\documentclass{article} % For LaTeX2e
\usepackage{iclr2027_conference,times}
\usepackage{amsmath,amsfonts,bm}

\def\eqref#1{equation~\ref{#1}}
\def\1{\bm{1}}

\DeclareMathAlphabet{\mathsfit}{\encodingdefault}{\sfdefault}{m}{sl}
\SetMathAlphabet{\mathsfit}{bold}{\encodingdefault}{\sfdefault}{bx}{n}

\usepackage{booktabs,graphicx,amsmath,amssymb,bm,wrapfig,xurl,url,makecell}
\usepackage{enumitem}
\usepackage{capt-of}
\usepackage{multirow}
\usepackage[table]{xcolor}
\newcommand{\fitop}[1]{\cellcolor{gray!15}#1}
\definecolor{typeblue}{RGB}{55,126,184}
\definecolor{typegreen}{RGB}{77,175,74}
\definecolor{princetonorange}{rgb}{1.0, 0.56, 0.0}
\newcommand{\ngicon}{\textcolor{typeblue}{\ensuremath{\bm{\bullet}}}}
\newcommand{\gicon}{\textcolor{typegreen}{\ensuremath{\bm{\blacktriangle}}}}
\newcommand{\dagg}{\textcolor{princetonorange}{\ensuremath{^{\boldsymbol{\dagger}}}}}

\newcommand{\spcAlign}{\Pi}                      % input-resolution alignment transform
\newcommand{\spcSt}{\operatorname{St}}           % multivariate Student-t density
\newcommand{\spcId}{\mathbf{Id}}                 % identity matrix
\newcommand{\spcDec}{\Phi}                       % binary decision function
\newcommand{\spcSigN}{\sigma_{\mathrm{n}}}       % additive-noise std
\newcommand{\spcSigG}{\sigma_{\mathrm{g}}}       % Gaussian-kernel std
\newcommand{\spcKer}{\kappa}                     % Gaussian-kernel size

\definecolor{citecolor}{RGB}{0,102,204}
\definecolor{refcolor}{RGB}{218, 47, 138}
\definecolor{skyblue}{RGB}{181, 212, 244}
\usepackage[normalem]{ulem}
\usepackage{hyperref}
\definecolor{linkblue}{RGB}{0,70,140}
\hypersetup{
  colorlinks=true,
  linkcolor=refcolor,
  citecolor=citecolor,
  urlcolor=black
}

\title{Not the Scalar, but Its Evolution: Robust Source Attribution for Generated Images}

\author{
Chao Wang$^{1}$,
Zijin Yang$^{1}$,
Yaofei Wang$^{2}$,
Yuang Qi$^{3}$,
Weiming Zhang$^{1}$,
Kejiang Chen$^{1}$\thanks{Corresponding author.} \\
$^{1}$University of Science and Technology of China \\
$^{2}$Hefei University of Technology \\
$^{3}$Beijing Zhongguancun Academy \\
\texttt{\{chaowang0708,bsmhmmlf\}@mail.ustc.edu.cn} \\
\texttt{wyf@hfut.edu.cn, qiyuang@bza.edu.cn} \\
\texttt{\{zhangwm,chenkj\}@ustc.edu.cn}
}

\iclrfinalcopy
\begin{document}
\maketitle
\addtocontents{toc}{\protect\setcounter{tocdepth}{-1}}
%======================================================================================================
\vspace{-0.15cm}
\begin{abstract}
\vspace{-0.15cm}
% non-adversarial
The widespread deployment of latent generative models (LGMs) makes generated image source attribution essential for copyright auditing and data governance, yet practical attribution should remain reliable under common post-processing and platform-induced processing. Existing reconstruction-based attribution methods summarize an image's reconstruction response to a single global \textit{scalar}, such as the reconstruction error or the ratio of two consecutive errors. These summaries discard channel-frequency structure, allowing image content and post-processing variation to obscure source evidence and making attribution after spreading difficult. We instead uncover a more stable source-discriminative cue under post-processing: the frequency-resolved \textit{evolution} of residual energy across repeated applications of the same reconstruction operator. Motivated by this insight, we characterize a phenomenon we term \textbf{\textit{Spectral Contraction}} and use it to build \textbf{\textit{SpCon}}, a source-attribution framework that accesses only the target model's VAE. SpCon obtains adjacent residuals through two deterministic reconstructions, builds a spectral-contraction representation from their frequency-resolved relative energy change, and aggregates Student-$t$ response densities for each source-condition pair over a finite set of fitted reference components. The same representation supports three settings according to non-belonging reference availability: none (\textit{Target-Only}, TO; belonging references only), external out-of-domain sources (\textit{External-Reference}, ER), or known in-domain competitors (\textit{Known-Reference}, KR). Across seven LGMs, TO, ER, and KR attain average AUROC/AP of 0.8786/0.6344, 0.9404/0.7886, and 0.9652/0.8626 over the 35 conditions, all outperforming the strongest baseline (0.8154/0.5277), and achieve 0.8581/0.5507, 0.9369/0.7549, and 0.9734/0.8773 on the three social platforms.
% , with a per-image inference time of only 0.384\,s.
\end{abstract}
%===============================================================================================

%===============================================================================================
\vspace{-0.4cm}
\section{Introduction}

Latent generative models continue to lower the barrier to high-fidelity image synthesis~\citep{LDM}, advancing applications such as artistic design and film production~\citep{DM-use} while also introducing risks of misuse. A bad actor may pass off a high-performance model's outputs as their own to exaggerate their capability, or disguise commercially generated content as original work for financial gain~\citep{IP-Problem}. Meanwhile, if unidentified generated data is mixed into the subsequent training of models in the same family, it may cause distribution degradation or even model collapse~\citep{Generated-Data}. Deciding merely whether an image is AI-generated is insufficient for these needs~\citep{FDU-ZXP}; one must further identify its source to support copyright auditing and source-aware data governance~\citep{AEDR, ZBH}.

This need is moving from voluntary practice toward regulation. In July 2026, Meta announced its commitment to the Code of Practice on Transparency of AI-Generated Content under the EU AI Act and continued to advance provenance verification~\citep{Meta}, while the California AI Transparency Act made AI content provenance a legal obligation and required covered providers to offer publicly accessible detection tools~\citep{California}. Beyond establishing provenance as a foundation of AI governance, this raises the bar on real-world applicability: generated images typically spread through social platforms and undergo unknown post-processing such as JPEG compression and noise (see \textcolor{refcolor}{Figure}~\ref{SpCon_Spread}). Attribution must therefore be accurate not only on clean images, but also robust to unknown post-processing and real platform processing chains.

Existing attribution methods fall into proactive and passive categories. Proactive methods embed a watermark or fingerprint during training or generation~\citep{Finger-1, Finger-2, Signature, GS}; their evidence is explicit but requires control over model parameters or sampling and cannot trace existing models without pre-deployed identifiers. Passive methods~\citep{UniversalAttribution, Forensic, MGI} instead exploit the statistical traces left by the generator. Supervised methods learn source features from many samples but generalize poorly out-of-domain, whereas reconstruction-based methods exploit the stronger reconstruction consistency of the target model on in-distribution images and summarize that response with a scalar, either the error itself~\citep{RONAN, LatentTracer} or a ratio derived from two successive reconstructions~\citep{AEDR}. Requiring no discriminator training and performing strongly on clean images, they are regarded as promising; yet post-processing can shift these global summaries and obscure source-dependent differences, degrading attribution performance (see \textcolor{refcolor}{Figure}~\ref{SpCon_Spread}). This raises a key question:

\begin{figure}[t]
\centering
\vspace{-0.1cm}
\includegraphics[width=\columnwidth]{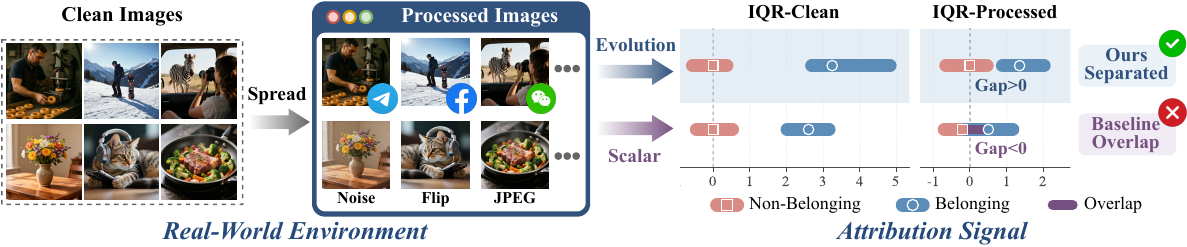}
\vspace{-0.7cm}
\caption{\textbf{Attribution signals must survive real-world spreading.} Interquartile range (IQR) of the standardized attribution score on clean and post-processed images. The \emph{scalar} reconstruction signal (\textbf{Baseline, AEDR}) separates belonging from non-belonging IQRs only when clean, whereas the spectral-contraction \emph{evolution} score of SpCon-ER (\textbf{Ours}) stays separated under both.}
\label{SpCon_Spread}
\vspace{-0.3cm}
\end{figure}

{\centering
\emph{\textbf{What reconstruction representation makes source evidence survive post-processing?}}
\par}

Despite their different procedures, reconstruction-based methods~\citep{RONAN, LatentTracer, AEDR} summarize the response globally rather than explicitly retaining its structure. Such summaries reveal the overall reconstruction discrepancy or its aggregate change, but not how source-dependent behavior is distributed across frequency scales. The core insight is \textbf{\textit{Not the Scalar, but Its Evolution}}: global scalar summaries of reconstruction responses can be unstable under post-processing, obscuring the separation between belonging and non-belonging images (see \textcolor{refcolor}{Figure}~\ref{SpCon_Spread}). Robust attribution therefore calls for richer, more stable signals. Rather than reducing reconstruction responses to an error or an aggregate comparison, we characterize how residual energy evolves across frequency scales over two successive reconstructions. Resolving the residuals by frequency preserves \emph{where} their energy changes, while comparing successive residuals captures \emph{how much} it contracts or expands at each scale. Unlike existing methods, which aggregate the residual response globally before scoring, this band-wise comparison retains structure that such summaries discard.

% This retains source-discriminative structure that global summaries discard. 

Concretely, we find that repeated application of the same target VAE induces source-related contraction and expansion across frequency scales, a phenomenon we term \textbf{\textit{Spectral Contraction}}. Building on it, we propose \textbf{\textit{SpCon}}, which measures this evolution as signed log-energy changes between two reconstruction residuals across RGB channels and radial frequency bands, preserving their relative channel-frequency profile rather than a scalar summary. It then models each fitted source-condition component with a Student-$t$ density and aggregates these components with equal weights. According to non-belonging reference availability, the same core representation supports three settings: \textit{Target-Only} (TO), \textit{External-Reference} (ER), and \textit{Known-Reference} (KR). The framework accesses only the target model's VAE and requires no discriminator training. Our main contributions include:

\vspace{-0.15cm}

\begin{enumerate}[leftmargin=*]
\item \textbf{Scalar Vulnerability.} We identify a limitation of globally summarized reconstruction responses: reconstruction errors and aggregate error comparisons discard response structure, allowing content and post-processing variation to obscure source evidence retained in richer representations.

\item \textbf{Spectral Contraction.} We propose SpCon, which characterizes the signed evolution of residual energy across RGB channels and radial frequency bands over two applications of a frozen target VAE. By combining frequency-resolved structure with relative residual comparison, SpCon captures source-discriminative behavior beyond global reconstruction statistics. Joint reference-density modeling supports TO, ER, and KR based on non-belonging reference availability.

\item \textbf{Experimental Validation.} Across seven LGMs, TO, ER, and KR attain average AUROC/AP of 0.8786/0.6344, 0.9404/0.7886, and 0.9652/0.8626 over 35 conditions, all above the strongest baseline AEDR (0.8154/0.5277), and reach 0.8581/0.5507, 0.9369/0.7549, and 0.9734/0.8773 on three real-world platforms, with ablations validating the contribution of Spectral Contraction.
\end{enumerate}
%======================================================================================================

%======================================================================================================
\section{Related Work}
\vspace{-0.1cm}
\textbf{Latent Generative Models.}
LGMs have become an important paradigm for image generation. They compress an image into a low-dimensional latent space with an autoencoder, run a generative process such as denoising diffusion~\citep{DM} or flow matching~\citep{FM} there, and decode back to pixels. Unlike the original pixel-space DDPM~\citep{DDPM}, latent diffusion~\citep{LDM} and subsequent latent-space models~\citep{SDXL} perform generation in an autoencoder latent space. In the generators studied here, the VAE~\citep{VAE} connects latent and pixel spaces. This makes it a relatively unified interface for cross-model source analysis, the premise that lets SpCon attribute using only a frozen target VAE.

\textbf{Proactive Source Attribution.}
Proactive methods embed a verifiable identifier during training or generation, mainly through image watermarking and model fingerprinting. Watermarking embeds the identifier in the pixel or frequency domain~\citep{MBRS, FIN, PIMoG}, or in the latent variables and initial noise of generation~\citep{Tree-Ring, GS, GS++}; fingerprinting binds model identity to generative behavior by modifying weights, fine-tuning the decoder, or introducing LoRA~\citep{Finger-1, Finger-2, Finger2Sem, Finger2DM}. Both give explicit and relatively robust evidence, but require control over model parameters or sampling, cannot trace existing models without pre-deployed identifiers, and may be vulnerable to removal or forgery~\citep{Black-Box, Removal-CCS}. In contrast, SpCon modifies neither training nor generation.

\textbf{Passive Source Attribution.}
Passive methods exploit the statistical traces left naturally by the generator, in supervised or reconstruction-based form. Supervised methods train a source classifier from generated samples~\citep{GANFingerprints-Data, DE-FAKE, UniversalAttribution, CDAL}, or cast attribution as few-shot one-class classification~\citep{OCC-CLIP}; they perform well in-distribution but generalize poorly out-of-domain. Reconstruction-based methods instead exploit the stronger reconstruction consistency of a target model on in-distribution images: RONAN~\citep{RONAN} optimizes the reconstruction input from a random start, LatentTracer~\citep{LatentTracer} moves this into the latent space using the VAE encoding as the starting point to reduce cost, and AEDR~\citep{AEDR} uses the ratio of two consecutive errors to mitigate the bias from image complexity. Their common limitation is that the attribution statistic does not explicitly preserve the channel-frequency structure of the reconstruction response. SpCon instead models the frequency-resolved evolution across adjacent residual levels. Compared with reconstruction-error methods~\citep{RONAN, LatentTracer}, it captures relative residual behavior rather than reconstruction fit alone; compared with AEDR~\citep{AEDR}, it preserves the spectral structure of this behavior rather than collapsing it into a global ratio. More detailed descriptions and comparisons are provided in \textcolor{refcolor}{Appendix}~\ref{Appendix_Baseline}.

%======================================================================================================

%======================================================================================================
\section{Method}
\begin{figure}[t]
\centering
\includegraphics[width=\columnwidth]{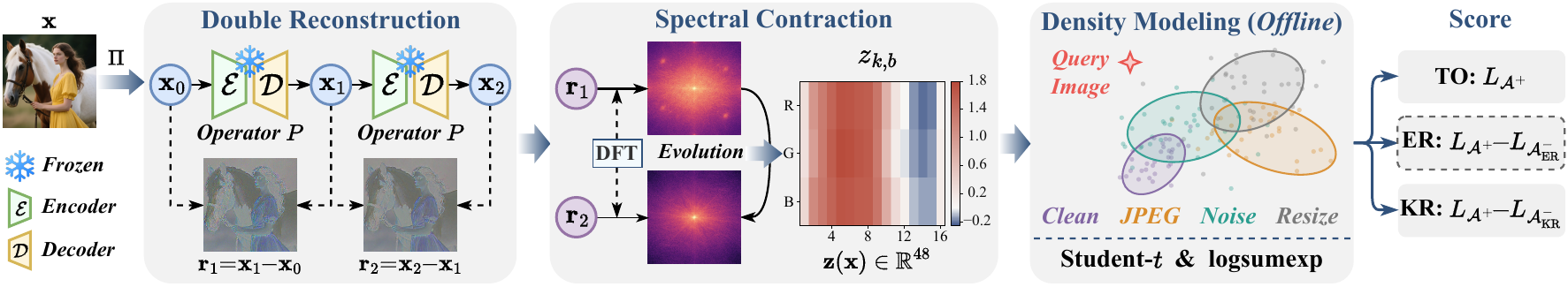}
\vspace{-0.5cm}
\caption{\textbf{Framework of SpCon.} A frozen target VAE $\mathcal{R}=(E,D)$ is applied twice to the aligned input to yield residuals $\mathbf{r}_1,\mathbf{r}_2$, whose band-wise log-energy difference forms the representation $\mathbf{z}(\mathbf{x})\in\mathbb{R}^{48}$. Each reference source-condition pair is modeled by a Student-$t$ and fused by logsumexp, giving a reference-adaptive score under the TO, ER, and KR settings.}
\label{Framework}
\vspace{-0.2cm}
\end{figure}

\vspace{-0.1cm}
In this section, we first define the problem and access assumptions, then detail the full SpCon pipeline: target-conditioned double reconstruction, spectral-contraction response, conditional Student-$t$ density modeling, and reference-adaptive scoring. The target model's VAE remains frozen, and no additional discriminator is trained. \textcolor{refcolor}{Figure}~\ref{Framework} shows the overall framework.

\subsection{Problem Formulation and Access Assumptions}
\label{Problem-Access}
\vspace{-0.1cm}
\textbf{Belonging and Non-Belonging Images.}
Given a target latent generative model $\mathcal{M}$, let $\mathcal{X}_{\mathcal{M}}$ denote the set of images generated by $\mathcal{M}$. Let $\mathbf{x}^{\star}$ be the original image before post-processing, and let the observed image be $\mathbf{x}=\mathcal{T}(\mathbf{x}^{\star})$, where $\mathcal{T}$ is a post-processing transform whose type and severity are unknown at test time and which includes the identity map corresponding to the clean condition ($\mathbf{x}^{\star}=\mathbf{x}$). Source verification for the target model $\mathcal{M}$ is defined as a binary hypothesis test: if $\mathbf{x}^{\star}\in\mathcal{X}_{\mathcal{M}}$, then $\mathbf{x}$ is a \textit{Belonging Image}; if $\mathbf{x}^{\star}\notin\mathcal{X}_{\mathcal{M}}$, then $\mathbf{x}$ is a \textit{Non-Belonging Image}.

\textbf{Attribution Goal.} 
Given a target model $\mathcal{M}$ and an observed image $\mathbf{x}$, the auditor aims to decide whether $\mathbf{x}$ originates from $\mathcal{M}$. In this process, the auditor neither modifies nor interferes with the training or generation of $\mathcal{M}$, and trains no additional discriminator. Formally, let $\mathcal{I}$ denote the image space; the attribution goal can be expressed as a binary decision function $\spcDec:\mathcal{I}\rightarrow\{0,1\}$, where $\spcDec(\mathbf{x})=1$ indicates that $\mathbf{x}$ is a belonging image and $\spcDec(\mathbf{x})=0$ indicates that it is non-belonging.

\textbf{Access Assumptions.}
We distinguish model-side access from reference-data access. On the model side, at query time the auditor invokes only the frozen VAE $\mathcal{R}=(E,D)$ of $\mathcal{M}$, not its generative backbone or sampler. On the reference-data side, all three settings need a small set of belonging images from $\mathcal{M}$ to fit the positive response distribution offline; these need not be auditor-generated and may come from provider-released samples, official model releases, or previously provenance-verified audit records. TO therefore denotes the absence of \emph{non-belonging} references, not of reference data. ER adds external non-belonging sources disjoint from the evaluated in-domain models; KR uses disjoint reference images from the evaluated in-domain non-belonging sources. The three settings share the same representation and probabilistic modeling (see \textcolor{refcolor}{Section}~\ref{Attribution_Score}).

\subsection{Double Reconstruction}
Since one reconstruction captures only the static input-output difference~\citep{RONAN, LatentTracer} while a second gives the local dynamic response of applying the same $\mathcal{R}$ again, two consecutive reconstructions form the shortest sequence for this response (see \textcolor{refcolor}{Appendix}~\ref{Multi_Step_SpCon}). We first align $\mathbf{x}$ to the native input resolution of $\mathcal{R}$ with a fixed alignment transform $\spcAlign$, and define the reconstruction operator $P$ through the deterministic center of the encoding posterior:
\begin{equation}
\mathbf{x}_0 = \spcAlign(\mathbf{x}), \quad P(\mathbf{v}) = D\bigl(\mathrm{mode}(E(\mathbf{v}))\bigr).
\label{Double_Recon}
\end{equation}
Keeping both reconstructions at this native resolution prevents extra resizing from perturbing the residual. We then apply $P$ twice and take the residual between adjacent states:
\begin{equation}
\mathbf{x}_1 = P(\mathbf{x}_0), \ \mathbf{r}_1 = \mathbf{x}_1 - \mathbf{x}_0, \qquad \mathbf{x}_2 = P(\mathbf{x}_1), \ \mathbf{r}_2 = \mathbf{x}_2 - \mathbf{x}_1.
\end{equation}
Here $\mathbf{r}_1$ and $\mathbf{r}_2$ are the changes induced by the first and second applications of $P$ to the aligned input $\mathbf{x}_0$. SpCon relies not on the absolute magnitude of either residual, but on the relative evolution from $\mathbf{r}_1$ to $\mathbf{r}_2$, characterizing the source-specific response under repeated application of $\mathcal{R}$.

\subsection{Spectral-Contraction Response}
\begin{wrapfigure}{r}{0.5\textwidth}
\begin{center}
\vspace{-1.5cm}
\centerline{\includegraphics[width=0.47\columnwidth]{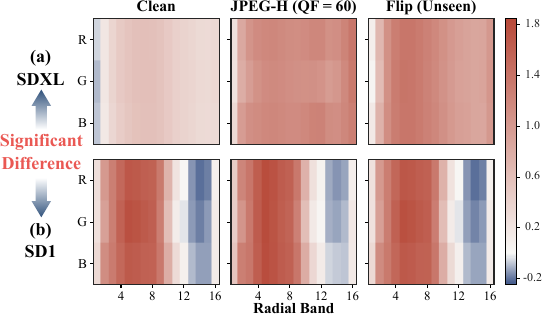}}
\vspace{-0.33cm}
\caption{\textbf{Spectral Contraction.} Mean $z_{k,b}$ under the SDXL target VAE for prompt-matched (a) belonging SDXL and (b) non-belonging SD1 images across Clean, JPEG-H, and Flip. All panels share a color scale; positive and negative values indicate residual-energy contraction and expansion.}
\label{SpCon_Phenomenon}
\end{center}
\vspace{-0.6cm}
\end{wrapfigure}

SpCon uses frequency bands as scale-indexed coordinates to compare two reconstruction residuals. Rather than using only their total energies, it preserves how their relative change is distributed across frequencies. For reconstruction level $j\in\{1,2\}$ and color channel $k\in\{R,G,B\}$, let $r_{j,k}$ be the $k$-th channel of $\mathbf{r}_j$. We compute the centered orthonormal 2D Discrete Fourier Transform (DFT) $\mathcal{F}$ and partition the frequency plane into $B$ fixed equal-width radial bands $\{\Omega_b\}_{b=1}^{B}$, with band energy
\begin{equation}
\mathcal{B}_b(g) = \sum_{\omega\in\Omega_b}\bigl|\mathcal{F}(g)(\omega)\bigr|^2, \quad b=1,\dots,B,
\label{Band_Energy}
\end{equation}
where $g$ denotes an arbitrary single-channel map, instantiated below as $r_{j,k}$. We use $B$=16 throughout unless stated otherwise. Radial aggregation preserves the low-to-high-frequency scale structure while reducing sensitivity to specific frequency directions. For each color channel and band, the log difference of the two residual energies defines the spectral-contraction response:
\begin{equation}
z_{k,b}(\mathbf{x}) =
\log\bigl(\max\{\mathcal{B}_b(r_{1,k}),\varepsilon\}\bigr) -
\log\bigl(\max\{\mathcal{B}_b(r_{2,k}),\varepsilon\}\bigr),
\label{SpCon_Z}
\end{equation}
where $\varepsilon=10^{-12}$ ensures numerical stability. If $z_{k,b}(\mathbf{x})>0$, the residual energy at this channel-frequency scale contracts after the second reconstruction; if $z_{k,b}(\mathbf{x})<0$, it expands (see \textcolor{refcolor}{Figure}~\ref{SpCon_Phenomenon}). 

Two properties matter when all relevant energies exceed the numerical floor. First, at $B{=}1$ with RGB energies summed before taking logarithms, Parseval's identity gives $z{=}-2\log(\|\mathbf{r}_2\|_2/\|\mathbf{r}_1\|_2)$, a monotone transform of the global MSE ratio computed from the same residual pair and thus connected to AEDR's error-ratio component. Finer partitions retain a vector of relative responses rather than one global statistic. Second, multiplicative factors shared by the two residual levels cancel within each band, even when they vary across bands; such band-dependent factors generally do not cancel after global aggregation. \textcolor{refcolor}{Appendix}~\ref{Response_Properties} states both properties precisely together with their scope, and \textcolor{refcolor}{Appendix}~\ref{Multi_Step_SpCon} analyzes longer reconstruction sequences. We concatenate the responses of the three color channels and 16 bands in a fixed order to obtain the spectral-contraction representation $\mathbf{z}(\mathbf{x})\in\mathbb{R}^{48}$. Under the SDXL VAE, belonging images follow a more uniform profile, whereas non-belonging SD1 images contract strongly at low-to-mid frequencies (see \textcolor{refcolor}{Figure}~\ref{SpCon_Phenomenon}).

\vspace{-0.05cm}
\subsection{Conditional Density Modeling}
The retained channel-frequency responses are evaluated jointly rather than scored independently or averaged using their raw energy magnitudes. SpCon models each fitted source-condition pair with a multivariate Student-$t$ density and a full scale matrix~\citep{Student-Mix}. Ledoit-Wolf shrinkage stabilizes covariance estimation from limited references, while the Student-$t$ tails accommodate off-center responses (see \textcolor{refcolor}{Appendices}~\ref{Nu_Sensitivity} and~\ref{Scale_Matrix_Ablation}). Let $a$ be a reference source and $c\in\mathcal{C}_{\mathrm{ref}}$ a reference condition with transform $\mathcal{T}_c$; under the target VAE, the images of $(a,c)$ map to $\mathbf{z}(\mathbf{x})\in\mathbb{R}^{48}$, with density
\begin{equation}
q_{a,c}(\mathbf{z}) = \spcSt\bigl(\mathbf{z};\boldsymbol{\mu}_{a,c},\boldsymbol{\Sigma}_{a,c},\nu\bigr),
\end{equation}
where $\spcSt$ denotes the 48-dimensional multivariate Student-$t$ density, $\boldsymbol{\mu}_{a,c}$ and $\boldsymbol{\Sigma}_{a,c}$ are the location parameter and full scale matrix, and $\nu$ is the degrees of freedom. Given $n$ reference responses for each component, $\boldsymbol{\mu}_{a,c}$ uses the sample mean, and the covariance uses the Ledoit-Wolf~\citep{Ledoit-Wolf} shrinkage estimate $\widehat{\mathbf{C}}_{a,c}$. We set degrees of freedom and full scale matrix as
\begin{equation}
\nu = 2 + \sqrt{n}, \quad \boldsymbol{\Sigma}_{a,c} = \widehat{\mathbf{C}}_{a,c}\,\frac{\nu-2}{\nu}.
\label{Student_Params}
\end{equation}
Under the main protocol $n=100$, so $\nu=12$. The rescaling aligns the Student-$t$ covariance with $\widehat{\mathbf{C}}_{a,c}$, while finite $\nu$ retains heavier tails than a Gaussian to accommodate off-center responses under unfitted transforms. All parameters are estimated from the reference-fitting set and fixed before testing. At test time SpCon does not infer the unknown transform; it aggregates the finite set of fitted source-condition components with an equal-weight mixture~\citep{Student-Mix}:
\begin{equation}
L_{\mathcal{A}}(\mathbf{z}) = \log\!\bigg[\frac{1}{|\mathcal{A}|}
\sum_{(a,c)\in\mathcal{A}} q_{a,c}(\mathbf{z})\bigg].
\end{equation}

\vspace{-0.45cm}
\subsection{Reference-Adaptive Scoring}
\label{Attribution_Score}
We consider three settings based on reference availability: \textit{Target-Only} (TO) uses only target-model outputs as belonging references; \textit{External-Reference} (ER) and \textit{Known-Reference} (KR) additionally use external and in-domain non-belonging references, respectively. Let $\rho\in\{\mathrm{TO},\mathrm{ER},\mathrm{KR}\}$ index these settings. $\mathcal{A}^{+}$ pairs the target source with each fitted condition in $\mathcal{C}_{\mathrm{ref}}$; $\mathcal{A}^{-}_{\rho}$ pairs the same conditions with external sources for ER or evaluated non-belonging sources for KR. All components are fitted using reference images disjoint from the test set. The attribution score is
\begin{equation}
S^{(\rho)}_{\mathcal{M}}(\mathbf{x}) =
\begin{cases}
L_{\mathcal{A}^{+}}(\mathbf{z}), & \rho = \mathrm{TO}, \\[4pt]
L_{\mathcal{A}^{+}}(\mathbf{z}) - L_{\mathcal{A}^{-}_{\rho}}(\mathbf{z}),
& \rho\in\{\mathrm{ER},\mathrm{KR}\}.
\end{cases}
\label{Attribution_Score_Eq}
\end{equation}
The three settings share the same double reconstruction, spectral-contraction response, and density modeling, differing only in the reference set. A larger score more strongly supports belonging. Binary deployment requires a target-specific threshold (see \textcolor{refcolor}{Appendix}~\ref{More_Metrics}).
%======================================================================================================

%======================================================================================================
\begin{table}[t]
\centering
\caption{Post-processing operation families and severity levels. Gray cells mark the six medium-severity Seen operations used for fitting; ``\ngicon''/``\gicon'' denote non-geometric/geometric distortions. Resize and Crop factors apply to both height and width. Flip has no severity levels.}
\vspace{0.1cm}
\scriptsize
\setlength{\tabcolsep}{4pt}
\renewcommand{\arraystretch}{1.05}
\resizebox{\columnwidth}{!}{
\begin{tabular}{ccccc@{\hspace{6pt}\vrule width 0.3pt\hspace{6pt}}ccccc}
\toprule
\multicolumn{5}{c}{\textbf{Seen Operations}} & \multicolumn{5}{c}{\textbf{Unseen Operations}} \\
\cmidrule{1-10}
Operation & Low (L) & Medium (M) & High (H) & Type & Operation & Low (L) & Medium (M) & High (H) & Type \\
\midrule
JPEG & $\mathrm{QF}{=}90$ & \fitop{$\mathrm{QF}{=}75$} & $\mathrm{QF}{=}60$ & \ngicon & WebP & $\mathrm{QF}{=}90$ & $\mathrm{QF}{=}75$ & $\mathrm{QF}{=}60$ & \ngicon \\
Noise & $\spcSigN{=}5$ & \fitop{$\spcSigN{=}10$} & $\spcSigN{=}15$ & \ngicon & Sharpen & $\alpha{=}0.5$ & $\alpha{=}1.0$ & $\alpha{=}1.5$ & \ngicon \\
Blur & $\spcKer{=}3$ & \fitop{$\spcKer{=}5$} & $\spcKer{=}9$ & \ngicon & Contrast & $\times1.2$ & $\times1.6$ & $\times2.0$ & \ngicon \\
Brightness & $\times1.2$ & \fitop{$\times1.6$} & $\times2.0$ & \ngicon & Saturation & $\times0.8$ & $\times0.6$ & $\times0.4$ & \ngicon \\
Resize & $\times7/8$ & \fitop{$\times6/8$} & $\times5/8$ & \gicon & Crop & $\times7/8$ & $\times6/8$ & $\times5/8$ & \gicon \\
Rotation & $\pm2^\circ$ & \fitop{$\pm5^\circ$} & $\pm8^\circ$ & \gicon & Flip & \multicolumn{3}{c}{Random Horizontal / Vertical / Both} & \gicon \\
\bottomrule
\end{tabular}}
\label{Post_Processing}
\vspace{-0.5cm}
\end{table}

\begin{table*}[t]
\centering
\caption{Attribution performance across different target models under clean conditions.}
\vspace{0.1cm}
\scriptsize
\setlength{\tabcolsep}{1.5pt}
\renewcommand{\arraystretch}{1.1}
\resizebox{\textwidth}{!}{
\begin{tabular}{l|ccccccc}
\toprule
\textbf{Method} & \textbf{SD1} & \textbf{SD2} & \textbf{SD3} & \textbf{SDXL} & \textbf{FLUX.1} & \textbf{FLUX.2} & \textbf{PixArt} \\
\midrule
OCC-CLIP & 0.5287 / 0.1345 & 0.5839 / 0.1583 & 0.7203 / 0.2640 & 0.6689 / 0.2136 & 0.6741 / 0.1981 & 0.6862 / 0.1952 & 0.8132 / 0.3322 \\
LatentTracer & 0.9978 / 0.9775 & 0.9976 / 0.9855 & 0.7697 / 0.3436 & 0.7645 / 0.2759 & 0.6789 / 0.2049 & 0.6152 / 0.1703 & 0.5437 / 0.1295 \\
AEDR & \underline{0.9993} / \underline{0.9954} & 0.9981 / 0.9754 & 0.9961 / 0.9638 & 0.9993 / 0.9884 & 0.9929 / 0.9250 & 0.9979 / 0.9858 & 0.9980 / 0.9846 \\
\rowcolor{skyblue!20}
SpCon-TO & 0.9609 / 0.9033 & 0.9695 / 0.9229 & 0.9849 / 0.9630 & 0.9873 / 0.9592 & 0.9971\dagg / 0.9898\dagg & 0.9917 / 0.9748 & 0.9810 / 0.9431 \\
\rowcolor{skyblue!30}
SpCon-ER & 0.9981 / 0.9913 & \underline{0.9997}\dagg / \underline{0.9979}\dagg & \underline{0.9982} / \underline{0.9927}\dagg & \underline{0.9999} / \underline{0.9991}\dagg & \underline{0.9999}\dagg / \underline{0.9994}\dagg & \underline{0.9998}\dagg / \underline{0.9988}\dagg & \underline{0.9999}\dagg / \underline{0.9994}\dagg \\
\rowcolor{skyblue!40}
SpCon-KR & \textbf{0.9999}\dagg / \textbf{0.9991}\dagg & \textbf{1.0000}\dagg / \textbf{1.0000}\dagg & \textbf{1.0000}\dagg / \textbf{0.9997}\dagg & \textbf{0.9999}\dagg / \textbf{0.9995}\dagg & \textbf{1.0000}\dagg / \textbf{0.9999}\dagg & \textbf{1.0000}\dagg / \textbf{1.0000}\dagg & \textbf{1.0000}\dagg / \textbf{0.9997}\dagg \\
\bottomrule
\end{tabular}}
\label{SpCon_Clean}
\vspace{-0.3cm}
\end{table*}

\section{Experiments}

\subsection{Experimental Setup}
\label{Experimental_Setup}
\textbf{Models and Data.}
We experiment on the public DNA-30K dataset~\citep{DNA}, using nine models and three natural-image datasets: Stable-Diffusion-v1-5 (SD1), Stable-Diffusion-2-1-base (SD2), Stable-Diffusion-3.5-large-turbo (SD3), Stable-Diffusion-xl-base-1.0 (SDXL), FLUX.1-dev (FLUX.1), FLUX.2-dev (FLUX.2), PixArt-XL-2-1024-MS (PixArt), Qwen-Image (Qwen), Kandinsky-3 (KD3), MS COCO, LAION, and ImageNet. The main evaluation covers seven models (SD1, SD2, SD3, SDXL, FLUX.1, FLUX.2, PixArt) with COCO as the natural-image source. We fit distributions on the DNA-30K calibration set (100 images / source) and evaluate on the test set (500 images / source). SpCon-ER uses KD3 and ImageNet as default external references disjoint from the main evaluation, while Qwen and LAION are reserved for the ablation study in \textcolor{refcolor}{Appendix}~\ref{Appendix_External_Reference}.

\textbf{Post-Processing Protocol.}
We consider 12 post-processing operations, evenly split into 6 Seen and 6 Unseen (see \textcolor{refcolor}{Table}~\ref{Post_Processing} and \textcolor{refcolor}{Appendix}~\ref{Appendix_Post_Processing}). SpCon is fitted only on Clean images and the medium severity of each of the 6 Seen operations; the low and high Seen severities and all Unseen operations are used only for testing, with operation type and severity unknown at test time. The synthetic evaluation thus comprises 35 conditions (1 Clean, 18 Seen, 16 Unseen). We test on images uploaded through Telegram, WeChat, and Facebook, none of which participate in fitting (see \textcolor{refcolor}{Appendix}~\ref{Real_World_Platform}).

\textbf{Baselines.}
We compare three representative passive methods: OCC-CLIP~\citep{OCC-CLIP}, which performs few-shot one-class classification on frozen CLIP; LatentTracer~\citep{LatentTracer}, which optimizes the VAE latent by gradient descent to minimize reconstruction error; and AEDR~\citep{AEDR}, which attributes via a homogeneity-calibrated loss ratio. We follow each method's official implementation; extensions and protocol-matched variants are detailed in \textcolor{refcolor}{Appendix}~\ref{Appendix_Baseline}.

\textbf{Evaluation Metrics.}
We report AUROC and Average Precision (AP), both computed from the full score ranking without threshold selection. Since each evaluation unit has a positive ratio of 1/8, a random ranking yields AUROC=0.5 and AP=0.125. We assess significance with 1,000 stratified paired bootstrap resamples and the two-sided 95\% confidence interval (CI) of the difference: for each scope and metric, we compare each SpCon variant against the strongest official baseline and mark significance with \dagg \,when the interval's lower bound exceeds zero. Full significance results are reported in \textcolor{refcolor}{Appendix}~\ref{BSA}; prevalence sensitivity, low-FPR operating points, and fixed-threshold evaluation are reported in \textcolor{refcolor}{Appendix}~\ref{More_Metrics}. Best and second-best results are \textbf{bold} and \underline{underlined}.

\textbf{Implementation Details.}
Experiments were conducted using Python 3.12.0, PyTorch 2.9.1, and CUDA 12.8, with each run independently using a single NVIDIA RTX PRO 6000 GPU.

\subsection{Performance under Clean Conditions}
\textcolor{refcolor}{Table}~\ref{SpCon_Clean} reports attribution performance on clean images. KR is near-perfect across all seven targets (average AUROC/AP=1.0000/0.9997), ER achieves AUROC$\geq$0.9981 and AP$\geq$0.9913 for every target, and TO achieves at least 0.9609 AUROC on every target without any non-belonging reference. The baselines instead show clear model-family sensitivity: LatentTracer is strong on SD1 and SD2 but drops on the other targets (consistent with prior work~\citep{AEDR}), while OCC-CLIP also varies across targets. AEDR remains strong across all seven targets, as do the SpCon variants.

\begin{table*}[t]
\centering
\vspace{-0.3cm}
\caption{Performance under Seen post-processing operations at different severity levels.}
\vspace{0.1cm}
\scriptsize
\setlength{\tabcolsep}{4pt}
\renewcommand{\arraystretch}{0.4}
\resizebox{\textwidth}{!}{
\begin{tabular}{cc|ccc|>{\columncolor{skyblue!20}}c>{\columncolor{skyblue!30}}c>{\columncolor{skyblue!40}}c}
\toprule
\textbf{Operation} & \textbf{Severity} & \textbf{OCC-CLIP} & \textbf{LatentTracer} & \textbf{AEDR}
& \multicolumn{1}{c}{\textbf{SpCon-TO}}
& \multicolumn{1}{c}{\textbf{SpCon-ER}}
& \multicolumn{1}{c}{\textbf{SpCon-KR}} \\
\midrule
\multirow{5.5}{*}{JPEG}
& L & 0.6660 / 0.2145 & 0.6879 / 0.2485 & 0.9503 / 0.6878 & 0.8447 / 0.4975 & \underline{0.9597}\dagg / \underline{0.8305}\dagg & \textbf{0.9914}\dagg / \textbf{0.9583}\dagg \\
& M (Fit) & 0.6570 / 0.2061 & 0.6279 / 0.1820 & 0.8728 / 0.4733 & 0.8906\dagg / 0.6637\dagg & \underline{0.9438}\dagg / \underline{0.7676}\dagg & \textbf{0.9811}\dagg / \textbf{0.9138}\dagg \\
& H & 0.6449 / 0.2010 & 0.5920 / 0.1609 & 0.8191 / 0.3748 & 0.8547\dagg / 0.5369\dagg & \underline{0.9168}\dagg / \underline{0.6730}\dagg & \textbf{0.9577}\dagg / \textbf{0.8068}\dagg \\
\cmidrule{1-8}
\multirow{5.5}{*}{Noise}
& L & 0.6486 / 0.2057 & 0.7496 / 0.2352 & 0.8575 / 0.5678 & 0.8185 / 0.3519 & \underline{0.9028}\dagg / \underline{0.6658}\dagg & \textbf{0.9545}\dagg / \textbf{0.8010}\dagg \\
& M (Fit) & 0.6365 / 0.1946 & 0.6355 / 0.1928 & 0.6647 / 0.2915 & 0.8783\dagg / 0.5487\dagg & \underline{0.9219}\dagg / \underline{0.6642}\dagg & \textbf{0.9671}\dagg / \textbf{0.8485}\dagg \\
& H & 0.6305 / 0.1911 & 0.5868 / 0.1916 & 0.5747 / 0.1999 & 0.8456\dagg / 0.4924\dagg & \underline{0.8665}\dagg / \underline{0.5330}\dagg & \textbf{0.9201}\dagg / \textbf{0.6530}\dagg \\
\cmidrule{1-8}
\multirow{5.5}{*}{Blur}
& L & 0.6620 / 0.2099 & 0.6637 / 0.2931 & 0.8976 / 0.5689 & 0.8126 / 0.5291 & \underline{0.9439}\dagg / \underline{0.8058}\dagg & \textbf{0.9637}\dagg / \textbf{0.8614}\dagg \\
& M (Fit) & 0.6578 / 0.2079 & 0.6277 / 0.2564 & 0.8305 / 0.4265 & 0.8692\dagg / 0.6141\dagg & \underline{0.9573}\dagg / \underline{0.8226}\dagg & \textbf{0.9765}\dagg / \textbf{0.8886}\dagg \\
& H & 0.6560 / 0.2070 & 0.6198 / 0.2434 & 0.7984 / 0.3666 & 0.8474\dagg / 0.5877\dagg & \underline{0.9413}\dagg / \underline{0.7759}\dagg & \textbf{0.9615}\dagg / \textbf{0.8369}\dagg \\
\cmidrule{1-8}
\multirow{5.5}{*}{Brightness}
& L & 0.6617 / 0.2102 & 0.7483 / 0.3702 & 0.9939 / 0.9437 & 0.9493 / 0.8363 & \underline{0.9971}\dagg / \underline{0.9853}\dagg & \textbf{0.9996}\dagg / \textbf{0.9981}\dagg \\
& M (Fit) & 0.6528 / 0.2079 & 0.6871 / 0.2288 & 0.9734 / 0.8127 & 0.9280 / 0.7858 & \underline{0.9917}\dagg / \underline{0.9568}\dagg & \textbf{0.9978}\dagg / \textbf{0.9884}\dagg \\
& H & 0.6449 / 0.2078 & 0.6421 / 0.1932 & 0.9373 / 0.6594 & 0.8915 / 0.6908\dagg & \underline{0.9791}\dagg / \underline{0.8937}\dagg & \textbf{0.9882}\dagg / \textbf{0.9426}\dagg \\
\cmidrule{1-8}
\multirow{5.5}{*}{Resize}
& L & 0.6678 / 0.2137 & 0.7154 / 0.2966 & 0.6017 / 0.1721 & 0.8580\dagg / 0.4108\dagg & \underline{0.9414}\dagg / \underline{0.7936}\dagg & \textbf{0.9894}\dagg / \textbf{0.9224}\dagg \\
& M (Fit) & 0.6700 / 0.2155 & 0.6840 / 0.2234 & 0.5464 / 0.1412 & 0.9507\dagg / 0.8525\dagg & \underline{0.9936}\dagg / \underline{0.9653}\dagg & \textbf{0.9988}\dagg / \textbf{0.9926}\dagg \\
& H & 0.6718 / 0.2170 & 0.6574 / 0.2026 & 0.5217 / 0.1359 & 0.7013\dagg / 0.3292\dagg & \textbf{0.9379}\dagg / \textbf{0.7429}\dagg & \underline{0.9001}\dagg / \underline{0.7005}\dagg \\
\cmidrule{1-8}
\multirow{5.5}{*}{Rotation}
& L & 0.6670 / 0.2121 & 0.5628 / 0.1592 & 0.6469 / 0.1900 & 0.9158\dagg / 0.7074\dagg & \underline{0.9530}\dagg / \underline{0.7777}\dagg & \textbf{0.9789}\dagg / \textbf{0.8835}\dagg \\
& M (Fit) & 0.6663 / 0.2110 & 0.5579 / 0.1580 & 0.6326 / 0.1865 & 0.9157\dagg / 0.7152\dagg & \underline{0.9520}\dagg / \underline{0.7746}\dagg & \textbf{0.9767}\dagg / \textbf{0.8748}\dagg \\
& H & 0.6677 / 0.2114 & 0.5548 / 0.1561 & 0.6255 / 0.1889 & 0.9091\dagg / 0.6799\dagg & \underline{0.9467}\dagg / \underline{0.7451}\dagg & \textbf{0.9704}\dagg / \textbf{0.8463}\dagg \\
\bottomrule
\end{tabular}}
\label{SpCon_Seen}
\vspace{-0.5cm}
\end{table*}

\begin{table*}[t]
\centering
\caption{Performance under Unseen post-processing operations at different severity levels.}
\vspace{0.1cm}
\scriptsize
\setlength{\tabcolsep}{4pt}
\renewcommand{\arraystretch}{0.4}
\resizebox{\textwidth}{!}{
\begin{tabular}{cc|ccc|>{\columncolor{skyblue!20}}c>{\columncolor{skyblue!30}}c>{\columncolor{skyblue!40}}c}
\toprule
\textbf{Operation} & \textbf{Severity} & \textbf{OCC-CLIP} & \textbf{LatentTracer} & \textbf{AEDR}
& \multicolumn{1}{c}{\textbf{SpCon-TO}}
& \multicolumn{1}{c}{\textbf{SpCon-ER}}
& \multicolumn{1}{c}{\textbf{SpCon-KR}} \\
\midrule
\multirow{5.5}{*}{WebP}
& L & 0.6662 / 0.2136 & 0.6974 / 0.2728 & \underline{0.9558} / 0.7164 & 0.8331 / 0.4745 & 0.9487 / \underline{0.8048}\dagg & \textbf{0.9878}\dagg / \textbf{0.9435}\dagg \\
& M & 0.6653 / 0.2131 & 0.6277 / 0.1782 & 0.8675 / 0.4479 & 0.8331 / 0.5361\dagg & \underline{0.9115}\dagg / \underline{0.6604}\dagg & \textbf{0.9492}\dagg / \textbf{0.7994}\dagg \\
& H & 0.6606 / 0.2087 & 0.6044 / 0.1656 & 0.8317 / 0.3760 & 0.7947 / 0.4637\dagg & \underline{0.8812}\dagg / \underline{0.5873}\dagg & \textbf{0.9219}\dagg / \textbf{0.7069}\dagg \\
\cmidrule{1-8}
\multirow{5.5}{*}{Sharpen}
& L & 0.6717 / 0.2169 & 0.7721 / 0.4083 & 0.9951 / 0.9526 & 0.9718 / 0.9144 & \underline{0.9981}\dagg / \underline{0.9909}\dagg & \textbf{0.9998}\dagg / \textbf{0.9990}\dagg \\
& M & 0.6730 / 0.2184 & 0.7520 / 0.3168 & 0.9863 / 0.8730 & 0.9538 / 0.8351 & \underline{0.9950}\dagg / \underline{0.9766}\dagg & \textbf{0.9994}\dagg / \textbf{0.9965}\dagg \\
& H & 0.6740 / 0.2194 & 0.7268 / 0.2577 & 0.9695 / 0.7650 & 0.9335 / 0.7454 & \underline{0.9874}\dagg / \underline{0.9400}\dagg & \textbf{0.9977}\dagg / \textbf{0.9871}\dagg \\
\cmidrule{1-8}
\multirow{5.5}{*}{Contrast}
& L & 0.6641 / 0.2100 & 0.7491 / 0.3850 & 0.9940 / 0.9443 & 0.9525 / 0.8499 & \underline{0.9976}\dagg / \underline{0.9872}\dagg & \textbf{0.9997}\dagg / \textbf{0.9983}\dagg \\
& M & 0.6569 / 0.2063 & 0.7035 / 0.2580 & 0.9764 / 0.8102 & 0.9250 / 0.7631 & \underline{0.9907}\dagg / \underline{0.9521}\dagg & \textbf{0.9975}\dagg / \textbf{0.9866}\dagg \\
& H & 0.6496 / 0.2036 & 0.6671 / 0.2094 & 0.9497 / 0.6745 & 0.8914 / 0.6617 & \underline{0.9748}\dagg / \underline{0.8802}\dagg & \textbf{0.9882}\dagg / \textbf{0.9440}\dagg \\
\cmidrule{1-8}
\multirow{5.5}{*}{Saturation}
& L & 0.6664 / 0.2118 & 0.7575 / 0.4266 & 0.9967 / 0.9693 & 0.9777 / 0.9379 & \underline{0.9991}\dagg / \underline{0.9960}\dagg & \textbf{0.9999}\dagg / \textbf{0.9996}\dagg \\
& M & 0.6645 / 0.2093 & 0.7448 / 0.3973 & 0.9953 / 0.9567 & 0.9699 / 0.9134 & \underline{0.9985}\dagg / \underline{0.9934}\dagg & \textbf{0.9998}\dagg / \textbf{0.9992}\dagg \\
& H & 0.6618 / 0.2067 & 0.7241 / 0.3498 & 0.9922 / 0.9293 & 0.9582 / 0.8697 & \underline{0.9967}\dagg / \underline{0.9861}\dagg & \textbf{0.9998}\dagg / \textbf{0.9986}\dagg \\
\cmidrule{1-8}
\multirow{5.5}{*}{Crop}
& L & 0.6753 / 0.2175 & 0.5535 / 0.1513 & 0.6155 / 0.1923 & 0.8465\dagg / 0.4477\dagg & \underline{0.8986}\dagg / \underline{0.5720}\dagg & \textbf{0.9330}\dagg / \textbf{0.6713}\dagg \\
& M & 0.6787 / 0.2196 & 0.5212 / 0.1374 & 0.5581 / 0.1595 & 0.7229\dagg / 0.2527\dagg & \underline{0.7523}\dagg / \underline{0.3186}\dagg & \textbf{0.8343}\dagg / \textbf{0.4098}\dagg \\
& H & \underline{0.6785} / \underline{0.2173} & 0.4830 / 0.1246 & 0.5091 / 0.1411 & 0.6460 / 0.1866 & 0.6435 / 0.2072 & \textbf{0.7542}\dagg / \textbf{0.2796}\dagg \\
\cmidrule{1-8}
Flip
& Random & 0.6670 / 0.2063 & 0.5593 / 0.1609 & 0.6032 / 0.2001 & 0.8787\dagg / 0.5728\dagg & \underline{0.8965}\dagg / \underline{0.5795}\dagg & \textbf{0.9481}\dagg / \textbf{0.7541}\dagg \\
\bottomrule
\end{tabular}}
\label{SpCon_Unseen}
\vspace{-0.3cm}
\end{table*}

\subsection{Robustness to Seen Post-Processing}
We separate the 18 Seen conditions by fitting exposure, since only the medium severity of each Seen operation participates in fitting (see \textcolor{refcolor}{Table}~\ref{SpCon_Seen}). On the six Seen-Fit conditions, TO, ER, and KR attain 0.9054/0.6967, 0.9600/0.8252, and 0.9830/0.9178 AUROC/AP; on the 12 Seen-Heldout severities they reach 0.8540/0.5542, 0.9405/0.7685, and 0.9646/0.8509, against 0.7687/0.4213 for AEDR. ER and KR significantly exceed the strongest official baseline on all 2$\times$12=24 held-out condition-metric comparisons, so we read severity generalization from Seen-Heldout rather than from the aggregate Seen result. KR leads ER on all six fitted conditions but trails it under Resize (H), the only inversion across the 18 conditions. This severity retains less than 40\% of the original image area and redistributes spectral energy, making it a strong perturbation to the frequency-domain response used by SpCon. Yet ER and KR remain close at 0.9379/0.7429 and 0.9001/0.7005 AUROC/AP, respectively, and both significantly outperform the strongest baseline OCC-CLIP.

\subsection{Generalization to Unseen Post-Processing}
Across the 16 Unseen conditions, TO, ER, and KR improve over AEDR by 0.0183/0.0198, 0.0671/0.1453, and 0.0946/0.2104 in mean AUROC/AP, respectively (see \textcolor{refcolor}{Table}~\ref{SpCon_Unseen}). Against the strongest official baseline per condition, KR achieves significant gains in all 32 condition-metric comparisons and ER in 29. Gains are especially pronounced under Crop and Flip; for Sharpen, Contrast, and Saturation, AUROC margins narrow as AEDR approaches the ceiling, while AP gains remain clear at higher severities. Across the 28 held-out conditions (12 Seen severities and 16 Unseen conditions), TO, ER, and KR reach 0.8691/0.6099, 0.9342/0.7734, and 0.9602/0.8459 AUROC/AP, versus 0.8222/0.5415 for AEDR. Sensitivity to alternative partitions is reported in \textcolor{refcolor}{Appendix}~\ref{Appendix_Seen_Unseen}.

\begin{figure}[t]
\centering
\includegraphics[width=\columnwidth]{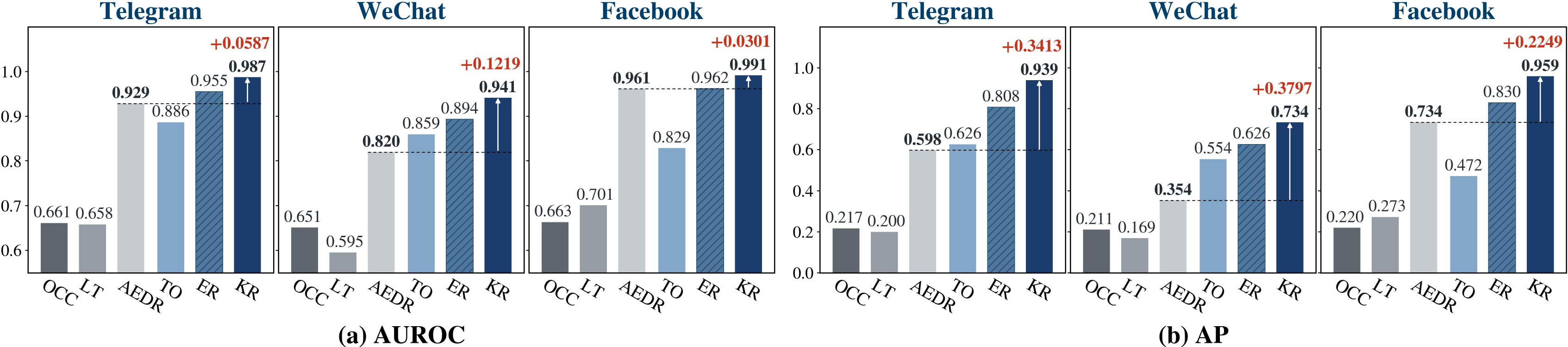}
\vspace{-0.6cm}
\caption{Attribution performance on images shared through Telegram, WeChat, and Facebook, averaged over the seven models. Annotations give the gain of SpCon-KR over the strongest baseline.}
\label{SpCon_Platform}
\vspace{-0.2cm}
\end{figure}

\begin{table*}[t]
\centering
\begin{minipage}[t]{0.37\textwidth}
\centering
\vspace{-0.1cm}
\caption{Inference time (s/image).}
\vspace{0.1cm}
\scriptsize
\setlength{\tabcolsep}{3pt}
\renewcommand{\arraystretch}{1}
\resizebox{\linewidth}{!}{
\begin{tabular}{l|ccc>{\columncolor{skyblue!30}}c}
\toprule
\textbf{Model} & \textbf{OCC} & \textbf{LT} & \textbf{AEDR} &
\multicolumn{1}{c}{\textbf{SpCon}} \\
\midrule
SD1     & 0.024 & 7.506  & 0.135 & 0.132 \\
SD2     & 0.024 & 7.487  & 0.130 & 0.123 \\
SD3     & 0.061 & 26.819 & 0.462 & 0.431 \\
SDXL    & 0.066 & 37.763 & 0.605 & 0.571 \\
FLUX.1  & 0.064 & 26.841 & 0.460 & 0.436 \\
FLUX.2  & 0.065 & 26.851 & 0.457 & 0.426 \\
PixArt  & 0.064 & 35.852 & 0.591 & 0.568 \\
\midrule
Average & \textbf{0.053} & 24.160 & 0.406 & \underline{0.384} \\
Speedup & \textbf{455.8$\times$} & 1.0$\times$ & 59.5$\times$ & \underline{62.9$\times$} \\
\bottomrule
\end{tabular}}
\label{SpCon_Time}
\end{minipage}
\hfill
\begin{minipage}[t]{0.62\textwidth}
\centering
\vspace{-0.1cm}
\caption{Ablation on the spectral-contraction response.}
\vspace{0.1cm}
\scriptsize
\setlength{\tabcolsep}{3pt}
\renewcommand{\arraystretch}{0.918}
\resizebox{\linewidth}{!}{
\begin{tabular}{cc|ccc}
\toprule
\textbf{Response} & \textbf{Type} & \textbf{Clean} & \textbf{Seen} & \textbf{Unseen} \\
\midrule
\multirow[c]{3}{*}{\shortstack{First Residual\\[1.2pt]{\scriptsize(\textit{Single-level})}}}
& TO & 0.8059 / 0.4865 & 0.7692 / 0.4047 & 0.7732 / 0.3956 \\
& ER & 0.9159 / 0.7351 & 0.8422 / 0.5294 & 0.8441 / 0.5334 \\
& KR & 0.9708 / 0.8745 & 0.9156 / 0.6939 & 0.9090 / 0.6739 \\
\cmidrule{1-5}
\multirow[c]{3}{*}{\shortstack{Second Residual\\[1.2pt]{\scriptsize(\textit{Single-level})}}}
& TO & 0.7253 / 0.3135 & 0.7079 / 0.2817 & 0.7015 / 0.2809 \\
& ER & 0.8221 / 0.4210 & 0.7895 / 0.3637 & 0.7938 / 0.3617 \\
& KR & 0.8747 / 0.5432 & 0.8385 / 0.4709 & 0.8512 / 0.4782 \\
\cmidrule{1-5}
& TO
& \cellcolor{skyblue!20}0.9807 / 0.9469
& \cellcolor{skyblue!20}0.8691 / 0.5911
& \cellcolor{skyblue!20}0.8808 / 0.6497 \\
& ER
& \cellcolor{skyblue!30}\underline{0.9989} / \underline{0.9952}
& \cellcolor{skyblue!30}\underline{0.9494} / \underline{0.7917}
& \cellcolor{skyblue!30}\underline{0.9301} / \underline{0.7812} \\
\multirow[c]{-3}{*}{\shortstack{SpCon\\[1pt]{\scriptsize(\textit{Evolution}})}}
& KR
& \cellcolor{skyblue!40}\textbf{0.9999} / \textbf{0.9994}
& \cellcolor{skyblue!40}\textbf{0.9699} / \textbf{0.8681}
& \cellcolor{skyblue!40}\textbf{0.9569} / \textbf{0.8438} \\
\bottomrule
\end{tabular}}
\label{One_Two}
\end{minipage}
\vspace{-0.2cm}
\end{table*}

\subsection{Robustness to Real-World Social Media}
As shown in \textcolor{refcolor}{Figure}~\ref{SpCon_Platform}, ER and KR exceed the strongest baseline~\citep{AEDR} on both metrics across all three platforms. On average, ER improves AUROC/AP by 0.034/0.193, while KR increases the gains to 0.070/0.315. Even on Facebook, where AEDR reaches 0.961 AUROC, ER attains 0.962 while raising AP from 0.734 to 0.830. These results show that SpCon more stably preserves source-discriminative evidence across the real-world platforms. Image acquisition and deployment-oriented low-FPR/fixed-threshold results are detailed in \textcolor{refcolor}{Appendices}~\ref{Real_World_Platform} and \ref{More_Metrics}.

\subsection{Computational Efficiency}
SpCon requires only two VAE reconstructions and a vectorized spectral computation, averaging 0.384\,s per image (see \textcolor{refcolor}{Table}~\ref{SpCon_Time}): a 62.9$\times$ speedup over LatentTracer (LT) and on par with AEDR. OCC-CLIP (OCC) is faster at inference but requires supervised training for each target model.

\subsection{Ablation Studies}
\label{SpCon_Ablation}
\textbf{Protocol.}
Unless stated otherwise, ablations use 100 fitting images and 100 disjoint test images per source across all seven models and 35 conditions, varying one factor at a time. Additional studies on reconstruction depth, density and fusion, Student-$t$ degrees of freedom, covariance structure, external-reference composition, and Seen/Unseen partitions are provided in \textcolor{refcolor}{Appendix}~\ref{More_Ablation}.

\textbf{Spectral-Contraction Response.}
We compare the full spectral-contraction response with single-level residual spectra (see \textcolor{refcolor}{Table}~\ref{One_Two}), using the 48-dimensional log band power of $\mathbf{r}_1$ or $\mathbf{r}_2$ alone while keeping all other components unchanged. Both variants perform substantially worse than SpCon on Seen and Unseen, showing that the gain arises from the relative spectral evolution across residual levels rather than either residual alone. Likewise, substituting the scalar AEDR statistic under the same references and density backend lowers ER from 0.9387/0.7825 to 0.7444/0.4496 over all post-processing conditions (\textcolor{refcolor}{Table}~\ref{AEDR_Mix}), demonstrating a representation gain under the same backend.

% Under the family-balanced All-13 aggregation, the tested per-channel centering recovers at most 0.0176 AUROC and 0.0358 AP of the gap, supporting the relative response beyond this single-residual normalization; details are provided in \textcolor{refcolor}{Appendix}~\ref{Repr_Controls_Sec}.

\begin{figure*}[t]
\centering
\begin{minipage}[t]{0.49\textwidth}
    \centering
    \includegraphics[width=\linewidth]{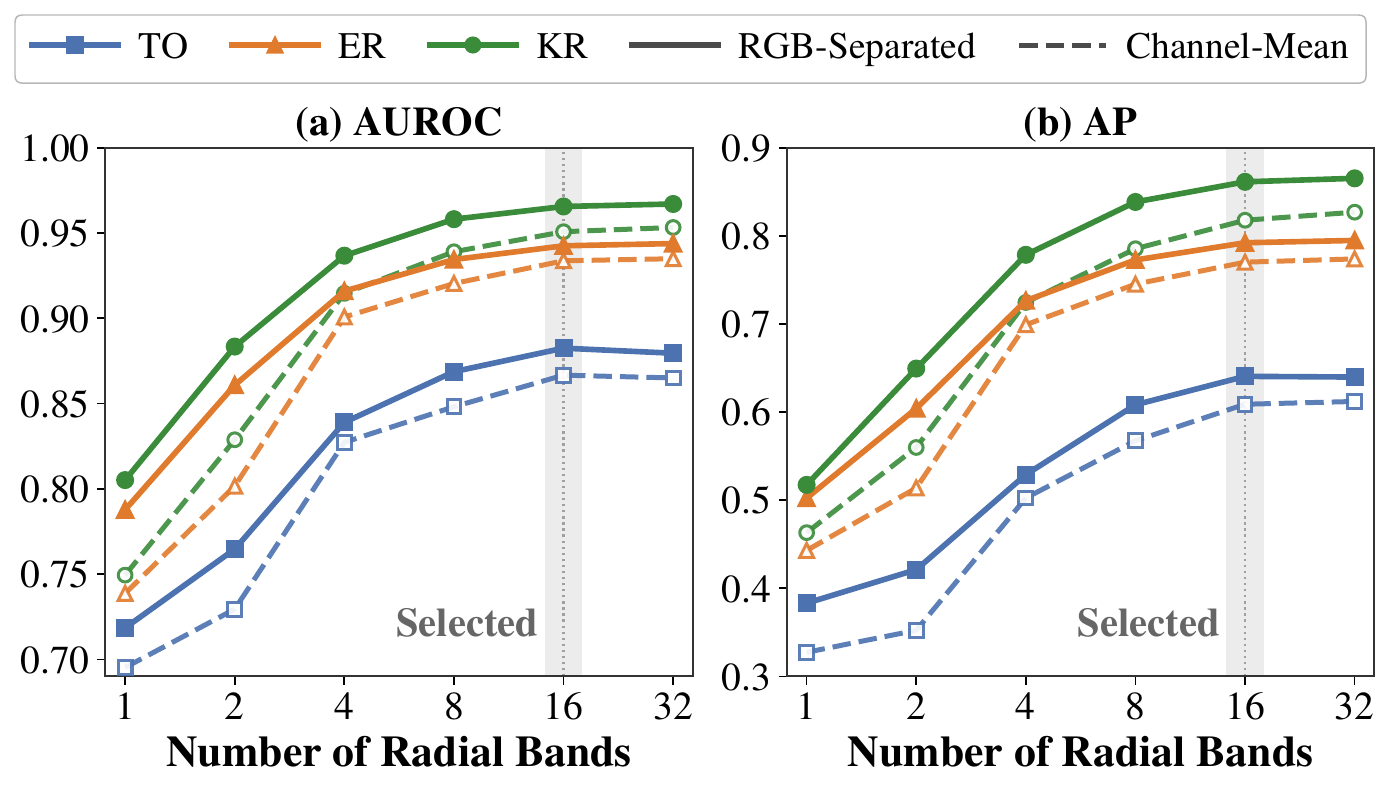}
    \vspace{-0.6cm}
    \caption{Effect of the radial bands and channels.}
    \label{Channel_Bands}
\end{minipage}
\hfill
\begin{minipage}[t]{0.49\textwidth}
    \centering
    \includegraphics[width=\linewidth]{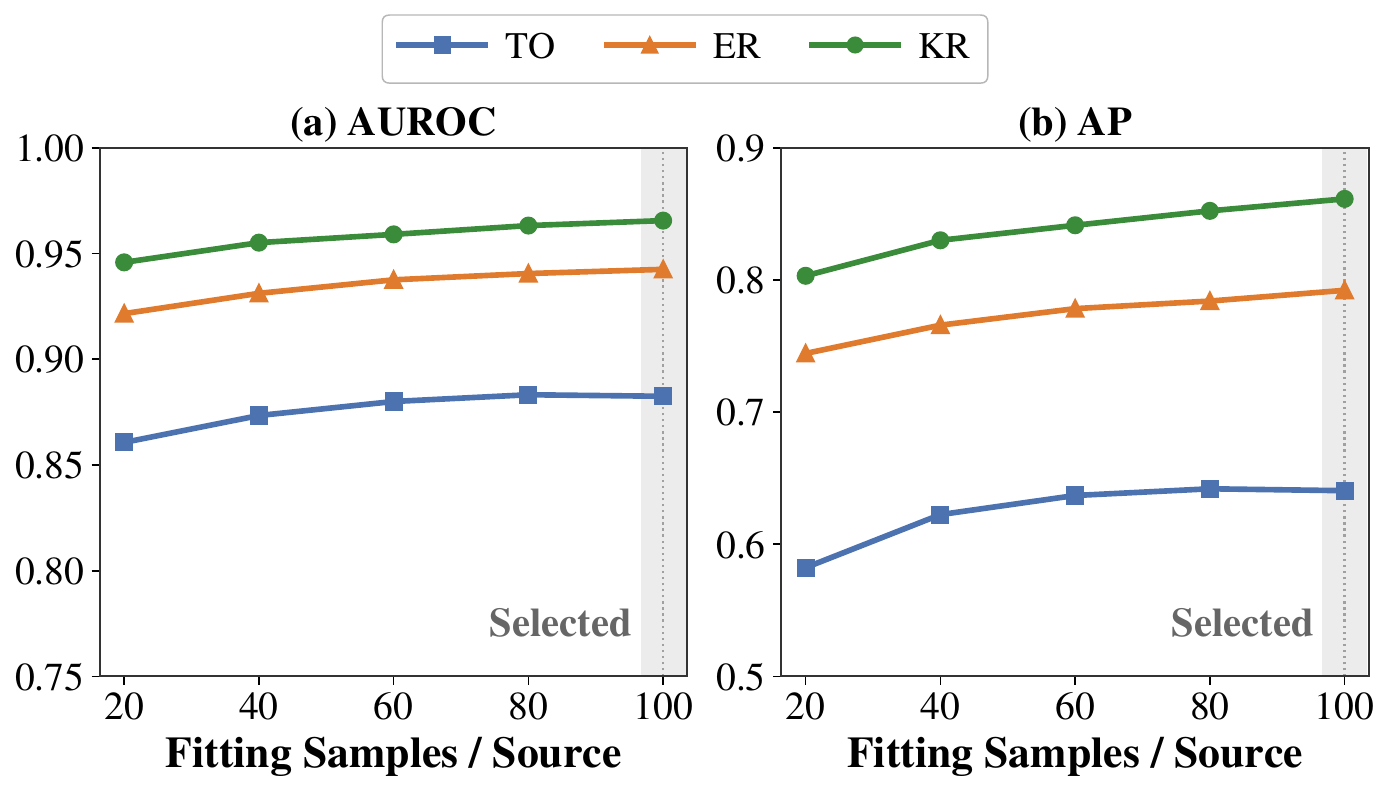}
    \vspace{-0.6cm}
    \caption{Effect of the number of fitting samples.}
    \label{Fitting_Samples}
\end{minipage}
\vspace{-0.3cm}
\end{figure*}

\textbf{Radial Bands and RGB Channels.}
Increasing radial bands from 1 to 16 improves all three settings monotonically (see \textcolor{refcolor}{Figure}~\ref{Channel_Bands}). At $B{=}1$ with summed RGB energies, the response reduces to a transform of AEDR's global error-ratio component~\citep{AEDR} (see \textcolor{refcolor}{Appendix}~\ref{Response_Properties}); SpCon therefore generalizes this component, and each finer partition adds attribution evidence under a fixed reconstruction count. Separating RGB channels further improves performance, while 32 bands yield negligible gains at twice the dimension; we therefore use 16 bands with separate channels.

\textbf{Number of Fitting Samples.}
We vary the number of fitting images per reference source (see \textcolor{refcolor}{Figure}~\ref{Fitting_Samples}). Performance improves with more samples and saturates after 60 to 80 images; even with only 20 images per source, ER and KR remain strong, showing a low reference-data requirement.

\textbf{Additional Analyses.}
Further studies cover prevalence shift, low-FPR operation, and condition-agnostic thresholds (\textcolor{refcolor}{Appendix}~\ref{More_Metrics}); unseen multi-operation chains (\ref{Chained_Postprocessing}); closed-set attribution from candidate-conditioned scores (\ref{Closed_Set_Attribution}); representation and backend controls (\ref{More_Ablation}, \ref{Repr_Controls_Sec}); and an adaptive-adversary stress test of SpCon using standard gradient-based perturbations (\ref{Adaptive_Adversary}).

%======================================================================================================
\vspace{-0.1cm}
\section{Discussion}
\label{Discussion}
\vspace{-0.1cm}

\textbf{Closed-Set Attribution.}
Although the main formulation evaluates one target at a time, the seven candidate-conditioned scores can be composed into a single source decision. SpCon-KR achieves Top-1 accuracy of 0.9983/0.8472/0.8498/0.8724 on Clean/Seen/Unseen/Platform inputs (see \textcolor{refcolor}{Appendix}~\ref{Closed_Set_Attribution}), showing that its binary evidence extends naturally to closed-set multi-source attribution.

\textbf{Limitations and Future Work.}
SpCon provides robust VAE-level attribution across diverse post-processing and platform pipelines. This granularity reflects a general limitation of attribution methods relying primarily on VAE-level evidence~\citep{RONAN, LatentTracer, AEDR}: when generators share the same or highly compatible VAEs, such evidence alone cannot reliably distinguish individual models (see \textcolor{refcolor}{Appendix}~\ref{Capability_Boundary}), and combining it with generator-level cues is a natural extension. Multi-operation chains reduce KR AUROC from 0.9838 on single operations to 0.8633 and 0.6427 on two- and three-operation chains (\textcolor{refcolor}{Appendix}~\ref{Chained_Postprocessing}), whereas adaptive perturbations built from public artifacts leave AUROC above 0.946 and AP above 0.759 at $\epsilon=8/255$ (\textcolor{refcolor}{Appendix}~\ref{Adaptive_Adversary}).
% Access to the auditor's fitted densities defines a stronger threat model presupposing auditor-side parameters (\textcolor{refcolor}{Appendix}~\ref{Adaptive_Whitebox}). 
%======================================================================================================
\vspace{-0.1cm}
\section{Conclusion}
\vspace{-0.1cm}
We studied source attribution of generated images under post-processing and proposed SpCon, built on Spectral Contraction, which recasts attribution from global reconstruction statistics to modeling how the two-step residual evolves across frequency scales under the same target VAE. Across seven models, 35 synthetic conditions, and three real platforms, SpCon is robust on clean images, Seen/Unseen post-processing, and real propagation, at a cost comparable to the cheapest reconstruction baseline. SpCon combines frequency-resolved structure with relative residual evolution, capturing source evidence beyond reconstruction accuracy or its aggregate change alone.
%======================================================================================================
\newpage

\subsection*{AI use statement}
A large language model (LLM) was used to assist with the writing and editing of this paper, specifically to polish and rephrase sentences for improved clarity, readability, and stylistic consistency. The authors have carefully reviewed all such content and take full responsibility for the manuscript.

\subsection*{Ethics statement}
This work is intended to support defensive applications, including content-provenance verification, copyright auditing, and model accountability. All experiments are conducted using publicly available models, publicly released datasets, and images generated from these models; the study involves no human-subject experiments and no collection of new personal data.

\subsection*{Reproducibility statement}
The main paper specifies the method, access assumptions, data splits, post-processing protocol, baselines, and evaluation metrics, and together with the appendix, which provides implementation details, hyperparameters, ablations, and complete results, offers a sufficient description of the method to support reproduction. The complete code will be released upon acceptance.


\bibliography{iclr2027_conference}
\bibliographystyle{iclr2027_conference}

\newpage
\appendix
\addtocontents{toc}{\protect\setcounter{tocdepth}{2}}

%--------------------------------
% Appendix table of contents
%--------------------------------
\makeatletter
\begingroup
  \hypersetup{hidelinks}
  \pdfbookmark[1]{Appendix Contents}{apptoc}
  \begin{center}
    {\LARGE\bfseries Appendix}\\[0.5em]
    {\large\bfseries Table of Contents}
  \end{center}
  \vskip 0.8em
  \hrule height 0.8pt
  \vskip 1.6em
  \renewcommand*\l@section[2]{%
    \ifnum \c@tocdepth >\z@
      \addpenalty\@secpenalty
      \addvspace{1.3em \@plus\p@}%
      \setlength\@tempdima{2.2em}%
      \begingroup
        \parindent\z@ \rightskip\@pnumwidth
        \parfillskip -\@pnumwidth
        \leavevmode \large\bfseries
        \advance\leftskip\@tempdima \hskip -\leftskip
        #1\nobreak\hfil \nobreak\hb@xt@\@pnumwidth{\hss #2}\par
      \endgroup
    \fi}%
  \renewcommand*\l@subsection{\@dottedtocline{2}{2.2em}{2.8em}}%
  \normalsize
  \@starttoc{toc}%
  \vskip 1.2em
  \hrule height 0.8pt
\endgroup
\makeatother

\newpage

%========================================================================================================
\section{Properties and Scope of the Spectral-Contraction Response}
\label{Response_Properties}

This section states what the band-resolved response of \textcolor{refcolor}{Equation}~\ref{SpCon_Z} measures, in what sense it is more stable than a scalar reconstruction statistic, and the scope within which it holds.

\textbf{Connection to AEDR's error-ratio component.}
At $B=1$, after summing the three color-channel energies, the single radial band spans the entire frequency plane, and Parseval's identity for the orthonormal DFT gives $\sum_k \mathcal{B}_1(r_{j,k})=\|\mathbf{r}_j\|_2^2$. Provided that both total residual energies exceed $\varepsilon$, the response reduces to
\begin{equation}
z=\log\|\mathbf{r}_1\|_2^{2}-\log\|\mathbf{r}_2\|_2^{2}=-2\log\frac{\|\mathbf{r}_2\|_2}{\|\mathbf{r}_1\|_2},
\end{equation}
a monotone transform of the global MSE ratio computed from the same residual pair. This establishes a connection to the error-ratio component of AEDR~\citep{AEDR}, but not equivalence to its complete score, which additionally incorporates homogeneity calibration and differs in its native reconstruction protocol. At finer resolutions, SpCon retains a vector of relative responses that cannot in general be recovered from the global ratio. As shown by \textcolor{refcolor}{Equation}~\ref{Global_Ratio_Decomposition}, the global ratio mixes these responses with input-dependent residual-energy weights. The gain from frequency resolution therefore concerns information retained before scoring, rather than a mere reparameterization of one scalar. \textcolor{refcolor}{Table}~\ref{AEDR_Mix} compares the representations after matching the reconstruction protocol, reference access, and density backend.

\textbf{Why global error ratios can drift.}
An orthonormal Fourier transform preserves total squared energy; applying it and then summing over all frequencies would therefore recover the same global reconstruction error. SpCon instead retains the relative change within each band. To expose the distinction, let $d=(k,b)$ index a channel-band coordinate and let $\mathcal{B}_{j,d}=\mathcal{B}_b(r_{j,k})$ denote the corresponding residual-band energy. For energies above the numerical floor, the global second-to-first residual-energy ratio can be written as
\begin{equation}
\begin{aligned}
R_{\mathrm{glob}}&=\frac{\|\mathbf{r}_2\|_2^2}{\|\mathbf{r}_1\|_2^2}=\frac{\sum_d \mathcal{B}_{2,d}}{\sum_d \mathcal{B}_{1,d}}=\sum_d w_d\exp(-z_d), \\w_d&=\frac{\mathcal{B}_{1,d}}{\sum_{d'}\mathcal{B}_{1,d'}}.
\end{aligned}
\label{Global_Ratio_Decomposition}
\end{equation}
Thus, the global error ratio is an energy-weighted average of the exponentiated negative band-wise responses, $\exp(-z_d)$. Its weights depend on the input residual rather than on the attribution relevance of each band. Different response profiles can produce the same global ratio, and changing the energy allocation $w_d$ can shift that ratio even when every $z_d$ remains unchanged. AEDR's MSE ratio, supplemented by its homogeneity calibration, does not retain these individual band-wise responses.

\textbf{Band-wise common-mode cancellation.}
Assuming that all original and processed energies exceed $\varepsilon$, suppose a processed input induces
\begin{equation}
\widetilde{\mathcal{B}}_{1,d}=\gamma_d \mathcal{B}_{1,d},\qquad\widetilde{\mathcal{B}}_{2,d}=\gamma_d \mathcal{B}_{2,d},\qquad \gamma_d>0,
\end{equation}
where $\gamma_d$ may vary across channels and bands. The resulting response satisfies
\begin{equation}
\widetilde{z}_d=\log(\gamma_d \mathcal{B}_{1,d})-\log(\gamma_d \mathcal{B}_{2,d})=z_d.
\end{equation}
By contrast, a scalar ratio formed after aggregation becomes
\begin{equation}
\widetilde{s}_{\mathrm{glob}}=\log\frac{\sum_d \gamma_d \mathcal{B}_{1,d}}{\sum_d \gamma_d \mathcal{B}_{2,d}},
\end{equation}
which generally differs from $s_{\mathrm{glob}}=\log\frac{\sum_d \mathcal{B}_{1,d}}{\sum_d \mathcal{B}_{2,d}}$ when $\gamma_d$ varies with $d$. If the two residual levels instead receive different factors $\gamma^{(1)}_d$ and $\gamma^{(2)}_d$, the remaining response shift is
\begin{equation}
\widetilde{z}_d-z_d=\log\gamma^{(1)}_d-\log\gamma^{(2)}_d.
\label{Bandwise_Differential_Shift}
\end{equation}
The band-resolved response therefore removes multiplicative changes shared by the two residual levels within each band and retains only their differential component, whereas frequency-dependent changes generally alter the relative energy weights of the global ratio even when every band-wise ratio is unchanged. Relative comparison and frequency resolution thus play complementary roles: the former cancels shared multiplicative energy changes within each band, while the latter preserves the band-wise response pattern for joint modeling. The distinction is not the presence or absence of normalization, but its granularity: SpCon compares corresponding bands before aggregation, whereas a global error ratio compares already-aggregated energies.

\textbf{Scope.}
The cancellation identity is a sufficient condition for response stability: when the two residual levels receive different band-wise factors, the response retains their differential component (\textcolor{refcolor}{Equation}~\ref{Bandwise_Differential_Shift}), which the reference densities model over the fitted conditions. The behavior of $\mathcal{R}$ under repeated application is measured rather than assumed; \textcolor{refcolor}{Appendix}~\ref{Multi_Step_SpCon} shows that the source-discriminative evidence is concentrated in the first transition.

\textbf{What the source cue looks like.}
\textcolor{refcolor}{Figure}~\ref{SpCon_Phenomenon} illustrates this distinction under a fixed SDXL target VAE. The non-belonging SD1 images exhibit pronounced low-to-mid-frequency contraction together with weak or negative high-frequency responses, whereas the belonging SDXL images follow a different, more uniform profile. JPEG-H and Flip change the response magnitudes without eliminating this qualitative source contrast.

%========================================================================================================

\section{Details of the Post-Processing Protocol}

\subsection{Models and Data}
All generators use the same ordered prompt set within each split: images with the same index across generators are conditioned on the same prompt. The generated belonging and non-belonging pools are therefore matched in prompt identity and frequency, preventing source comparisons from being confounded by mismatched prompt distributions. This controls textual conditioning and prompt frequency but does not assume pixel- or style-matched outputs across generators.

\subsection{Post-Processing Operations}
\label{Appendix_Post_Processing}

\begin{figure}[t]
\centering
\includegraphics[width=1\columnwidth]{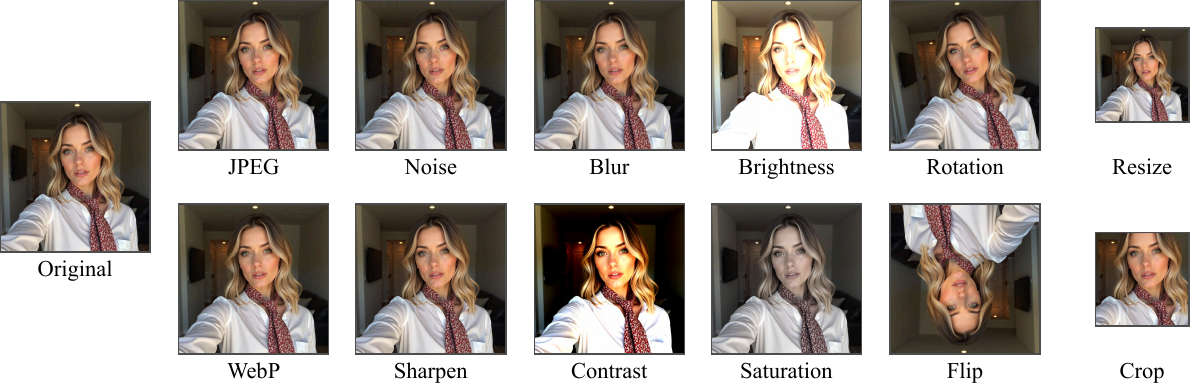}
\caption{The twelve post-processing operations applied to an example image.}
\vspace{-0.3cm}
\label{12_Operations}
\end{figure}

For each non-Clean condition in the main synthetic benchmark, we apply exactly one post-processing operation to each test image (see \textcolor{refcolor}{Figure}~\ref{12_Operations}). All methods receive the identical processed image. Only \emph{Clean} and the medium severities of the six \emph{Seen} operations are used to fit SpCon; the low and high \emph{Seen} severities and all \emph{Unseen} operations are used exclusively for evaluation. Any subsequent input-size canonicalization required by an individual method is method-specific preprocessing and is not counted as a post-processing operation.

\subsubsection{Seen Operations}
\textbf{(1) JPEG.} Each RGB image is encoded into an in-memory JPEG byte stream using Pillow and then decoded back to RGB. The three severity levels use quality factors of $\mathrm{QF}\in\{90,75,60\}$, with fixed 4:2:0 chroma subsampling. A lower quality factor indicates stronger compression.

\textbf{(2) Noise.} In the 8-bit RGB domain $[0,255]$, zero-mean Gaussian noise $\boldsymbol{\eta}$ is added, with each entry independently drawn from $\mathcal{N}(0,\spcSigN^{2})$ and $\spcSigN\in\{5,10,15\}$. The processed image is $\mathbf{x}_{\mathrm{out}}=\operatorname{clip}(\mathbf{x}+\boldsymbol{\eta},0,255)$.

\textbf{(3) Blur.} The three severity levels use $\spcKer\times\spcKer$ Gaussian kernels with $\spcKer\in\{3,5,9\}$. For each base image, the Gaussian standard deviation $\spcSigG$ is deterministically sampled from $\mathcal{U}(0.1,2.0)$ using the image-content hash. The same value of $\spcSigG$ is shared across the three kernel sizes for that image. Thus, the three severity levels differ primarily in the spatial support of the Gaussian kernel.

\textbf{(4) Brightness.} Global brightness is performed using Pillow's \texttt{ImageEnhance.Brightness}, with multiplicative enhancement factors of $1.2$, $1.6$, and $2.0$. A larger factor produces a brighter image. This operation changes neither the image dimensions nor the spatial locations of pixels.

\textbf{(5) Rotation.} Each image is rotated around its center by $2^\circ$, $5^\circ$, or $8^\circ$, while the output canvas size remains unchanged. The rotation direction is determined by the global index of the base image: even indices use positive angles, whereas odd indices use negative angles. The three experimental conditions are therefore denoted by $\theta\in\{\pm2^\circ,\pm5^\circ,\pm8^\circ\}$. Bilinear interpolation and reflective boundary handling are used to avoid introducing fixed black borders.

\textbf{(6) Resize.} Both the image width and height are reduced using scale factors $s\in\{7/8,6/8,5/8\}$. The resulting dimensions are then independently aligned to the nearest multiples of 8. Lanczos resampling is used, and the resized image is not restored to its original dimensions.

\subsubsection{Unseen Operations}
\textbf{(1) WebP.} Each RGB image is encoded into an in-memory WebP byte stream using Pillow and then decoded back to RGB. The three severity levels use quality factors of $\mathrm{QF}\in\{90,75,60\}$, with the encoder search parameter fixed to \texttt{method=6}. The \texttt{method} parameter controls the encoder search quality and computational cost rather than the distortion severity.

\textbf{(2) Sharpen.} Sharpening is performed using unsharp masking. A blurred image is first generated using a fixed $3\times3$ Gaussian kernel with $\spcSigG=1$ and reflective boundary handling. The high-frequency residual is then added back to the original image with strength $\alpha$. The three severity levels use $\alpha\in\{0.5,1.0,1.5\}$. A larger value of $\alpha$ produces stronger enhancement of edges.

\textbf{(3) Contrast.} Global contrast enhancement is performed using \texttt{ImageEnhance.Contrast}, with enhancement factors of $1.2$, $1.6$, and $2.0$. A larger factor increases the difference between bright and dark regions. This operation changes neither the image dimensions nor the spatial structure.

\textbf{(4) Saturation.} Color saturation is adjusted using Pillow's \texttt{ImageEnhance.Color}, with factors of $0.8$, $0.6$, and $0.4$. A lower factor moves the image closer to grayscale; therefore, $0.4$ corresponds to the highest severity. This operation changes neither the image dimensions nor the spatial structure.

\textbf{(5) Flip.} Flip is treated as a single Unseen operation without severity levels. For each base image, its content hash is used to deterministically select, with equal probability, horizontal flipping, vertical flipping, or their composition. The three variants jointly constitute a single Flip condition rather than three separate experiments. Flipping only rearranges pixel locations; it introduces no interpolation and changes neither the pixel values nor the image dimensions.

\textbf{(6) Crop.} The image is center-cropped using retained side-length ratios $s\in\{7/8,6/8,5/8\}$, producing a central region of approximately $sW\times sH$. The cropped region is then resized back to the pre-operation dimensions $W\times H$ using Lanczos interpolation.

\subsection{Real-World Platform Processing}
\label{Real_World_Platform}
We evaluate three platform-processing pipelines using Facebook, Telegram, and WeChat. For Facebook, images generated by each model were uploaded to a separate album named after the corresponding source model. After all uploads were completed, the processed images were retrieved collectively through Facebook's \emph{Download Your Information} function. For Telegram, we created a dedicated channel and uploaded the same image set used for Facebook, with images from each model uploaded as a separate batch and subsequently downloaded in bulk. For WeChat, images were sent through the File Transfer Assistant by direct copy-and-paste, rather than through the \emph{File} or \emph{Original} transmission modes, and were then downloaded in batches. All three pipelines used the platforms' default upload, processing, and download settings; no quality, compression, resolution, or transmission options were manually modified. The downloaded images were used only for evaluation and did not participate in reference fitting.

%======================================================================================================
\vspace{-0.1cm}
\section{Baseline Implementation and Controlled Variants}
\label{Appendix_Baseline}
We implement OCC-CLIP~\citep{OCC-CLIP}, LatentTracer~\citep{LatentTracer}, and AEDR~\citep{AEDR} following their official configurations. In addition to the standard baselines used in the main paper, we construct controlled variants of OCC-CLIP and AEDR that progressively match SpCon in input protocol, sample budget, or post-processing prior while preserving their core algorithms and scoring mechanisms. These controls examine whether the gains of SpCon can be explained by simple adaptations of existing methods rather than by the proposed signal itself. 

We additionally clarify the omission of several cited methods. DNA~\citep{DNA} extends AEDR to a broader attribution pipeline; under our one-target binary verification protocol, its applicable reconstruction-based attribution component reduces to AEDR and therefore does not provide an independent baseline. Forensic Self-Descriptions~\citep{Forensic} does not publicly provide the training pipeline required to reproduce its open-set attribution setting. DE-FAKE~\citep{DE-FAKE}, while reproducible, relies on a substantially heavier supervised regime (80,000 training images in its attribution setup), which is fundamentally different from our lightweight reference-based setting. We therefore report these methods as context rather than as baselines under this protocol.

\vspace{-0.1cm}
\subsection{OCC-CLIP}
\textbf{Native.} The main-paper OCC-CLIP~\citep{OCC-CLIP} follows the official 50-shot training recipe. For each target generator, we train on 50 clean belonging images and 50 clean COCO images. This configuration is denoted simply as OCC-CLIP in the main paper.

\textbf{Matched.} OCC-CLIP-Matched keeps the architecture, optimization procedure, and official fast gradient sign method (FGSM) augmentation of the native configuration unchanged, while increasing the number of clean base images from 50 to 100 per source. This matches the number of unique base images available per reference source to SpCon and controls for differences in sample budget.

\textbf{Matched-Aug.} OCC-CLIP-Matched-Aug further provides the same post-processing prior used during SpCon fitting while retaining the same 100 base images per source. Each base image contributes the \textit{Clean} view and six medium-severity \textit{Seen} views: JPEG, Noise, Blur, Brightness, Rotation, and Resize. This results in 700 training views per source while keeping the number of unique base images fixed at 100. The additional post-processing views therefore match the Seen-operation prior available to SpCon without modifying OCC-CLIP's classification signal. This provides a stronger supervised control for evaluating how much benefit comes from the same post-processing prior.

As shown by the performance comparison in \textcolor{refcolor}{Table}~\ref{OCC_variant} and the significance analysis in \textcolor{refcolor}{Table}~\ref{OCC_Boost}, even after OCC-CLIP-Matched and OCC-CLIP-Matched-Aug progressively match SpCon in base-image budget and Seen post-processing prior, a substantial and statistically significant gap to SpCon remains. Compared with the strongest Matched-Aug baseline, SpCon-ER achieves absolute gains of 0.2358 and 0.5242 in AUROC and AP on All Post., while the gains remain 0.2439 and 0.5030 on real-world Platforms. Since Matched-Aug receives the same base-image budget and Seen post-processing prior as SpCon, the remaining gap cannot be explained by these factors alone, supporting the contribution of SpCon's attribution signal rather than additional samples or augmentation.

\begin{table}[t]
\centering
\vspace{-0.2cm}
\caption{Comparison with OCC-CLIP~\citep{OCC-CLIP} and its variants. \dagg~marks a significant improvement over the strongest variant in this table under the two-sided 95\% CI of \textcolor{refcolor}{Table}~\ref{OCC_Boost}.}
\vspace{0.1cm}
\scriptsize
\setlength{\tabcolsep}{4pt}
\renewcommand{\arraystretch}{1.2}
\resizebox{\columnwidth}{!}{
\begin{tabular}{l|ccccc}
\toprule
\textbf{Method} & \textbf{Clean} & \textbf{Seen} & \textbf{Unseen} & \textbf{All Post.} & \textbf{Platforms} \\
\midrule
Native      & 0.6679 / 0.2137 & 0.6572 / 0.2080 & 0.6671 / 0.2124 & 0.6618 / 0.2101 & 0.6585 / 0.2161 \\
Matched     & 0.6781 / 0.2171 & 0.6669 / 0.2113 & 0.6773 / 0.2167 & 0.6718 / 0.2139 & 0.6698 / 0.2167 \\
Matched-Aug & 0.7071 / 0.2631 & 0.7007 / 0.2580 & 0.7054 / 0.2587 & 0.7029 / 0.2583 & 0.6930 / 0.2519 \\
\rowcolor{skyblue!20}
SpCon-TO    & 0.9818\dagg / 0.9509\dagg & 0.8712\dagg / 0.6017\dagg & 0.8805\dagg / 0.6516\dagg & 0.8756\dagg / 0.6251\dagg & 0.8581\dagg / 0.5507\dagg \\
\rowcolor{skyblue!30}
SpCon-ER    & \underline{0.9994}\dagg / \underline{0.9969}\dagg & \underline{0.9470}\dagg / \underline{0.7874}\dagg & \underline{0.9294}\dagg / \underline{0.7770}\dagg & \underline{0.9387}\dagg / \underline{0.7825}\dagg & \underline{0.9369}\dagg / \underline{0.7549}\dagg \\
\rowcolor{skyblue!40}
SpCon-KR    & \textbf{1.0000}\dagg / \textbf{0.9997}\dagg & \textbf{0.9708}\dagg / \textbf{0.8732}\dagg & \textbf{0.9569}\dagg / \textbf{0.8421}\dagg & \textbf{0.9642}\dagg / \textbf{0.8586}\dagg & \textbf{0.9734}\dagg / \textbf{0.8773}\dagg \\
\bottomrule
\end{tabular}}
\label{OCC_variant}
\vspace{-0.4cm}
\end{table}

\begin{table}[t]
\centering
\caption{Bootstrap significance analysis for the comparison with OCC-CLIP and its variants. We report the performance difference $\Delta$ relative to the specified reference, together with the 95\% CI.}
\vspace{0.1cm}
\scriptsize
\setlength{\tabcolsep}{2.5pt}
\renewcommand{\arraystretch}{1.2}
\resizebox{\columnwidth}{!}{
\begin{tabular}{llc|ccc}
\toprule
\textbf{Scope} & \textbf{Metric} & \textbf{Reference} & \textbf{TO: $\Delta$ / 95\% CI} & \textbf{ER: $\Delta$ / 95\% CI} & \textbf{KR: $\Delta$ / 95\% CI} \\
\midrule
\multirow{2}{*}{Clean}
& AUROC & Matched-Aug & +0.2747 / [0.2690, 0.2802] & +0.2923 / [0.2875, 0.2969] & +0.2929 / [0.2880, 0.2975] \\
& AP & Matched-Aug & +0.6878 / [0.6765, 0.6948] & +0.7339 / [0.7243, 0.7389] & +0.7366 / [0.7271, 0.7418] \\
\cmidrule{1-6}
\multirow{2}{*}{Seen}
& AUROC & Matched-Aug & +0.1705 / [0.1646, 0.1760] & +0.2463 / [0.2415, 0.2509] & +0.2700 / [0.2655, 0.2746] \\
& AP & Matched-Aug & +0.3436 / [0.3331, 0.3516] & +0.5294 / [0.5197, 0.5360] & +0.6152 / [0.6054, 0.6215] \\
\cmidrule{1-6}
\multirow{2}{*}{Unseen}
& AUROC & Matched-Aug & +0.1752 / [0.1697, 0.1807] & +0.2240 / [0.2194, 0.2284] & +0.2515 / [0.2470, 0.2561] \\
& AP & Matched-Aug & +0.3929 / [0.3821, 0.4005] & +0.5183 / [0.5096, 0.5244] & +0.5834 / [0.5745, 0.5891] \\
\cmidrule{1-6}
\multirow{2}{*}{All Post.}
& AUROC & Matched-Aug & +0.1727 / [0.1670, 0.1782] & +0.2358 / [0.2313, 0.2403] & +0.2613 / [0.2568, 0.2658] \\
& AP & Matched-Aug & +0.3668 / [0.3564, 0.3743] & +0.5242 / [0.5153, 0.5302] & +0.6002 / [0.5910, 0.6059] \\
\cmidrule{1-6}
\multirow{2}{*}{Platforms}
& AUROC & Matched-Aug & +0.1651 / [0.1559, 0.1739] & +0.2439 / [0.2368, 0.2504] & +0.2804 / [0.2744, 0.2860] \\
& AP & Matched-Aug & +0.2988 / [0.2774, 0.3166] & +0.5030 / [0.4842, 0.5137] & +0.6255 / [0.6058, 0.6344] \\
\bottomrule
\end{tabular}}
\label{OCC_Boost}
\vspace{-0.2cm}
\end{table}

\vspace{-0.1cm}
\subsection{LatentTracer}
\textbf{Native.} LatentTracer~\citep{LatentTracer} performs attribution through per-image VAE inversion. For each test image and target generator, the latent variable is initialized from the target VAE encoder and independently optimized for 100 iterations using Adam~\citep{Adam} to reconstruct the input image. We use the negative minimum reconstruction error obtained during optimization as the attribution score, such that a larger score indicates stronger evidence of belonging to the target generator. The released implementation natively supports SD1.5 and SD2.1-base in our benchmark. For the remaining generators, we preserve the same VAE-inversion objective and 100-step optimization procedure while making only the necessary architectural adaptations; we therefore report them under the same LatentTracer entry.

\begin{table}[t]
\centering
\caption{Comparison with AEDR~\citep{AEDR} and its protocol-matched variant. \dagg~marks a significant improvement over the strongest variant in this table under the 95\% CI of \textcolor{refcolor}{Table}~\ref{AEDR_Boost}.}
\vspace{0.1cm}
\scriptsize
\setlength{\tabcolsep}{4pt}
\renewcommand{\arraystretch}{1.2}
\resizebox{\textwidth}{!}{
\begin{tabular}{l|ccccc}
\toprule
\textbf{Method} & \textbf{Clean} & \textbf{Seen} & \textbf{Unseen} & \textbf{All Post.} & \textbf{Platforms} \\
\midrule
Native   & 0.9974 / 0.9741 & 0.7636 / 0.4104 & 0.8623 / 0.6317 & 0.8100 / 0.5146 & 0.9031 / 0.5620 \\
Matched  & 0.9641 / 0.8002 & 0.7425 / 0.4500 & 0.8008 / 0.5653 & 0.7699 / 0.5043 & 0.7910 / 0.5216 \\
\rowcolor{skyblue!20}
SpCon-TO & 0.9818 / 0.9509 & 0.8712\dagg / 0.6017\dagg & 0.8805\dagg / 0.6516\dagg & 0.8756\dagg / 0.6251\dagg & 0.8581 / 0.5507 \\
\rowcolor{skyblue!30}
SpCon-ER & \underline{0.9994}\dagg / \underline{0.9969}\dagg & \underline{0.9470}\dagg / \underline{0.7874}\dagg & \underline{0.9294}\dagg / \underline{0.7770}\dagg & \underline{0.9387}\dagg / \underline{0.7825}\dagg & \underline{0.9369}\dagg / \underline{0.7549}\dagg \\
\rowcolor{skyblue!40}
SpCon-KR & \textbf{1.0000}\dagg / \textbf{0.9997}\dagg & \textbf{0.9708}\dagg / \textbf{0.8732}\dagg & \textbf{0.9569}\dagg / \textbf{0.8421}\dagg & \textbf{0.9642}\dagg / \textbf{0.8586}\dagg & \textbf{0.9734}\dagg / \textbf{0.8773}\dagg \\
\bottomrule
\end{tabular}}
\label{AEDR_variant}
\vspace{-0.2cm}
\end{table}

\subsection{AEDR}
\textbf{Native.} For the main AEDR~\citep{AEDR}, we reproduce the native implementation. AEDR performs two consecutive stochastic reconstructions using the target VAE~\citep{VAE} and combines the ratio of the two reconstruction MSEs with the gray-level co-occurrence matrix (GLCM) homogeneity~\citep{Hom} of the input image. We use its negative logarithm only to unify the score direction across methods; this monotonic transformation does not affect AUROC or AP. This configuration is denoted simply as AEDR in the main paper.

\textbf{Matched.} This variant preserves AEDR's two-reconstruction pipeline, MSE ratio, GLCM homogeneity, and final scalar statistic, while resizing the input to the native square resolution of the target VAE, replacing posterior sampling with deterministic posterior modes, and computing GLCM homogeneity on the same native-resolution input. Thus, AEDR-Matched changes how the inputs to the native AEDR statistic are constructed rather than redesigning the attribution signal itself.

As shown by the performance comparison in \textcolor{refcolor}{Table}~\ref{AEDR_variant} and the significance analysis in \textcolor{refcolor}{Table}~\ref{AEDR_Boost}, AEDR-Matched aligns the input resolution, VAE projection, and GLCM homogeneity computation protocol while preserving the core AEDR statistic, yet these adaptations do not systematically improve performance or close the gap to SpCon. In contrast, SpCon-ER achieves 0.9387 AUROC and 0.7825 AP on All Post., corresponding to absolute gains of 0.1287 and 0.2679 over the strongest AEDR baseline, respectively; on real-world Platforms, the absolute gains remain 0.0338 and 0.1928. These results indicate that the advantage of SpCon cannot be explained by simple modifications to AEDR's input or projection protocol, but instead arises from the design of its attribution signal.

\begin{table}[t]
\centering
\caption{Bootstrap significance analysis for the comparison with AEDR and its variant. We report the performance difference $\Delta$ relative to the specified reference, together with the 95\% CI.}
\vspace{0.1cm}
\scriptsize
\setlength{\tabcolsep}{2.5pt}
\renewcommand{\arraystretch}{1.2}
\resizebox{\columnwidth}{!}{
\begin{tabular}{llc|ccc}
\toprule
\textbf{Scope} & \textbf{Metric} & \textbf{Reference} & \textbf{TO: $\Delta$ / 95\% CI} & \textbf{ER: $\Delta$ / 95\% CI} & \textbf{KR: $\Delta$ / 95\% CI} \\
\midrule
\multirow{2}{*}{Clean}
& AUROC & Native & -0.0156 / [-0.0181, -0.0131] & +0.0020 / [0.0014, 0.0028] & +0.0026 / [0.0019, 0.0033] \\
& AP & Native & -0.0232 / [-0.0320, -0.0145] & +0.0229 / [0.0152, 0.0307] & +0.0256 / [0.0180, 0.0334] \\
\cmidrule{1-6}
\multirow{2}{*}{Seen}
& AUROC & Native & +0.1076 / [0.1035, 0.1118] & +0.1834 / [0.1804, 0.1864] & +0.2071 / [0.2043, 0.2101] \\
& AP & Matched & +0.1516 / [0.1430, 0.1586] & +0.3374 / [0.3304, 0.3430] & +0.4232 / [0.4163, 0.4282] \\
\cmidrule{1-6}
\multirow{2}{*}{Unseen}
& AUROC & Native & +0.0183 / [0.0147, 0.0219] & +0.0671 / [0.0650, 0.0691] & +0.0946 / [0.0925, 0.0967] \\
& AP & Native & +0.0198 / [0.0091, 0.0298] & +0.1453 / [0.1362, 0.1538] & +0.2104 / [0.2010, 0.2191] \\
\cmidrule{1-6}
\multirow{2}{*}{All Post.}
& AUROC & Native & +0.0656 / [0.0619, 0.0692] & +0.1287 / [0.1263, 0.1309] & +0.1542 / [0.1518, 0.1565] \\
& AP & Native & +0.1106 / [0.1017, 0.1186] & +0.2679 / [0.2598, 0.2742] & +0.3440 / [0.3360, 0.3503] \\
\cmidrule{1-6}
\multirow{2}{*}{Platforms}
& AUROC & Native & -0.0450 / [-0.0530, -0.0366] & +0.0338 / [0.0279, 0.0396] & +0.0702 / [0.0653, 0.0753] \\
& AP & Native & -0.0114 / [-0.0381, 0.0106] & +0.1928 / [0.1677, 0.2078] & +0.3153 / [0.2904, 0.3278] \\
\bottomrule
\end{tabular}}
\label{AEDR_Boost}
\vspace{-0.2cm}
\end{table}

\begin{table}[t]
\centering
\vspace{-0.2cm}
\caption{Representation-level control based on the AEDR ratio under the same final evaluation protocol as the main results. \dagg~marks a significant improvement over the AEDR-Mix variant under the same access setting, using the stratified paired bootstrap and two-sided 95\% CI of \textcolor{refcolor}{Section}~\ref{Experimental_Setup}.}
\vspace{0.1cm}
\scriptsize
\setlength{\tabcolsep}{4pt}
\renewcommand{\arraystretch}{1.1}
\resizebox{\textwidth}{!}{
\begin{tabular}{l|ccccc}
\toprule
\textbf{Method} & \textbf{Clean} & \textbf{Seen} & \textbf{Unseen} & \textbf{All Post.} & \textbf{Platforms} \\
\midrule
AEDR-Mix-TO & 0.8633 / 0.5361 & 0.7334 / 0.3994 & 0.7677 / 0.4509 & 0.7495 / 0.4236 & 0.7971 / 0.4792 \\
\rowcolor{skyblue!20}
SpCon-TO & 0.9818\dagg / 0.9509\dagg & 0.8712\dagg / 0.6017\dagg & 0.8805\dagg / 0.6516\dagg & 0.8756\dagg / 0.6251\dagg & 0.8581\dagg / 0.5507\dagg \\
AEDR-Mix-ER & 0.8091 / 0.6200 & 0.7431 / 0.4323 & 0.7459 / 0.4691 & 0.7444 / 0.4496 & 0.8043 / 0.5354 \\
\rowcolor{skyblue!30}
SpCon-ER & \underline{0.9994}\dagg / \underline{0.9969}\dagg & \underline{0.9470}\dagg / \underline{0.7874}\dagg & \underline{0.9294}\dagg / \underline{0.7770}\dagg & \underline{0.9387}\dagg / \underline{0.7825}\dagg &
\underline{0.9369}\dagg / \underline{0.7549}\dagg \\
AEDR-Mix-KR & 0.9272 / 0.7551 & 0.7428 / 0.4370 & 0.7748 / 0.5123 & 0.7579 / 0.4724 & 0.8136 / 0.5362 \\
\rowcolor{skyblue!40}
SpCon-KR & \textbf{1.0000}\dagg / \textbf{0.9997}\dagg & \textbf{0.9708}\dagg / \textbf{0.8732}\dagg & \textbf{0.9569}\dagg / \textbf{0.8421}\dagg & \textbf{0.9642}\dagg / \textbf{0.8586}\dagg & \textbf{0.9734}\dagg / \textbf{0.8773}\dagg \\
\bottomrule
\end{tabular}}
\label{AEDR_Mix}
\vspace{-0.4cm}
\end{table}

\textbf{Mix.}
AEDR-Mix is an access- and backend-matched representation control. It uses the same 100 base images per reference source, Clean and six medium-severity fitted views, positive and negative reference identities, Student-$t$ estimation, and equal-weight component aggregation as the corresponding SpCon setting. It preserves AEDR-Matched's two-reconstruction pipeline, GLCM homogeneity, and scalar ratio; only the response representation differs, with the one-dimensional AEDR statistic replacing SpCon's 48-dimensional spectral response.

As shown in \textcolor{refcolor}{Table}~\ref{AEDR_Mix}, matching the probabilistic backend does not close the gap to SpCon. On All Post., AEDR-Mix-ER achieves 0.7444 AUROC and 0.4496 AP, whereas SpCon-ER reaches 0.9387 and 0.7825, corresponding to absolute gains of 0.1943 and 0.3329, respectively. Under KR, SpCon improves from 0.7579/0.4724 to 0.9642/0.8586, yielding gains of 0.2063 and 0.3862. The advantage also persists on real-world Platforms, where SpCon-ER improves over AEDR-Mix-ER by 0.1326 AUROC and 0.2195 AP. Within this access-matched control, the remaining gap is not explained by reference availability, fitted-condition exposure, or the probabilistic backend alone, supporting the contribution of the frequency-resolved response representation.
%======================================================================================================

\section{More Metric Evaluations}
\label{More_Metrics}

\subsection{Average Precision under Varying Attribution Prevalence}
\label{AP_Prior}

\begin{table}[t]
\centering
\caption{Prior-adjusted AP under different belonging-image prevalences $\pi$. The original scores and target-condition cells are fixed; only the class prior used to evaluate precision is changed. $
\pi=1/8$ reproduces the native one-positive-source versus seven-negative-source protocol.}
\label{AP_Prior_Table}
\vspace{0.1cm}
\scriptsize
\setlength{\tabcolsep}{5pt}
\renewcommand{\arraystretch}{1.05}
\resizebox{0.8\columnwidth}{!}{
\begin{tabular}{l|ccccc}
\toprule
\textbf{Method} &
$\boldsymbol{\pi=1/4}$ &
$\boldsymbol{\pi=1/8}$ &
$\boldsymbol{\pi=1/32}$ &
$\boldsymbol{\pi=1/64}$ &
$\boldsymbol{\pi=1/128}$ \\
\midrule
\multicolumn{6}{c}{\textit{All Post-Processing}} \\
\midrule
OCC-CLIP     & 0.3724 & 0.2101 & 0.0606 & 0.0320 & 0.0171 \\
LatentTracer & 0.3895 & 0.2409 & 0.0930 & 0.0587 & 0.0374 \\
AEDR         & 0.6354 & 0.5146 & 0.3302 & 0.2569 & 0.1940 \\
\rowcolor{skyblue!20}
SpCon-TO     & 0.7420 & 0.6251 & 0.4354 & 0.3643 & 0.3076 \\
\rowcolor{skyblue!30}
SpCon-ER     & \underline{0.8572} & \underline{0.7825}
           & \underline{0.6396} & \underline{0.5763}
           & \underline{0.5202} \\
\rowcolor{skyblue!40}
SpCon-KR     & \textbf{0.9119} & \textbf{0.8586}
           & \textbf{0.7448} & \textbf{0.6893}
           & \textbf{0.6373} \\
\midrule
\multicolumn{6}{c}{\textit{Real Platforms}} \\
\midrule
OCC-CLIP     & 0.3776 & 0.2161 & 0.0667 & 0.0379 & 0.0228 \\
LatentTracer & 0.3692 & 0.2138 & 0.0656 & 0.0356 & 0.0195 \\
AEDR         & 0.7119 & 0.5620 & 0.3088 & 0.2158 & 0.1462 \\
\rowcolor{skyblue!20}
SpCon-TO     & 0.6977 & 0.5507 & 0.3176 & 0.2366 & 0.1774 \\
\rowcolor{skyblue!30}
SpCon-ER     & \underline{0.8415} & \underline{0.7549}
           & \underline{0.5877} & \underline{0.5144}
           & \underline{0.4536} \\
\rowcolor{skyblue!40}
SpCon-KR     & \textbf{0.9294} & \textbf{0.8773}
           & \textbf{0.7542} & \textbf{0.6937}
           & \textbf{0.6394} \\
\bottomrule
\end{tabular}}
\vspace{-0.4cm}
\end{table}

As shown in \textcolor{refcolor}{Table}~\ref{AP_Prior_Table}, we evaluate AP under deployment-motivated belonging-image prevalences from $1/4$ to $1/128$ without resampling or changing any prediction score. Within each target-condition cell, positive and negative samples are assigned total weights $\pi$ and $1-\pi$, respectively, before AP is computed, and the results are macro-averaged over seven targets and either 34 post-processing conditions or three real platforms; consequently, the no-skill AP equals $\pi$. Although AP decreases as positives become rarer, SpCon-KR remains best at every prior in both evaluation scopes. Even at $\pi=1/128$, it retains APs of 0.6373 on all post-processing conditions and 0.6394 on real platforms, compared with 0.1940 and 0.1462 for the strongest baseline, AEDR; SpCon-ER also retains 0.5202 and 0.4536. These results show that the advantage of SpCon is not an artifact of the native $1{:}7$ test composition and persists in substantially more imbalanced attribution scenarios.

\subsection{Low-False-Positive Operating Performance}
We additionally report TPR@1\%FPR (see \textcolor{refcolor}{Table}~\ref{TPR1}) and TPR@5\%FPR (see \textcolor{refcolor}{Table}~\ref{TPR2}) to characterize attribution in low-false-positive operating regions. Following the condition-level robustness protocol used for AUROC and AP, we construct an empirical ROC independently for every target-condition cell and retain the maximum TPR whose empirical FPR does not exceed the specified budget. Each image-domain cell contains 500 belonging and 3,500 non-belonging samples, while each platform cell contains 100 and 700 samples, respectively. We then macro-average the resulting target-condition values within Clean, Seen, Unseen, all 34 post-processing conditions (All Post.), and the three real platforms (Plats.). OCC-CLIP~\citep{OCC-CLIP} is evaluated independently for each of its five official training seeds before averaging. These ROC operating points are test-set evaluation statistics rather than fitted method parameters or deployment thresholds.

SpCon-KR achieves the strongest result in every evaluation scope at both operating points. Relative to the strongest official baseline in Clean, Seen, Unseen, All Post., and Plats., its absolute TPR gains are 0.0526, 0.5098, 0.2907, 0.4066, and 0.4738 at 1\% FPR and 0.0029, 0.5136, 0.1953, 0.3639, and 0.3629 at 5\% FPR, respectively. SpCon-ER also consistently improves over the strongest baseline at 1\% FPR and matches or exceeds it throughout at 5\% FPR. SpCon-TO improves robustness without negative references, particularly on Seen post-processing, and the reference-assisted variants further raise clean and strict low-FPR performance.

\begin{table*}[t]
\centering
\begin{minipage}[t]{0.49\textwidth}
\centering
\caption{TPR@1\%FPR under the complete 35-condition protocol. Values are macro-averaged over target-condition ROC operating points.}
\vspace{0.1cm}
\scriptsize
\setlength{\tabcolsep}{2pt}
\renewcommand{\arraystretch}{1.1}
\resizebox{\linewidth}{!}{
\begin{tabular}{l|ccccc}
\toprule
\textbf{Method} & \textbf{Clean} & \textbf{Seen} & \textbf{Unseen} & \textbf{All Post.} & \textbf{Plats.} \\
\midrule
OCC-CLIP     & 0.0318 & 0.0288 & 0.0303 & 0.0295 & 0.0330 \\
LatentTracer & 0.3003 & 0.0376 & 0.0757 & 0.0556 & 0.0233 \\
AEDR         & 0.9460 & 0.1806 & 0.4253 & 0.2958 & 0.2367 \\
\rowcolor{skyblue!20}
SpCon-TO     & 0.8803 & 0.3268 & 0.4256 & 0.3733 & 0.2576 \\
\rowcolor{skyblue!30}
SpCon-ER     & \underline{0.9889} & \underline{0.5646} & \underline{0.6228} & \underline{0.5920} & \underline{0.5176} \\
\rowcolor{skyblue!40}
SpCon-KR     & \textbf{0.9986} & \textbf{0.6904} & \textbf{0.7160} & \textbf{0.7024} & \textbf{0.7105} \\
\bottomrule
\end{tabular}}
\label{TPR1}
\vspace{-0.2cm}
\end{minipage}
\hfill
\begin{minipage}[t]{0.49\textwidth}
\centering
\caption{TPR@5\%FPR under the complete 35-condition protocol. Values are macro-averaged over target-condition ROC operating points.}
\vspace{0.1cm}
\scriptsize
\setlength{\tabcolsep}{2pt}
\renewcommand{\arraystretch}{1.1}
\resizebox{\linewidth}{!}{
\begin{tabular}{l|ccccc}
\toprule
\textbf{Method} & \textbf{Clean} & \textbf{Seen} & \textbf{Unseen} & \textbf{All Post.} & \textbf{Plats.} \\
\midrule
OCC-CLIP     & 0.1196 & 0.1138 & 0.1187 & 0.1161 & 0.1237 \\
LatentTracer & 0.3783 & 0.1235 & 0.1813 & 0.1507 & 0.1110 \\
AEDR         & 0.9971 & 0.3374 & 0.6207 & 0.4707 & 0.4995 \\
\rowcolor{skyblue!20}
SpCon-TO     & 0.9306 & 0.5173 & 0.5850 & 0.5492 & 0.4752 \\
\rowcolor{skyblue!30}
SpCon-ER     & \underline{0.9971} & \underline{0.7504} & \underline{0.7498} & \underline{0.7501} & \underline{0.7248} \\
\rowcolor{skyblue!40}
SpCon-KR     & \textbf{1.0000} & \textbf{0.8510} & \textbf{0.8160} & \textbf{0.8346} & \textbf{0.8624} \\
\bottomrule
\end{tabular}}
\label{TPR2}
\vspace{-0.2cm}
\end{minipage}
\end{table*}

\subsection{Condition-Agnostic Fixed-Threshold Evaluation}

\begin{table}[t]
\centering
\caption{Condition-agnostic fixed-threshold performance (TPR/FPR) with a 1\% calibration FPR.}
\vspace{0.1cm}
\scriptsize
\setlength{\tabcolsep}{5pt}
\renewcommand{\arraystretch}{1}
\resizebox{\columnwidth}{!}{
\begin{tabular}{l|ccccc}
\toprule
\textbf{Method}
& \textbf{Clean}
& \textbf{Seen}
& \textbf{Unseen}
& \textbf{All Post.}
& \textbf{Platforms} \\
\midrule
OCC-CLIP
& 0.0379 / 0.0126
& 0.0248 / 0.0086
& 0.0306 / 0.0107
& 0.0275 / 0.0095
& 0.0335 / 0.0117 \\
LatentTracer
& 0.2725 / 0.0021
& 0.0360 / 0.0134
& 0.0447 / 0.0023
& 0.0401 / 0.0082
& 0.0019 / 0.0010 \\
AEDR
& 0.7543 / 0.0102
& 0.1302 / 0.0108
& 0.3215 / 0.0080
& 0.2203 / 0.0095
& 0.1600 / 0.0046 \\
\rowcolor{skyblue!20}
SpCon-TO
& 0.7032 / 0.0016
& 0.1543 / 0.0063
& 0.2098 / 0.0022
& 0.1804 / 0.0043
& 0.0905 / 0.0021 \\
\rowcolor{skyblue!30}
SpCon-ER
& \underline{0.9575} / 0.0034
& \underline{0.4414} / 0.0093
& \underline{0.4749} / 0.0022
& \underline{0.4572} / 0.0060
& \underline{0.3952} / 0.0033 \\
\rowcolor{skyblue!40}
SpCon-KR
& \textbf{0.9982} / 0.0051
& \textbf{0.6904} / 0.0145
& \textbf{0.6893} / 0.0079
& \textbf{0.6898} / 0.0114
& \textbf{0.6605} / 0.0069 \\
\bottomrule
\end{tabular}}
\label{Fixed_Threshold}
\vspace{-0.2cm}
\end{table}

To evaluate deployment-oriented performance without selecting a separate operating point from each test ROC, we calibrated one condition-agnostic threshold for every method-target pair at a target FPR of 1\%. For this auxiliary experiment, the 500-image held-out pool of each source was partitioned using a fixed random seed: 100 images were sampled from indices 100-499 for threshold calibration, while the other 400 images were retained for synthetic evaluation. Restricting threshold calibration to indices 100-499 also kept it disjoint from the first 100 base images used to construct the platform sets. The threshold-calibration subset was additionally disjoint from the 100-image reference-fitting set used by the main protocol.

Each base image was evaluated under Clean and the six medium-severity Seen conditions. We selected the most permissive empirical threshold whose calibration FPR did not exceed 1\%, and subsequently applied it unchanged to Clean, all 18 Seen and 16 Unseen conditions, and the three real-platform pipelines. Because raw score scales differ across methods and candidate VAEs, thresholds were calibrated separately for each method-target pair. Here, \emph{condition-agnostic} means that one threshold is shared across all conditions for a given method-target pair, rather than that an identical raw threshold is shared across different target VAEs. All methods used the same held-out labeled negative pool solely for calibrating the final decision threshold. The TO row therefore reports the TO scoring function under this shared calibration protocol.

As shown in \textcolor{refcolor}{Table}~\ref{Fixed_Threshold}, SpCon-KR achieves the highest TPR in every evaluation scope under the frozen thresholds. Over all post-processing conditions, it reaches a TPR/FPR of 0.6898/0.0114, compared with 0.2203/0.0095 for the strongest baseline, AEDR~\citep{AEDR}, yielding a TPR gain of 0.4695 with an absolute FPR increase of 0.0019. On real platforms, SpCon-KR reaches 0.6605/0.0069, improving the TPR over AEDR by 0.5005 with an absolute FPR increase of 0.0023. SpCon-ER further provides a Pareto improvement over AEDR in every scope, simultaneously achieving higher TPR and lower realized FPR. In particular, it obtains 0.4572/0.0060 over all post-processing conditions and 0.3952/0.0033 on real platforms. These results show that the sensitivity gain of the reference-assisted SpCon variants largely persists without condition-specific or test-time threshold adaptation.

%======================================================================================================
\section{Additional Ablation Studies}
\label{More_Ablation}

\subsection{Multi-Step Reconstruction and First-Transition Dominance}
\label{Multi_Step_SpCon}

\textbf{Protocol.}
To characterize how the response evolves under repeated reconstruction, we extend the deterministic sequence to five steps:
\begin{equation}
\mathbf{x}_{j}=P(\mathbf{x}_{j-1}),
\qquad
\mathbf{r}_{j}=\mathbf{x}_{j}-\mathbf{x}_{j-1},
\qquad j=1,\ldots,5.
\end{equation}
Each adjacent residual pair defines a frequency-resolved transition
\begin{equation}
z^{(j)}_{k,b}(\mathbf{x})
=
\log\!\bigl(\max\{\mathcal{B}_b(r_{j,k}),\varepsilon\}\bigr)
-
\log\!\bigl(\max\{\mathcal{B}_b(r_{j+1,k}),\varepsilon\}\bigr),
\qquad j=1,\ldots,4.
\end{equation}
The adopted SpCon response is therefore $z^{(1)}$, which requires exactly two reconstructions.

For each transition, we independently fit the same KR density backend while keeping the reference sources, fitted conditions, Student-$t$ parameterization, and component aggregation unchanged. Density fitting uses 100 images per fitted source-condition component, while evaluation uses 20 independent images per source under all 35 conditions. Fitting and evaluation images do not overlap. This mechanism-analysis split covers all seven target VAEs but differs from the 100-image ablation test split used elsewhere; its absolute values should therefore not be compared directly with the main evaluation tables. For Seen and Unseen results, severities are first averaged within each operation family, followed by equal averaging over operation families and the seven targets. We measure the normalized residual energy as
\begin{equation}
\bar{\mathcal{B}}_j(\mathbf{x})=\frac{\sum_{k}\sum_{b}\mathcal{B}_b(r_{j,k})}{\sum_{k}\sum_{b}\mathcal{B}_b(r_{1,k})},\qquad j=1,\ldots,5.
\end{equation}

To quantify whether later transitions preserve the source-discriminative direction of the first transition, let $d\in\{1,\ldots,48\}$ index a channel-band dimension and define
\begin{equation}
\delta^{(j)}_{m,c,d}=\mathbb{E}\!\left[z^{(j)}_{d}\mid\mathrm{Belonging}\right]-\mathbb{E}\!\left[z^{(j)}_{d}\mid\mathrm{Non\mbox{-}Belonging}\right]
\end{equation}
for target $m$ and condition $c$. Its agreement with the clean first-transition direction is
\begin{equation}
\alpha^{(j)}_{m,c}=\frac{1}{48}\sum_{d=1}^{48}\mathbf{1}\!\left[\operatorname{sgn}\!\left(\delta^{(j)}_{m,c,d}\right)=\operatorname{sgn}\!\left(\delta^{(1)}_{m,\mathrm{Clean},d}\right)
\right].
\end{equation}
A value of $0.5$ corresponds to chance-level sign agreement, while $\alpha^{(1)}_{m,\mathrm{Clean}}=1$ by construction.

\begin{figure*}[t]
\centering
\includegraphics[width=\textwidth]{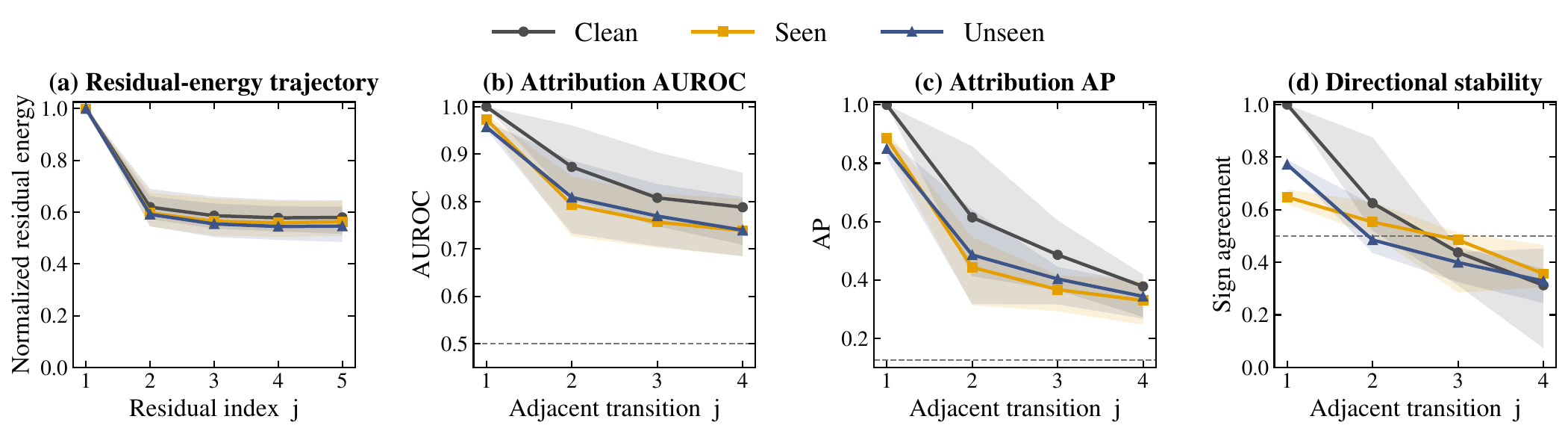}
\vspace{-0.5cm}
\caption{\textbf{Multi-step reconstruction and first-transition dominance.} (a) Normalized residual-energy trajectories over five reconstructions. (b-c) AUROC and AP obtained from the four adjacent transitions $z^{(1)},\ldots,z^{(4)}$ under the KR setting. (d) Agreement between each transition's source-discriminative direction and that of the clean first transition. The dashed lines indicate random-ranking AUROC, random-ranking AP, and chance-level sign agreement. Clean agreement at $j=1$ equals one by construction. Shaded regions indicate the interquartile range across evaluation units.}
\label{Multi_Step_Figure}
\end{figure*}

\begin{table}[t]
\centering
\caption{Attribution performance of adjacent residual transitions under the KR setting (AUROC / AP). Results use operation-family-balanced macro-averaging over seven targets.}
\vspace{0.1cm}
\scriptsize
\setlength{\tabcolsep}{6pt}
\renewcommand{\arraystretch}{1}
\resizebox{0.82\columnwidth}{!}{
\begin{tabular}{l|cccc}
\toprule
\textbf{Condition}
& \makecell{\textbf{$z^{(1)}$}\\$\mathbf{r}_1\!\rightarrow\!\mathbf{r}_2$}
& \makecell{\textbf{$z^{(2)}$}\\$\mathbf{r}_2\!\rightarrow\!\mathbf{r}_3$}
& \makecell{\textbf{$z^{(3)}$}\\$\mathbf{r}_3\!\rightarrow\!\mathbf{r}_4$}
& \makecell{\textbf{$z^{(4)}$}\\$\mathbf{r}_4\!\rightarrow\!\mathbf{r}_5$} \\
\midrule
Clean
& \cellcolor{skyblue!30}\textbf{1.0000 / 1.0000}
& 0.8730 / 0.6143
& 0.8077 / 0.4860
& 0.7879 / 0.3777 \\
Seen
& \cellcolor{skyblue!30}\textbf{0.9735 / 0.8856}
& 0.7927 / 0.4432
& 0.7565 / 0.3669
& 0.7387 / 0.3296 \\
Unseen
& \cellcolor{skyblue!30}\textbf{0.9570 / 0.8497}
& 0.8087 / 0.4857
& 0.7693 / 0.4031
& 0.7396 / 0.3435 \\
\bottomrule
\end{tabular}}
\label{Multi_Step_Table}
\vspace{-0.2cm}
\end{table}

\textbf{Results.}
The first adjacent transition is consistently dominant across Clean, Seen, and Unseen conditions (\textcolor{refcolor}{Figure}~\ref{Multi_Step_Figure} and \textcolor{refcolor}{Table}~\ref{Multi_Step_Table}). From $z^{(1)}$ to $z^{(4)}$, attribution decreases from 0.9735/0.8856 to 0.7387/0.3296 AUROC/AP on Seen conditions and from 0.9570/0.8497 to 0.7396/0.3435 on Unseen conditions. Thus, later transitions retain some source information but progressively weaken rather than improve attribution.

The normalized residual energy undergoes its dominant reduction during the first transition. For Clean, Seen, and Unseen conditions, it decreases from 1 to 0.6195, 0.5964, and 0.5914 at $\mathbf{r}_2$, respectively, but reaches only 0.5797, 0.5635, and 0.5461 by $\mathbf{r}_5$. The limited change after $\mathbf{r}_2$ is consistent with, but does not establish, local convergence of repeated reconstruction.

Directional stability exhibits the same pattern. Across the later transitions $j\in\{2,3,4\}$, the median sign agreement with the clean first-transition direction is only 49.1\% under Seen conditions and 39.9\% under Unseen conditions. The useful response is therefore concentrated in the first transition rather than preserved as a stable multi-step sign pattern. Together, these results provide an empirical justification for using two reconstructions. They are the minimum required to construct $z^{(1)}$ and capture the strongest source-discriminative transition. Later transitions require more computation and perform worse when scored individually.

\subsection{Density Model and Fusion Strategy}
We compare Student-$t$ against Gaussian and logsumexp against hard-min (see \textcolor{refcolor}{Table}~\ref{Student_Logsumexp}). Here hard-min minimizes the component negative log-likelihood, equivalently $L^{\mathrm{hard}}_{\mathcal{A}}(\mathbf{z})=\max_{(a,c)\in\mathcal{A}}\log q_{a,c}(\mathbf{z})$. It replaces each mixture log-density $L_{\mathcal{A}}$ in \textcolor{refcolor}{Equation}~\ref{Attribution_Score_Eq}. Student-$t$ with logsumexp performs best for ER and KR, while all four combinations lie within 0.0004 AUROC and 0.0046 AP under TO. For a fixed density, logsumexp consistently improves over hard-min. Performance remains stable when varying $\nu$ from 5 to 100 (see \textcolor{refcolor}{Appendix}~\ref{Nu_Sensitivity}). The density and fusion choices therefore provide a stable backend; with this backend held fixed, the representation accounts for the gain (see \textcolor{refcolor}{Table}~\ref{AEDR_Mix}).

\subsection{Sensitivity to \texorpdfstring{Student-$t$}{Student-t} Degrees of Freedom}
\label{Nu_Sensitivity}

The main protocol uses $n=100$ reference responses per source-condition component, for which the predefined rule $\nu=2+\sqrt{n}$ gives $\nu=12$. To determine whether the results depend on this particular tail thickness, we fix $n=100$ and vary $\nu\in\{5,10,12,20,30,50,75,100\}$. All variants use the same spectral-contraction responses, reference images, fitted condition components, Ledoit-Wolf covariance estimates, and equal-weight logsumexp aggregation; only $\nu$ changes. Following \textcolor{refcolor}{Equation}~\ref{Student_Params}, the scale matrix is rescaled for each $\nu$ so that the implied covariance remains $\widehat{\mathbf{C}}_{a,c}$ and the comparison isolates tail thickness. To avoid assigning greater weight to operation families with three severity levels, we use a family-balanced aggregation. Severities are first averaged within each operation family; Clean and the twelve operation families are then weighted equally within each target, followed by equal averaging over the seven targets. We refer to this aggregation as \emph{All-13}.

\begin{table}[t]
\centering
\caption{Sensitivity to the Student-$t$ degrees of freedom under the family-balanced All-13 protocol.}
\vspace{0.1cm}
\scriptsize
\setlength{\tabcolsep}{2pt}
\renewcommand{\arraystretch}{1.1}
\resizebox{\columnwidth}{!}{
\begin{tabular}{c|cccccccc}
\toprule
\textbf{Type} & \textbf{$\nu=5$} & \textbf{$\nu=10$} & \cellcolor{skyblue!30}\textbf{$\nu=12$} & \textbf{$\nu=20$} & \textbf{$\nu=30$} & \textbf{$\nu=50$} & \textbf{$\nu=75$} & \textbf{$\nu=100$} \\
\midrule
SpCon-TO & 0.8822 / 0.6382 & 0.8824 / 0.6400 & \cellcolor{skyblue!30}0.8825 / 0.6405 & 0.8826 / 0.6421 & 0.8827 / 0.6430 & 0.8827 / 0.6439 & 0.8827 / 0.6443 & 0.8827 / 0.6444 \\
SpCon-ER & 0.9405 / 0.7912 & 0.9421 / 0.7920 & \cellcolor{skyblue!30}0.9425 / 0.7921 & 0.9432 / 0.7916 & 0.9434 / 0.7906 & 0.9432 / 0.7885 & 0.9427 / 0.7865 & 0.9424 / 0.7851 \\
SpCon-KR & 0.9653 / 0.8622 & 0.9655 / 0.8616 & \cellcolor{skyblue!30}0.9655 / 0.8613 & 0.9652 / 0.8594 & 0.9647 / 0.8569 & 0.9637 / 0.8533 & 0.9626 / 0.8501 & 0.9619 / 0.8481 \\
\bottomrule
\end{tabular}}
\label{Nu_Sensitivity_Table}
\vspace{-0.2cm}
\end{table}

\begin{table}[t]
\centering
\caption{Ablation of the Student-$t$ scale-matrix structure under the family-balanced All-13 protocol (AUROC / AP). Scaled Identity preserves the empirical mean variance.}
\vspace{0.1cm}
\scriptsize
\setlength{\tabcolsep}{6pt}
\renewcommand{\arraystretch}{1}
\resizebox{0.65\columnwidth}{!}{
\begin{tabular}{l|ccc}
\toprule
\textbf{Scale Matrix}
& \textbf{SpCon-TO}
& \textbf{SpCon-ER}
& \textbf{SpCon-KR} \\
\midrule
Scaled Identity
& 0.8034 / 0.4436
& 0.9000 / 0.7101
& 0.9101 / 0.7199 \\
Diagonal
& 0.8152 / 0.4688
& 0.9074 / 0.7247
& 0.9202 / 0.7404 \\
Full Sample
& \underline{0.8499} / \underline{0.5670}
& \underline{0.9234} / \underline{0.7261}
& \underline{0.9564} / \underline{0.8281} \\
\rowcolor{skyblue!30}
Ledoit-Wolf
& \textbf{0.8825} / \textbf{0.6405}
& \textbf{0.9425} / \textbf{0.7921}
& \textbf{0.9655} / \textbf{0.8613} \\
\bottomrule
\end{tabular}}
\label{Scale_Matrix_Table}
\vspace{-0.2cm}
\end{table}

As shown in \textcolor{refcolor}{Table}~\ref{Nu_Sensitivity_Table}, SpCon shows limited sensitivity to the Student-$t$ degrees of freedom. Across $\nu\in[5,100]$, TO remains within 0.8822-0.8827 AUROC and 0.6382-0.6444 AP, ER within 0.9405-0.9434 and 0.7851-0.7921, and KR within 0.9619-0.9655 and 0.8481-0.8622. Relative to the default $\nu=12$, the largest absolute change in either metric is only 0.0039 for TO, 0.0070 for ER, and 0.0132 for KR. The default value is therefore located within a broad high-performing region rather than selected at a narrow optimum. As $\nu$ increases and the Student-$t$ tails become lighter, KR AP gradually decreases from 0.8622 at $\nu=5$ to 0.8481 at $\nu=100$. This pattern suggests that finite tail thickness provides a modest benefit when contrasting belonging responses against competing in-domain sources. Nevertheless, the limited variation across the full sweep shows that the conclusions are stable over a broad range of $\nu$.

\subsection{Scale-Matrix Structure}
\label{Scale_Matrix_Ablation}

We isolate the effect of the Student-$t$ scale matrix while keeping the 48-dimensional SpCon representation, $\nu=12$, $\varepsilon=10^{-12}$, reference sets, fitted condition components, and equal-weight logsumexp aggregation fixed. Each source-condition component is fitted using 100 responses and evaluated on 100 disjoint images. We compare four covariance structures. \emph{Scaled Identity} uses an isotropic covariance whose variance equals the empirical mean variance, thereby preserving the overall response scale. \emph{Diagonal} retains dimension-wise variances but sets all off-diagonal covariance entries to zero. \emph{Full Sample} uses the unregularized empirical covariance, while \emph{Ledoit-Wolf} applies shrinkage to the full covariance estimate. For each covariance structure, the Student-$t$ scale matrix is rescaled as in \textcolor{refcolor}{Equation}~\ref{Student_Params}, so that the implied Student-$t$ covariance matches the corresponding covariance estimate. Results use the family-balanced All-13 aggregation defined in \textcolor{refcolor}{Appendix}~\ref{Nu_Sensitivity}. The full sample covariance consistently outperforms the diagonal alternative, indicating that cross-channel and cross-band dependencies provide useful attribution information. Ledoit-Wolf shrinkage further improves all three settings over the unregularized full covariance (see \textcolor{refcolor}{Table}~\ref{Scale_Matrix_Table}): TO increases from 0.8499/0.5670 to 0.8825/0.6405, ER from 0.9234/0.7261 to 0.9425/0.7921, and KR from 0.9564/0.8281 to 0.9655/0.8613. These results support the use of a shrinkage full-scale model for covariance estimation in the 48-dimensional, 100-sample setting, rather than treating it as an arbitrary implementation choice.

\subsection{External Reference Composition}
\label{Appendix_External_Reference}
We vary both the type and composition of the external references, evaluating four single-source settings and all four natural-generated pairs (see \textcolor{refcolor}{Table}~\ref{External_Reference}). Each individual reference already provides a viable ER model, indicating that neither a particular dataset nor the simultaneous availability of both reference types is indispensable. Nevertheless, pairing a natural-image source with a generated-image source consistently improves Clean, Seen, and Unseen performance over the corresponding single-reference variants. The two KD3-based pairs perform best overall: ImageNet\,+\,KD3 achieves the highest Seen performance and Unseen AP, whereas LAION\,+\,KD3 obtains the highest Clean performance and Unseen AUROC. This consistent complementarity indicates that ER benefits primarily from reference diversity rather than relying on a specific external source.

\begin{table}[t]
\centering
\caption{Ablation of the response density and conditional fusion rule under the ablation protocol of \textcolor{refcolor}{Section}~\ref{SpCon_Ablation} (AUROC / AP), with all $7\times35$ target-condition cells weighted equally.}
\vspace{0.1cm}
\scriptsize
\setlength{\tabcolsep}{5pt}
\renewcommand{\arraystretch}{1}
\resizebox{0.7\columnwidth}{!}{
\begin{tabular}{cc|ccc}
\toprule
\textbf{Density} & \textbf{Fusion} & \textbf{SpCon-TO} & \textbf{SpCon-ER} & \textbf{SpCon-KR} \\
\midrule
Gaussian & hard-min & 0.8775 / \underline{0.6316} & 0.9382 / 0.7748 & 0.9556 / 0.8340 \\
Gaussian & logsumexp & \textbf{0.8778} / \textbf{0.6322} & 0.9386 / 0.7758 & 0.9567 / 0.8371 \\
Student-$t$ & hard-min & 0.8774 / 0.6276 & \underline{0.9415} / \underline{0.7917} & \underline{0.9641} / \underline{0.8584} \\
\rowcolor{skyblue!30}
Student-$t$ & logsumexp & \underline{0.8776} / 0.6281 & \textbf{0.9420} / \textbf{0.7927} & \textbf{0.9648} / \textbf{0.8607} \\
\bottomrule
\end{tabular}}
\label{Student_Logsumexp}
\vspace{-0.2cm}
\end{table}

\begin{table}[t]
\centering
\caption{Ablation of external reference sources (all 35 conditions); ImageNet\,+\,KD3 is the default.}
\vspace{0.1cm}
\scriptsize
\setlength{\tabcolsep}{5pt}
\renewcommand{\arraystretch}{1}
\resizebox{0.8\columnwidth}{!}{
\begin{tabular}{cc|ccc}
\toprule
\textbf{Natural} & \textbf{Generated} & \textbf{Clean} & \textbf{Seen} & \textbf{Unseen} \\
\midrule
LAION    & -    & 0.9850 / 0.9368 & 0.9027 / 0.6432 & 0.8911 / 0.6291 \\
ImageNet & -    & 0.9918 / 0.9720 & 0.9175 / 0.6947 & 0.9049 / 0.7022 \\
-       & Qwen  & 0.9939 / 0.9792 & 0.8694 / 0.6664 & 0.8823 / 0.6840 \\
-       & KD3   & 0.9987 / 0.9945 & 0.9328 / 0.7667 & 0.9121 / 0.7657 \\
\midrule
LAION    & Qwen  & 0.9979 / 0.9910 & 0.9172 / 0.7081 & 0.9179 / 0.7288 \\
ImageNet & Qwen  & 0.9975 / 0.9894 & 0.9206 / 0.7259 & 0.9162 / 0.7383 \\
LAION    & KD3   & \textbf{0.9991} / \textbf{0.9958}
                  & \underline{0.9468} / \underline{0.7808}
                  & \textbf{0.9336} / \underline{0.7772} \\
\rowcolor{skyblue!30}
ImageNet & KD3   & \underline{0.9989} / \underline{0.9952}
                  & \textbf{0.9494} / \textbf{0.7917}
                  & \underline{0.9301} / \textbf{0.7812} \\
\bottomrule
\end{tabular}}
\label{External_Reference}
\vspace{-0.2cm}
\end{table}

\subsection{Sensitivity to Seen/Unseen Partition}
\label{Appendix_Seen_Unseen}

We further examine the sensitivity of SpCon to the partition of post-processing operations into Seen and Unseen sets. We evaluate eight partitions, including the default, each with four non-geometric and two geometric Seen families and six Unseen families. Each family is represented by its medium-severity condition, except Flip, which has one condition without severity levels. SpCon is fitted using Clean and the six Seen conditions. All configurations share the same feature extraction, probabilistic modeling, and evaluation protocol; the only varying factor is which six operations are used for fitting.

As shown in \textcolor{refcolor}{Table}~\ref{Seen_Unseen}, the default partition is near-optimal for the reference-assisted settings. For ER, the default configuration achieves an All-12 AUROC/AP of 0.9439/0.7917, only 0.0012/0.0008 below the best result of 0.9451/0.7925. Similarly, KR achieves 0.9691/0.8703, with gaps of only 0.0008/0.0007 from the best result of 0.9699/0.8710; both rank second among the eight partitions. For TO, the default result of 0.8919/0.6634 is within 0.0133/0.0346 of its best All-12 performance of 0.9052/0.6980.

The default Seen set (JPEG, Noise, Blur, Brightness, Rotation, and Resize) covers diverse distortion mechanisms, including compression, random pixel perturbation, low-pass filtering, photometric variation, geometric transformation, and spatial resampling. These operations are also commonly encountered during image uploading, editing, resizing, re-encoding, and real-world platform distribution, making the configuration practically representative.

\begin{table}[t]
\centering
\caption{Sensitivity to the Seen/Unseen partition of post-processing operations. Results are reported as AUROC / AP. For each SpCon variant, the \textbf{best} and \underline{second-best} results across partitions are highlighted. Colored cells indicate the default configuration used in the main experiments.}
\vspace{0.1cm}
\scriptsize
\setlength{\tabcolsep}{4pt}
\renewcommand{\arraystretch}{1}
\resizebox{\columnwidth}{!}{
\begin{tabular}{@{}c|ccccc@{}}
\toprule
\textbf{Seen Operations} & \textbf{SpCon} & \textbf{Clean} & \textbf{Seen} & \textbf{Unseen} & \textbf{All-12} \\
\midrule

\multirow[c]{3}{*}{\makecell[c]{JPEG + Noise + Blur\\[2pt]Brightness + Resize + Crop}}
& TO & \underline{0.9891} / 0.9725 & 0.9129 / 0.7107 & \textbf{0.8976} / \textbf{0.6853} & \textbf{0.9052} / \textbf{0.6980} \\
& ER & \textbf{0.9995} / \underline{0.9976} & 0.9599 / 0.8242 & 0.8883 / 0.6915 & 0.9241 / 0.7579 \\
& KR & \textbf{0.9999} / \underline{0.9996} & 0.9831 / 0.9166 & 0.9378 / 0.7765 & 0.9605 / 0.8466 \\
\cmidrule{1-6}

\multirow[c]{3}{*}{\makecell[c]{JPEG + Noise + Blur\\[2pt]Rotation + Resize + Contrast}}
& TO & 0.9809 / 0.9475 & 0.9081 / 0.6945 & 0.8722 / 0.6305 & 0.8902 / 0.6625 \\
& ER & \underline{0.9989} / 0.9951 & 0.9637 / 0.8341 & 0.9227 / 0.7435 & 0.9432 / 0.7888 \\
& KR & \textbf{0.9999} / 0.9992 & 0.9832 / 0.9174 & 0.9549 / 0.8211 & 0.9690 / 0.8692 \\
\cmidrule{1-6}

\multirow[c]{3}{*}{\makecell[c]{JPEG + Noise + Blur\\[2pt]Resize + Contrast + Crop}}
& TO & 0.9890 / \underline{0.9728} & 0.9161 / 0.7192 & 0.8900 / 0.6695 & \underline{0.9031} / \underline{0.6943} \\
& ER & \textbf{0.9995} / \underline{0.9976} & 0.9606 / 0.8266 & 0.8864 / 0.6851 & 0.9235 / 0.7559 \\
& KR & \textbf{0.9999} / \underline{0.9996} & 0.9834 / 0.9177 & 0.9362 / 0.7724 & 0.9598 / 0.8451 \\
\cmidrule{1-6}

\multirow[c]{3}{*}{\makecell[c]{JPEG + Noise + Brightness\\[2pt]Resize + Contrast + Crop}}
& TO & \textbf{0.9892} / \textbf{0.9733} & \textbf{0.9272} / \textbf{0.7575} & 0.8523 / 0.6142 & 0.8898 / 0.6859 \\
& ER & \textbf{0.9995} / \textbf{0.9978} & \underline{0.9649} / \underline{0.8438} & 0.8696 / 0.6349 & 0.9173 / 0.7393 \\
& KR & \textbf{0.9999} / \textbf{0.9997} & \underline{0.9866} / \underline{0.9338} & 0.9131 / 0.7098 & 0.9498 / 0.8218 \\
\cmidrule{1-6}

\multirow[c]{3}{*}{\makecell[c]{Noise + Blur + Brightness\\[2pt]Rotation + Resize + WebP}}
& TO & 0.9803 / 0.9457 & 0.9018 / 0.6816 & 0.8740 / 0.6283 & 0.8879 / 0.6550 \\
& ER & \underline{0.9989} / 0.9952 & 0.9619 / 0.8261 & \textbf{0.9282} / \textbf{0.7588} & \textbf{0.9451} / \textbf{0.7925} \\
& KR & \textbf{0.9999} / 0.9993 & 0.9821 / 0.9109 & \textbf{0.9577} / \textbf{0.8312} & \textbf{0.9699} / \textbf{0.8710} \\
\cmidrule{1-6}

\multirow[c]{3}{*}{\makecell[c]{Noise + Blur + Brightness\\[2pt]Resize + WebP + Crop}}
& TO & 0.9883 / 0.9703 & 0.9081 / 0.6965 & \underline{0.8931} / \underline{0.6746} & 0.9006 / 0.6856 \\
& ER & \textbf{0.9995} / 0.9975 & 0.9582 / 0.8151 & 0.8940 / 0.7064 & 0.9261 / 0.7607 \\
& KR & \textbf{0.9999} / \underline{0.9996} & 0.9820 / 0.9116 & 0.9419 / 0.7862 & 0.9620 / 0.8489 \\
\cmidrule{1-6}

\multirow[c]{3}{*}{\makecell[c]{WebP + Sharpen + Contrast\\[2pt]Saturation + Flip + Crop}}
& TO & 0.9790 / 0.9386 & \underline{0.9173} / \underline{0.7364} & 0.6986 / 0.3794 & 0.8079 / 0.5579 \\
& ER & \underline{0.9989} / 0.9962 & \textbf{0.9688} / \textbf{0.8624} & 0.8414 / 0.5976 & 0.9051 / 0.7300 \\
& KR & \textbf{0.9999} / 0.9994 & \textbf{0.9882} / \textbf{0.9420} & 0.8765 / 0.6434 & 0.9323 / 0.7927 \\
\cmidrule{1-6}

& \cellcolor{skyblue!20}TO
& \cellcolor{skyblue!20}0.9807 / 0.9469
& \cellcolor{skyblue!20}0.9049 / 0.6862
& \cellcolor{skyblue!20}0.8790 / 0.6406
& \cellcolor{skyblue!20}0.8919 / 0.6634 \\

& \cellcolor{skyblue!30}ER
& \cellcolor{skyblue!30}\underline{0.9989} / 0.9952
& \cellcolor{skyblue!30}0.9629 / 0.8321
& \cellcolor{skyblue!30}\underline{0.9249} / \underline{0.7513}
& \cellcolor{skyblue!30}\underline{0.9439} / \underline{0.7917} \\

\multirow[c]{-3}{*}{\makecell[c]{JPEG + Noise + Blur\\[2pt]Brightness + Rotation + Resize}}
& \cellcolor{skyblue!40}KR
& \cellcolor{skyblue!40}\textbf{0.9999} / 0.9994
& \cellcolor{skyblue!40}0.9829 / 0.9160
& \cellcolor{skyblue!40}\underline{0.9552} / \underline{0.8246}
& \cellcolor{skyblue!40}\underline{0.9691} / \underline{0.8703} \\
\bottomrule
\end{tabular}}
\label{Seen_Unseen}
\vspace{-0.5cm}
\end{table}

%======================================================================================================
\section{Bootstrap Significance Analysis}
\label{BSA}

\subsection{Significance on Clean Images}

\begin{table*}[t]
\centering
\caption{Significance analysis under clean conditions for \textcolor{refcolor}{Table}~\ref{SpCon_Clean} (AUROC/AP).}
\vspace{0.1cm}
\scriptsize
\setlength{\tabcolsep}{2.5pt}
\renewcommand{\arraystretch}{1.1}
\resizebox{\textwidth}{!}{
\begin{tabular}{ccc|ccc}
\toprule
\textbf{Model} & \textbf{Metric} & \textbf{Best Baseline} & \textbf{TO: $\Delta$ / 95\% CI} & \textbf{ER: $\Delta$ / 95\% CI} & \textbf{KR: $\Delta$ / 95\% CI} \\
\midrule
\multirow{2}{*}{SD1}
& AUROC & 0.9993 & -0.03847 / [-0.04865, -0.02860] & -0.00123 / [-0.00260, -0.00029] & +0.00051 / [+0.00013, +0.00097] \\
& AP & 0.9954 & -0.09208 / [-0.10928, -0.07472] & -0.00411 / [-0.00889, +0.00064] & +0.00371 / [+0.00093, +0.00740] \\
\cmidrule{1-6}
\multirow{2}{*}{SD2}
& AUROC & 0.9981 & -0.02866 / [-0.03783, -0.02038] & +0.00155 / [+0.00039, +0.00302] & +0.00187 / [+0.00072, +0.00332] \\
& AP & 0.9855 & -0.06258 / [-0.08290, -0.04194] & +0.01247 / [+0.00428, +0.02542] & +0.01448 / [+0.00625, +0.02786] \\
\cmidrule{1-6}
\multirow{2}{*}{SD3}
& AUROC & 0.9961 & -0.01116 / [-0.01838, -0.00521] & +0.00209 / [-0.00001, +0.00407] & +0.00385 / [+0.00253, +0.00538] \\
& AP & 0.9638 & -0.00087 / [-0.02125, +0.02122] & +0.02892 / [+0.01116, +0.05006] & +0.03586 / [+0.01930, +0.05592] \\
\cmidrule{1-6}
\multirow{2}{*}{SDXL}
& AUROC & 0.9993 & -0.01205 / [-0.01748, -0.00749] & +0.00053 / [-0.00005, +0.00129] & +0.00059 / [+0.00004, +0.00136] \\
& AP & 0.9884 & -0.02919 / [-0.04734, -0.00800] & +0.01070 / [+0.00012, +0.02837] & +0.01105 / [+0.00055, +0.02866] \\
\cmidrule{1-6}
\multirow{2}{*}{FLUX.1}
& AUROC & 0.9929 & +0.00420 / [+0.00084, +0.00722] & +0.00695 / [+0.00487, +0.00922] & +0.00707 / [+0.00501, +0.00932] \\
& AP & 0.9250 & +0.06477 / [+0.03738, +0.09347] & +0.07436 / [+0.04774, +0.10381] & +0.07491 / [+0.04826, +0.10461] \\
\cmidrule{1-6}
\multirow{2}{*}{FLUX.2}
& AUROC & 0.9979 & -0.00622 / [-0.01000, -0.00280] & +0.00191 / [+0.00062, +0.00396] & +0.00211 / [+0.00073, +0.00442] \\
& AP & 0.9858 & -0.01102 / [-0.02281, +0.00234] & +0.01296 / [+0.00453, +0.02434] & +0.01420 / [+0.00562, +0.02554] \\
\cmidrule{1-6}
\multirow{2}{*}{PixArt}
& AUROC & 0.9980 & -0.01707 / [-0.02330, -0.01120] & +0.00185 / [+0.00092, +0.00300] & +0.00191 / [+0.00097, +0.00306] \\
& AP & 0.9846 & -0.04151 / [-0.05732, -0.02482] & +0.01472 / [+0.00666, +0.02490] & +0.01510 / [+0.00702, +0.02518] \\
\bottomrule
\end{tabular}}
\label{SpCon_Clean_bootstrap}
\vspace{-0.5cm}
\end{table*}

\textcolor{refcolor}{Table}~\ref{SpCon_Clean_bootstrap} compares each SpCon variant with the strongest official baseline for every target and metric. Each Clean target cell contains 500 belonging and 3,500 non-belonging images. We apply the shared resampling protocol above, including averaging OCC-CLIP~\citep{OCC-CLIP} metrics over five fixed training seeds. We report $\Delta$ = SpCon - baseline with its 95\% confidence interval; intervals entirely above zero indicate significant improvements. KR significantly outperforms the strongest baseline in all 14 target-metric comparisons. ER improves significantly in 10 of 14 comparisons, whereas TO does so in 2 of 14. AEDR~\citep{AEDR} is the strongest baseline in 13 comparisons; the only exception is AP on SD2, where LatentTracer~\citep{LatentTracer} performs best.

\subsection{Significance on Seen Post-Processing Operations}

\begin{table*}[t]
\centering
\caption{Significance analysis under Seen post-processing operations for \textcolor{refcolor}{Table}~\ref{SpCon_Seen} (AUROC).}
\vspace{0.1cm}
\scriptsize
\setlength{\tabcolsep}{2.5pt}
\renewcommand{\arraystretch}{1.1}
\resizebox{\textwidth}{!}{
\begin{tabular}{llc|ccc}
\toprule
\textbf{Operation} & \textbf{Severity} & \textbf{Best Baseline} & \textbf{SpCon-TO $\Delta$ / 95\% CI} & \textbf{SpCon-ER $\Delta$ / 95\% CI} & \textbf{SpCon-KR $\Delta$ / 95\% CI} \\
\midrule
\multirow{3}{*}{JPEG}
& L       & 0.9503 & -0.1055 / [-0.1117, -0.0991] & +0.0094 / [+0.0060, +0.0127] & +0.0412 / [+0.0381, +0.0444] \\
& M (Fit) & 0.8728 & +0.0178 / [+0.0120, +0.0238] & +0.0710 / [+0.0662, +0.0754] & +0.1083 / [+0.1037, +0.1125] \\
& H       & 0.8191 & +0.0356 / [+0.0289, +0.0424] & +0.0977 / [+0.0921, +0.1032] & +0.1385 / [+0.1331, +0.1436] \\
\cmidrule{1-6}
\multirow{3}{*}{Noise}
& L       & 0.8575 & -0.0390 / [-0.0487, -0.0297] & +0.0452 / [+0.0363, +0.0542] & +0.0970 / [+0.0894, +0.1048] \\
& M (Fit) & 0.6647 & +0.2136 / [+0.2055, +0.2221] & +0.2572 / [+0.2500, +0.2648] & +0.3024 / [+0.2948, +0.3104] \\
& H       & 0.6305 & +0.2151 / [+0.2091, +0.2211] & +0.2360 / [+0.2302, +0.2420] & +0.2896 / [+0.2844, +0.2948] \\
\cmidrule{1-6}
\multirow{3}{*}{Blur}
& L       & 0.8976 & -0.0850 / [-0.0920, -0.0783] & +0.0463 / [+0.0413, +0.0509] & +0.0661 / [+0.0614, +0.0705] \\
& M (Fit) & 0.8305 & +0.0386 / [+0.0311, +0.0461] & +0.1268 / [+0.1200, +0.1338] & +0.1459 / [+0.1392, +0.1524] \\
& H       & 0.7984 & +0.0491 / [+0.0415, +0.0568] & +0.1429 / [+0.1362, +0.1501] & +0.1631 / [+0.1564, +0.1699] \\
\cmidrule{1-6}
\multirow{3}{*}{Brightness}
& L       & 0.9939 & -0.0446 / [-0.0488, -0.0403] & +0.0032 / [+0.0023, +0.0041] & +0.0057 / [+0.0048, +0.0067] \\
& M (Fit) & 0.9734 & -0.0454 / [-0.0509, -0.0405] & +0.0183 / [+0.0160, +0.0207] & +0.0244 / [+0.0220, +0.0268] \\
& H       & 0.9373 & -0.0457 / [-0.0524, -0.0392] & +0.0419 / [+0.0384, +0.0456] & +0.0510 / [+0.0473, +0.0550] \\
\cmidrule{1-6}
\multirow{3}{*}{Resize}
& L       & 0.7154 & +0.1426 / [+0.1348, +0.1500] & +0.2260 / [+0.2194, +0.2323] & +0.2740 / [+0.2681, +0.2798] \\
& M (Fit) & 0.6840 & +0.2668 / [+0.2590, +0.2739] & +0.3096 / [+0.3028, +0.3157] & +0.3148 / [+0.3080, +0.3210] \\
& H       & 0.6718 & +0.0295 / [+0.0212, +0.0378] & +0.2661 / [+0.2603, +0.2718] & +0.2283 / [+0.2222, +0.2345] \\
\cmidrule{1-6}
\multirow{3}{*}{Rotation}
& L       & 0.6670 & +0.2488 / [+0.2422, +0.2548] & +0.2860 / [+0.2811, +0.2904] & +0.3119 / [+0.3075, +0.3158] \\
& M (Fit) & 0.6663 & +0.2494 / [+0.2426, +0.2559] & +0.2858 / [+0.2808, +0.2905] & +0.3104 / [+0.3059, +0.3145] \\
& H       & 0.6677 & +0.2415 / [+0.2340, +0.2477] & +0.2790 / [+0.2736, +0.2839] & +0.3028 / [+0.2978, +0.3071] \\
\bottomrule
\end{tabular}}
\label{SpCon_Seen_bootstrap_AUROC}
\vspace{-0.2cm}
\end{table*}

\begin{table*}[t]
\centering
\caption{Significance analysis under Seen post-processing operations for \textcolor{refcolor}{Table}~\ref{SpCon_Seen} (AP).}
\vspace{0.1cm}
\scriptsize
\setlength{\tabcolsep}{2.5pt}
\renewcommand{\arraystretch}{1.2}
\resizebox{\textwidth}{!}{
\begin{tabular}{llc|ccc}
\toprule
\textbf{Operation} & \textbf{Severity} & \textbf{Best Baseline} & \textbf{SpCon-TO $\Delta$ / 95\% CI} & \textbf{SpCon-ER $\Delta$ / 95\% CI} & \textbf{SpCon-KR $\Delta$ / 95\% CI} \\
\midrule
\multirow{3}{*}{JPEG}
& L       & 0.6878 & -0.1903 / [-0.2115, -0.1707] & +0.1427 / [+0.1261, +0.1575] & +0.2705 / [+0.2542, +0.2840] \\
& M (Fit) & 0.4733 & +0.1905 / [+0.1727, +0.2055] & +0.2943 / [+0.2775, +0.3097] & +0.4405 / [+0.4243, +0.4533] \\
& H       & 0.3748 & +0.1621 / [+0.1461, +0.1768] & +0.2981 / [+0.2832, +0.3123] & +0.4320 / [+0.4153, +0.4440] \\
\cmidrule{1-6}
\multirow{3}{*}{Noise}
& L       & 0.5678 & -0.2159 / [-0.2390, -0.1914] & +0.0980 / [+0.0738, +0.1207] & +0.2333 / [+0.2117, +0.2526] \\
& M (Fit) & 0.2915 & +0.2573 / [+0.2383, +0.2729] & +0.3728 / [+0.3580, +0.3862] & +0.5571 / [+0.5415, +0.5695] \\
& H       & 0.1999 & +0.2925 / [+0.2787, +0.3042] & +0.3331 / [+0.3206, +0.3448] & +0.4531 / [+0.4385, +0.4651] \\
\cmidrule{1-6}
\multirow{3}{*}{Blur}
& L       & 0.5689 & -0.0398 / [-0.0571, -0.0255] & +0.2369 / [+0.2203, +0.2518] & +0.2926 / [+0.2759, +0.3067] \\
& M (Fit) & 0.4265 & +0.1876 / [+0.1706, +0.2023] & +0.3961 / [+0.3805, +0.4106] & +0.4621 / [+0.4466, +0.4760] \\
& H       & 0.3666 & +0.2211 / [+0.2061, +0.2339] & +0.4092 / [+0.3941, +0.4213] & +0.4703 / [+0.4554, +0.4825] \\
\cmidrule{1-6}
\multirow{3}{*}{Brightness}
& L       & 0.9437 & -0.1073 / [-0.1203, -0.0946] & +0.0416 / [+0.0310, +0.0523] & +0.0544 / [+0.0434, +0.0662] \\
& M (Fit) & 0.8127 & -0.0269 / [-0.0476, -0.0093] & +0.1441 / [+0.1255, +0.1602] & +0.1757 / [+0.1574, +0.1924] \\
& H       & 0.6594 & +0.0314 / [+0.0103, +0.0494] & +0.2343 / [+0.2145, +0.2492] & +0.2832 / [+0.2634, +0.2981] \\
\cmidrule{1-6}
\multirow{3}{*}{Resize}
& L       & 0.2966 & +0.1143 / [+0.0966, +0.1321] & +0.4970 / [+0.4816, +0.5087] & +0.6259 / [+0.6112, +0.6378] \\
& M (Fit) & 0.2234 & +0.6291 / [+0.6147, +0.6386] & +0.7419 / [+0.7323, +0.7472] & +0.7691 / [+0.7602, +0.7734] \\
& H       & 0.2170 & +0.1122 / [+0.1018, +0.1223] & +0.5259 / [+0.5103, +0.5397] & +0.4835 / [+0.4697, +0.4962] \\
\cmidrule{1-6}
\multirow{3}{*}{Rotation}
& L       & 0.2121 & +0.4953 / [+0.4803, +0.5075] & +0.5656 / [+0.5506, +0.5774] & +0.6714 / [+0.6580, +0.6809] \\
& M (Fit) & 0.2110 & +0.5043 / [+0.4893, +0.5169] & +0.5636 / [+0.5495, +0.5755] & +0.6639 / [+0.6504, +0.6733] \\
& H       & 0.2114 & +0.4685 / [+0.4518, +0.4812] & +0.5337 / [+0.5189, +0.5470] & +0.6349 / [+0.6209, +0.6464] \\
\bottomrule
\end{tabular}}
\label{SpCon_Seen_bootstrap_AP}
\vspace{-0.2cm}
\end{table*}

These two tables report the significance results for AUROC (see \textcolor{refcolor}{Table}~\ref{SpCon_Seen_bootstrap_AUROC}) and AP (see \textcolor{refcolor}{Table}~\ref{SpCon_Seen_bootstrap_AP}), respectively, under Seen post-processing. Each entry gives the performance difference between SpCon and the strongest official baseline for the corresponding condition and metric, together with its 95\% confidence interval. SpCon-ER and SpCon-KR achieve significant improvements in all 36 condition-metric comparisons, while SpCon-TO does so in 25, showing the consistent benefit of reference-assisted scoring across the tested operations and severity levels.

\subsection{Significance on Unseen Post-Processing Operations}

\begin{table*}[t]
\centering
\caption{Significance analysis under Unseen post-processing operations for \textcolor{refcolor}{Table}~\ref{SpCon_Unseen} (AUROC).}
\vspace{0.1cm}
\scriptsize
\setlength{\tabcolsep}{3pt}
\renewcommand{\arraystretch}{1.2}
\resizebox{\textwidth}{!}{
\begin{tabular}{lcc|ccc}
\toprule
\textbf{Operation} & \textbf{Severity} & \textbf{Best Baseline} & \textbf{SpCon-TO $\Delta$ / 95\% CI} & \textbf{SpCon-ER $\Delta$ / 95\% CI} & \textbf{SpCon-KR $\Delta$ / 95\% CI} \\
\midrule
\multirow{3}{*}{WebP}
& L & 0.9558 & -0.1227 / [-0.1282, -0.1170] & -0.0070 / [-0.0104, -0.0036] & +0.0321 / [+0.0294, +0.0348] \\
& M & 0.8675 & -0.0344 / [-0.0404, -0.0282] & +0.0440 / [+0.0393, +0.0487] & +0.0817 / [+0.0770, +0.0864] \\
& H & 0.8317 & -0.0370 / [-0.0438, -0.0294] & +0.0495 / [+0.0440, +0.0554] & +0.0902 / [+0.0849, +0.0958] \\
\cmidrule{1-6}

\multirow{3}{*}{Sharpen}
& L & 0.9951 & -0.0232 / [-0.0263, -0.0198] & +0.0030 / [+0.0020, +0.0041] & +0.0048 / [+0.0038, +0.0058] \\
& M & 0.9863 & -0.0325 / [-0.0362, -0.0283] & +0.0086 / [+0.0069, +0.0104] & +0.0131 / [+0.0114, +0.0149] \\
& H & 0.9695 & -0.0360 / [-0.0406, -0.0312] & +0.0179 / [+0.0149, +0.0209] & +0.0282 / [+0.0254, +0.0310] \\
\cmidrule{1-6}

\multirow{3}{*}{Contrast}
& L & 0.9940 & -0.0415 / [-0.0454, -0.0373] & +0.0035 / [+0.0026, +0.0045] & +0.0057 / [+0.0048, +0.0066] \\
& M & 0.9764 & -0.0514 / [-0.0568, -0.0462] & +0.0144 / [+0.0123, +0.0165] & +0.0211 / [+0.0189, +0.0233] \\
& H & 0.9497 & -0.0583 / [-0.0639, -0.0522] & +0.0251 / [+0.0215, +0.0282] & +0.0385 / [+0.0351, +0.0418] \\
\cmidrule{1-6}

\multirow{3}{*}{Saturation}
& L & 0.9967 & -0.0190 / [-0.0219, -0.0158] & +0.0025 / [+0.0016, +0.0033] & +0.0032 / [+0.0024, +0.0041] \\
& M & 0.9953 & -0.0254 / [-0.0290, -0.0220] & +0.0033 / [+0.0022, +0.0042] & +0.0046 / [+0.0036, +0.0055] \\
& H & 0.9922 & -0.0340 / [-0.0376, -0.0305] & +0.0045 / [+0.0031, +0.0057] & +0.0076 / [+0.0064, +0.0087] \\
\cmidrule{1-6}

\multirow{3}{*}{Crop}
& L & 0.6753 & +0.1712 / [+0.1643, +0.1783] & +0.2233 / [+0.2176, +0.2289] & +0.2577 / [+0.2526, +0.2629] \\
& M & 0.6787 & +0.0442 / [+0.0372, +0.0511] & +0.0736 / [+0.0655, +0.0814] & +0.1557 / [+0.1487, +0.1626] \\
& H & 0.6785 & -0.0324 / [-0.0399, -0.0250] & -0.0350 / [-0.0431, -0.0259] & +0.0758 / [+0.0683, +0.0835] \\
\cmidrule{1-6}

Flip
& Random & 0.6670 & +0.2117 / [+0.2053, +0.2183] & +0.2295 / [+0.2239, +0.2354] & +0.2811 / [+0.2764, +0.2859] \\
\bottomrule
\end{tabular}}
\label{SpCon_Unseen_bootstrap_AUROC}
\vspace{-0.2cm}
\end{table*}

\begin{table*}[t]
\centering
\caption{Significance analysis under Unseen post-processing operations for \textcolor{refcolor}{Table}~\ref{SpCon_Unseen} (AP).}
\vspace{0.1cm}
\scriptsize
\setlength{\tabcolsep}{3pt}
\renewcommand{\arraystretch}{1.05}
\resizebox{\textwidth}{!}{
\begin{tabular}{lcc|ccc}
\toprule
\textbf{Operation} & \textbf{Severity} & \textbf{Best Baseline} & \textbf{SpCon-TO $\Delta$ / 95\% CI} & \textbf{SpCon-ER $\Delta$ / 95\% CI} & \textbf{SpCon-KR $\Delta$ / 95\% CI} \\
\midrule
\multirow{3}{*}{WebP}
& L & 0.7164 & -0.2419 / [-0.2620, -0.2241] & +0.0883 / [+0.0706, +0.1035] & +0.2271 / [+0.2107, +0.2408] \\
& M & 0.4479 & +0.0882 / [+0.0697, +0.1027] & +0.2125 / [+0.1949, +0.2260] & +0.3515 / [+0.3339, +0.3647] \\
& H & 0.3760 & +0.0878 / [+0.0713, +0.1013] & +0.2113 / [+0.1955, +0.2239] & +0.3309 / [+0.3142, +0.3437] \\
\cmidrule{1-6}

\multirow{3}{*}{Sharpen}
& L & 0.9526 & -0.0382 / [-0.0513, -0.0235] & +0.0383 / [+0.0270, +0.0509] & +0.0464 / [+0.0347, +0.0595] \\
& M & 0.8730 & -0.0378 / [-0.0586, -0.0169] & +0.1037 / [+0.0843, +0.1231] & +0.1236 / [+0.1033, +0.1431] \\
& H & 0.7650 & -0.0195 / [-0.0444, +0.0026] & +0.1750 / [+0.1528, +0.1958] & +0.2221 / [+0.1997, +0.2424] \\
\cmidrule{1-6}

\multirow{3}{*}{Contrast}
& L & 0.9443 & -0.0943 / [-0.1082, -0.0806] & +0.0429 / [+0.0316, +0.0540] & +0.0540 / [+0.0431, +0.0654] \\
& M & 0.8102 & -0.0470 / [-0.0692, -0.0268] & +0.1420 / [+0.1213, +0.1590] & +0.1764 / [+0.1550, +0.1945] \\
& H & 0.6745 & -0.0128 / [-0.0386, +0.0077] & +0.2058 / [+0.1842, +0.2233] & +0.2696 / [+0.2472, +0.2868] \\
\cmidrule{1-6}

\multirow{3}{*}{Saturation}
& L & 0.9693 & -0.0314 / [-0.0406, -0.0224] & +0.0268 / [+0.0181, +0.0346] & +0.0303 / [+0.0218, +0.0381] \\
& M & 0.9567 & -0.0433 / [-0.0555, -0.0330] & +0.0367 / [+0.0261, +0.0457] & +0.0425 / [+0.0321, +0.0518] \\
& H & 0.9293 & -0.0596 / [-0.0740, -0.0474] & +0.0568 / [+0.0430, +0.0683] & +0.0693 / [+0.0562, +0.0808] \\
\cmidrule{1-6}

\multirow{3}{*}{Crop}
& L & 0.2175 & +0.2303 / [+0.2170, +0.2447] & +0.3546 / [+0.3396, +0.3697] & +0.4539 / [+0.4369, +0.4711] \\
& M & 0.2196 & +0.0330 / [+0.0241, +0.0416] & +0.0990 / [+0.0866, +0.1115] & +0.1902 / [+0.1758, +0.2053] \\
& H & 0.2173 & -0.0307 / [-0.0378, -0.0250] & -0.0100 / [-0.0185, -0.0014] & +0.0624 / [+0.0522, +0.0723] \\
\cmidrule{1-6}

Flip
& Random & 0.2063 & +0.3665 / [+0.3515, +0.3798] & +0.3732 / [+0.3579, +0.3886] & +0.5477 / [+0.5326, +0.5617] \\
\bottomrule
\end{tabular}}
\label{SpCon_Unseen_bootstrap_AP}
\vspace{-0.2cm}
\end{table*}

\textcolor{refcolor}{Tables}~\ref{SpCon_Unseen_bootstrap_AUROC} and~\ref{SpCon_Unseen_bootstrap_AP} report the significance comparisons for AUROC and AP, respectively. Each entry gives the performance difference between SpCon and the strongest official baseline for the corresponding condition and metric, together with its 95\% confidence interval. SpCon-KR achieves significant improvements in all 32 comparisons, while SpCon-ER does so in 29, with the only exceptions occurring for AUROC under WebP (L, QF = 90) and for both metrics under severe Crop. SpCon-TO achieves significant gains in 8 comparisons, further supporting the importance of reference-assisted modeling for generalization to Unseen post-processing.

%======================================================================================================

\section{Closed-Set Seven-Way Source Attribution}
\label{Closed_Set_Attribution}

\subsection{Closed-Set Composition}
The main formulation evaluates one candidate target at a time. We additionally compose the seven candidate-conditioned scores into a single closed-set source decision. Let $\{\mathcal{M}_1,\ldots,\mathcal{M}_7\}$ denote the seven candidate sources. For a query $\mathbf{x}$, candidate $\mathcal{M}_m$ produces its own response $\mathbf{z}_m(\mathbf{x})$ through the corresponding frozen VAE. Its positive evidence is
\begin{equation}
u_m(\mathbf{x})
=
L_{\mathcal{A}^{+}_m}
\bigl(\mathbf{z}_m(\mathbf{x})\bigr),
\end{equation}
while its KR evidence is
\begin{equation}
e^{\mathrm{KR}}_m(\mathbf{x})
=
L_{\mathcal{A}^{+}_m}\bigl(\mathbf{z}_m(\mathbf{x})\bigr)
-
L_{\mathcal{A}^{-}_m}\bigl(\mathbf{z}_m(\mathbf{x})\bigr).
\end{equation}
For belonging-only scoring, we set $e_m(\mathbf{x})=u_m(\mathbf{x})$; for KR, we set $e_m(\mathbf{x})=e_m^{\mathrm{KR}}(\mathbf{x})$. The closed-set prediction is then
\begin{equation}
\widehat{m}(\mathbf{x})
=
\arg\max_{m\in\{1,\ldots,7\}}
e_m(\mathbf{x}).
\end{equation}

\subsection{Belonging-Only Calibration Controls}
Raw positive log-likelihoods may have candidate-dependent locations and scales. To test whether the gain of KR can be explained by fixed candidate-wise calibration alone, we construct z-score and empirical percentile controls using only held-out belonging responses.

For candidate $\mathcal{M}_m$, let $\mu_m$, $\sigma_m$, and $\widehat{F}_m$ denote the mean, standard deviation, and empirical cumulative distribution function estimated from 700 held-out belonging scores $u_m$. These scores use the same 100 base images (calibration set indices 100-199) under Clean and the six fitted medium-severity conditions. These images are disjoint from the density-fitting images (calibration set indices 0-99) and all test set and platform queries. The two calibrated scores are
\begin{equation}
e^{z}_m(\mathbf{x})
=
\frac{u_m(\mathbf{x})-\mu_m}{\sigma_m},
\qquad
e^{p}_m(\mathbf{x})
=
\widehat{F}_m
\bigl(u_m(\mathbf{x})\bigr).
\end{equation}
Neither calibration uses non-belonging images, Unseen conditions, platform queries, or test labels.

\subsection{Evaluation Protocol}
The evaluation uses the complete test cache. Each synthetic source-condition cell contains 500 queries, while each source-platform cell contains 100 queries. All seven true sources are
generated models; no natural image is included as a query or candidate class. Results are macro-averaged over the seven true sources. Within Seen and Unseen, severities are first averaged within each operation family, followed by equal averaging over operation families. Random Top-1 accuracy is 1/7=0.1429.

\begin{table}[t]
\centering
\caption{Seven-way closed-set source-attribution Top-1 accuracy. The z-score and percentile controls use only held-out belonging responses and no non-belonging reference.}
\vspace{0.1cm}
\scriptsize
\setlength{\tabcolsep}{5pt}
\renewcommand{\arraystretch}{1}
\resizebox{0.75\columnwidth}{!}{
\begin{tabular}{l|cccc}
\toprule
\textbf{Evidence}
& \textbf{Clean}
& \textbf{Seen}
& \textbf{Unseen}
& \textbf{Platforms} \\
\midrule
Random
& 0.1429
& 0.1429
& 0.1429
& 0.1429 \\
Raw $L_{\mathcal{A}^{+}}$
& \underline{0.9731}
& \underline{0.6725}
& \underline{0.7227}
& \underline{0.6343} \\
$L_{\mathcal{A}^{+}}$ with z-score
& 0.9700
& 0.6588
& 0.7080
& 0.6205 \\
$L_{\mathcal{A}^{+}}$ with percentile calibration
& 0.9683
& 0.6567
& 0.7088
& 0.6143 \\
\rowcolor{skyblue!40}
SpCon-KR
& \textbf{0.9983}
& \textbf{0.8472}
& \textbf{0.8498}
& \textbf{0.8724} \\
\bottomrule
\end{tabular}}
\label{Closed_Set_Table}
\vspace{-0.2cm}
\end{table}

\textbf{Results.}
Candidate-specific belonging-only calibration does not improve the raw positive likelihood. Compared with uncalibrated $L_{\mathcal{A}^{+}}$, z-score calibration lowers Top-1 accuracy from 0.6725 to 0.6588 on Seen, 0.7227 to 0.7080 on Unseen, and 0.6343 to 0.6205 on platform inputs; empirical-percentile calibration shows similar reductions (see \textcolor{refcolor}{Table}~\ref{Closed_Set_Table}). Thus, KR's gain cannot be explained by fixed candidate-dependent location or scale corrections alone, indicating that its query-dependent non-belonging likelihood provides additional cross-candidate evidence. SpCon-KR achieves 0.9983 Top-1 accuracy on Clean and 0.8472/0.8498/0.8724 on Seen/Unseen/Platform inputs. Since all query and candidate classes are generated sources, this shows that the closed-set result reflects discrimination among the seven generators rather than a generated-versus-natural shortcut.

\begin{table}[t]
\centering
\caption{Largest directed confusions of SpCon-KR in the seven-way
closed-set evaluation. An entry
$\mathcal{M}_a\rightarrow\mathcal{M}_b$
denotes the fraction of queries from source $\mathcal{M}_a$ assigned to
candidate $\mathcal{M}_b$.}
\vspace{0.1cm}
\scriptsize
\setlength{\tabcolsep}{7pt}
\renewcommand{\arraystretch}{1}
\resizebox{0.8\columnwidth}{!}{
\begin{tabular}{l|cc|cc}
\toprule
\textbf{Scope}
& \textbf{Largest Confusion}
& \textbf{Rate}
& \textbf{Second Largest}
& \textbf{Rate} \\
\midrule
Clean
& PixArt $\rightarrow$ FLUX.1
& 0.0040
& SD2 $\rightarrow$ SD1
& 0.0020 \\
Seen
& SD2 $\rightarrow$ SD1
& 0.1837
& FLUX.2 $\rightarrow$ FLUX.1
& 0.0863 \\
Unseen
& SD2 $\rightarrow$ SD1
& 0.2158
& FLUX.1 $\rightarrow$ SD3
& 0.0559 \\
Platforms
& SD2 $\rightarrow$ SD1
& 0.1767
& FLUX.2 $\rightarrow$ SD3
& 0.1333 \\
\bottomrule
\end{tabular}}
\label{Closed_Set_Confusions}
\vspace{-0.2cm}
\end{table}

The dominant processed-image confusion is SD2 $\rightarrow$ SD1 (see \textcolor{refcolor}{Table}~\ref{Closed_Set_Confusions}), two generators from the same model family. This evaluation assumes that the true source lies in the candidate set.
%======================================================================================================
\section{Attribution Granularity under Shared VAEs}
\label{Capability_Boundary}

\begin{table}[t]
\centering
\caption{Fine-grained checkpoint attribution within the SD3 shared-VAE family~\citep{SD3}. Random-chance AUROC is 0.5000; random-chance AP is 0.5000 for 1-vs-1 discrimination and 0.3333 for 1-vs-2 discrimination.}
\vspace{0.1cm}
\scriptsize
\setlength{\tabcolsep}{3pt}
\renewcommand{\arraystretch}{1.1}
\resizebox{\columnwidth}{!}{
\begin{tabular}{l|cccccc}
\toprule
\textbf{Method} & \textbf{M vs L} & \textbf{M vs LT} & \textbf{L vs LT} & \textbf{M vs L+LT} & \textbf{L vs M+LT} & \textbf{LT vs M+L} \\
\cmidrule{1-7}
SpCon-TO
& \underline{0.5276} / \underline{0.5333}
& 0.6525 / 0.6871
& 0.7099 / 0.7552
& 0.5901 / \underline{0.4305}
& 0.6451 / 0.5014
& \underline{0.6725} / \underline{0.5458} \\
SpCon-ER
& 0.5006 / 0.5001
& \textbf{0.7037} / \textbf{0.7415}
& \textbf{0.7718} / \textbf{0.8186}
& \underline{0.6022} / 0.4205
& \textbf{0.6891} / \textbf{0.5414}
& 0.6385 / 0.5173 \\
SpCon-KR
& \textbf{0.5518} / \textbf{0.5597}
& \underline{0.6797} / \underline{0.7202}
& \underline{0.7289} / \underline{0.7797}
& \textbf{0.6158} / \textbf{0.4648}
& \underline{0.6514} / \underline{0.5017}
& \textbf{0.6966} / \textbf{0.5650} \\
\bottomrule
\end{tabular}}
\label{Capability}
\vspace{-0.2cm}
\end{table}

To characterize SpCon's attribution boundary, we evaluate SD3.5-Medium (M), SD3.5-Large (L), and SD3.5-Large-Turbo (LT)~\citep{SD3}, whose VAE configurations and weights are byte-identical. For each comparison, the checkpoint on the left is treated as belonging; its positive distribution is fitted using 100 DNA-30K calibration images under Clean and six medium-severity Seen operations, while 500 Clean images per checkpoint are reserved for evaluation. Other shared-VAE checkpoints are excluded from ER and KR reference fitting. As expected when the queried operator is identical, separation is weaker (see \textcolor{refcolor}{Table}~\ref{Capability}) than among generators with distinct VAEs, including SD1 and SD2 from the same family, while comparisons involving LT retain moderate separability; since all checkpoints share prompt identities and frequencies, this residual separation likely reflects checkpoint-dependent output statistics. These results characterize SpCon as a VAE-family attribution method, consistent with its design of querying only the target VAE; model-level discrimination within a shared-VAE family calls for complementary evidence (see \textcolor{refcolor}{Section}~\ref{Discussion}).

%==============================================================================================

\section{Representation-Level Controls}
\label{Repr_Controls_Sec}

\begin{table}[t]
\centering
\caption{Representation-level controls under the ablation protocol of \textcolor{refcolor}{Table}~\ref{One_Two} (AUROC / AP). \emph{All-13} averages severities within each operation family and weights Clean and the twelve families equally; the Unseen column is family-balanced. Colored cells indicate the adopted configuration.}
\vspace{0.1cm}
\scriptsize
\setlength{\tabcolsep}{5pt}
\renewcommand{\arraystretch}{1}
\resizebox{0.95\linewidth}{!}{
\begin{tabular}{cc|cccc}
\toprule
\textbf{Response} & \textbf{Type} & \textbf{Clean} & \textbf{Seen} & \textbf{Unseen} & \textbf{All-13} \\
\midrule
\multirow[c]{3}{*}{\makecell{First Residual\\(48D)}}
& TO & 0.8059 / 0.4865 & 0.7692 / 0.4047 & 0.7701 / 0.3859 & 0.7725 / 0.4023 \\
& ER & 0.9159 / 0.7351 & 0.8422 / 0.5294 & 0.8414 / 0.5295 & 0.8475 / 0.5453 \\
& KR & 0.9708 / 0.8745 & 0.9156 / 0.6939 & 0.9083 / 0.6747 & 0.9165 / 0.6989 \\
\cmidrule{1-6}
\multirow[c]{3}{*}{\makecell{Normalized\\First Residual (48D)}}
& TO & 0.7997 / 0.4463 & 0.7787 / 0.4180 & 0.7672 / 0.3847 & 0.7750 / 0.4048 \\
& ER & 0.9387 / 0.7746 & 0.8682 / 0.5731 & 0.8496 / 0.5568 & 0.8651 / 0.5811 \\
& KR & 0.9686 / 0.8755 & 0.9232 / 0.7226 & 0.9080 / 0.6910 & 0.9197 / 0.7198 \\
\cmidrule{1-6}
\multirow[c]{3}{*}{\makecell{Concatenation\\(96D)}}
& TO & 0.9029 / 0.7214 & 0.8335 / 0.5382 & 0.8346 / 0.5135 & 0.8393 / 0.5409 \\
& ER & 0.9958 / 0.9873 & 0.9448 / 0.7957 & 0.9345 / 0.7755 & 0.9440 / 0.8011 \\
& KR & 0.9998 / 0.9986 & 0.9672 / 0.8745 & 0.9602 / 0.8449 & 0.9665 / 0.8704 \\
\cmidrule{1-6}
& TO
& \cellcolor{skyblue!20}0.9807 / 0.9469
& \cellcolor{skyblue!20}0.8691 / 0.5911
& \cellcolor{skyblue!20}0.8795 / 0.6389
& \cellcolor{skyblue!20}0.8825 / 0.6405 \\
& ER
& \cellcolor{skyblue!30}0.9989 / 0.9952
& \cellcolor{skyblue!30}0.9494 / 0.7917
& \cellcolor{skyblue!30}0.9261 / 0.7587
& \cellcolor{skyblue!30}0.9425 / 0.7921 \\
\multirow[c]{-3}{*}{\makecell{\textbf{SpCon}\\\textbf{(48D)}}}
& KR
& \cellcolor{skyblue!40}0.9999 / 0.9994
& \cellcolor{skyblue!40}0.9699 / 0.8681
& \cellcolor{skyblue!40}0.9555 / 0.8314
& \cellcolor{skyblue!40}0.9655 / 0.8613 \\
\bottomrule
\end{tabular}}
\label{Repr_Controls}
\vspace{-0.2cm}
\end{table}

\textcolor{refcolor}{Table}~\ref{One_Two} shows that a single 48-dimensional residual spectrum cannot match SpCon. Two further controls isolate why. The first tests per-channel centering of the first log-energy spectrum as a substitute for the second residual. The second evaluates the inductive bias of the relative response: the 96-dimensional concatenation contains the 48-dimensional log difference as the linear map $\mathbf{z}=[\spcId,-\spcId]\mathbf{z}_{\mathrm{cat}}$.

\textbf{Controls.} \textcolor{refcolor}{Table}~\ref{Repr_Controls} compares four response representations. \emph{First Residual} uses the 48-dimensional log band power of $\mathbf{r}_1$. \emph{Normalized First Residual} subtracts, per image and per channel, the mean of that channel's 16 log band powers, and never invokes the second reconstruction. \emph{Concatenation} stacks both levels into 96 dimensions. All four representations share the reconstruction operator, reference sources, fitting conditions, and density backend of \textcolor{refcolor}{Table}~\ref{One_Two}; density parameters are fitted separately for each representation. The First Residual and SpCon blocks reuse the runs of \textcolor{refcolor}{Table}~\ref{One_Two}; their Unseen entries differ slightly because Unseen operation families are weighted equally here.

\textbf{The second reconstruction is not replaceable by single-level centering.} Under All-13, SpCon exceeds the normalized first-residual control by 0.1074/0.2358, 0.0774/0.2110, and 0.0458/0.1415 AUROC/AP under TO, ER, and KR, respectively. Both representations have 48 dimensions and share the same backend, so the gap is attributable to the band-wise reference supplied by the second reconstruction.

\textbf{The relative response is a compact inductive bias.} The 96-dimensional concatenation contains the log difference as the linear map $\mathbf{z}=[\spcId,-\spcId]\mathbf{z}_{\mathrm{cat}}$. Under ER and KR, the 48-dimensional response performs on par with concatenation (All-13 differences within 0.0015 AUROC and 0.0091 AP) at half the dimension, with 1,176 rather than 4,656 unique scale-matrix entries. Under TO, where no non-belonging density supplies contrast at scoring time, SpCon outperforms concatenation by 0.0432 AUROC and 0.0996 AP: cancelling the component shared by the two residual levels removes input-dependent variation that a belonging-only density cannot separate. We therefore adopt the log difference as the SpCon response.

%======================================================================================================
\section{Compositional Post-Processing Stress Test}
\label{Chained_Postprocessing}

\textbf{Protocol.}
The main synthetic benchmark applies one controlled operation to each image. To examine compositional generalization separately, we construct three two-operation chains and three three-operation chains:
\begin{equation}
\begin{aligned}
&\text{Resize}\rightarrow\text{JPEG},
\qquad
\text{Brightness}\rightarrow\text{JPEG},
\qquad
\text{Blur}\rightarrow\text{JPEG},
\\
&\text{Noise}\rightarrow\text{Resize}\rightarrow\text{JPEG},
\\
&\text{Brightness}\rightarrow\text{Resize}\rightarrow\text{JPEG},
\\
&\text{Rotation}\rightarrow\text{Resize}\rightarrow\text{JPEG}.
\end{aligned}
\end{equation}
All constituent operations use the medium severities of the frozen single-operation protocol: Resize uses a side-length factor of $6/8$, Brightness a factor of $1.6$, Blur a kernel size of $5$, Noise $\spcSigN=10$, Rotation $\pm5^\circ$, and JPEG quality factor $75$. JPEG is placed last to represent a final re-encoding stage.

None of the six chains participates in reference fitting. The SpCon variants retain the same frozen Clean and single-operation reference components used by the main protocol, while AEDR-Matched is evaluated with its unchanged native scorer. Evaluation uses 100 images per source from an independent test split. Each target-chain cell contains 100 belonging and 700 non-belonging images, and results are macro-averaged over the seven targets. This stress test is separate from the complete 35-condition benchmark.

\textbf{Input Distortion Measurement.}
We additionally measure the pixel-level difference between each clean input $\mathbf{x}$ and its chained version $\widetilde{\mathbf{x}}$. Both images undergo the same RGB normalization and bicubic-antialias input alignment used during feature extraction. For candidate VAE $m$, let $\spcAlign_m$ denote the alignment transform of \textcolor{refcolor}{Equation}~\ref{Double_Recon}; we compute
\begin{equation}
\operatorname{PSNR}_{m}
(\mathbf{x},\widetilde{\mathbf{x}})
=
10\log_{10}
\frac{4}{
\operatorname{MSE}
\left(
\spcAlign_m(\mathbf{x}),
\spcAlign_m(\widetilde{\mathbf{x}})
\right)
},
\end{equation}
where the numerator is $2^2=4$ because the aligned inputs lie in $[-1,1]$. PSNR is used only to quantify input distortion and does not participate in density fitting or attribution scoring. Each reported mean and standard deviation aggregates 7$\times$8$\times$100=5,600 candidate-source-image comparisons.

\begin{table*}[t]
\centering
\caption{Attribution under six previously unseen sequential post-processing chains (AUROC / AP). Input PSNR is reported as mean $\pm$ standard deviation over candidate-source-image comparisons after candidate-VAE input alignment. Bold marks the best AUROC and AP within each row.}
\vspace{0.1cm}
\scriptsize
\setlength{\tabcolsep}{2.5pt}
\renewcommand{\arraystretch}{1.1}
\resizebox{\textwidth}{!}{
\begin{tabular}{lcc|cccc}
\toprule
\textbf{Sequential Operations}
& \textbf{Steps}
& \textbf{Input PSNR (dB)}
& \textbf{AEDR-Matched}
& \textbf{SpCon-TO}
& \textbf{SpCon-ER}
& \textbf{SpCon-KR} \\
\midrule
Resize $\rightarrow$ JPEG
& 2
& $34.33\pm4.30$
& 0.6675 / 0.3633
& 0.5683 / 0.1920
& \textbf{0.7877} / \textbf{0.4276}
& 0.7677 / 0.3308 \\
Brightness $\rightarrow$ JPEG
& 2
& $13.42\pm1.54$
& 0.7755 / 0.5058
& 0.8228 / 0.5106
& 0.9251 / 0.7094
& \textbf{0.9568} / \textbf{0.8126} \\
Blur $\rightarrow$ JPEG
& 2
& $32.61\pm5.12$
& 0.6373 / 0.2828
& 0.6798 / 0.3227
& 0.8533 / 0.5191
& \textbf{0.8654} / \textbf{0.5508} \\
\midrule
Noise $\rightarrow$ Resize $\rightarrow$ JPEG
& 3
& $32.43\pm3.01$
& 0.6194 / \textbf{0.3400}
& 0.4846 / 0.1497
& 0.6639 / 0.2751
& \textbf{0.6645} / 0.2249 \\
Brightness $\rightarrow$ Resize $\rightarrow$ JPEG
& 3
& $13.43\pm1.54$
& 0.6837 / \textbf{0.3611}
& 0.5481 / 0.1897
& \textbf{0.7336} / 0.3581
& 0.7271 / 0.2912 \\
Rotation $\rightarrow$ Resize $\rightarrow$ JPEG
& 3
& $14.81\pm2.33$
& 0.4557 / 0.1159
& 0.4587 / 0.1461
& 0.5250 / \textbf{0.1733}
& \textbf{0.5365} / 0.1696 \\
\bottomrule
\end{tabular}}
\label{Chained_Postprocessing_Table}
\vspace{-0.2cm}
\end{table*}

\begin{table}[t]
\centering
\caption{Aggregated performance under single and sequential post-processing scopes (AUROC / AP). The single-operation row uses the six medium-severity operations involved in the chains.}
\vspace{0.1cm}
\scriptsize
\setlength{\tabcolsep}{5pt}
\renewcommand{\arraystretch}{1}
\resizebox{0.9\columnwidth}{!}{
\begin{tabular}{l|cccc}
\toprule
\textbf{Evaluation Scope}
& \textbf{AEDR-Matched}
& \textbf{SpCon-TO}
& \textbf{SpCon-ER}
& \textbf{SpCon-KR} \\
\midrule
JPEG Q75 only
& 0.7354 / 0.4953
& 0.8901 / 0.6832
& 0.9479 / 0.7791
& \textbf{0.9818} / \textbf{0.9149} \\
Single medium operation
& 0.7353 / 0.4497
& 0.9050 / 0.7032
& 0.9623 / 0.8297
& \textbf{0.9838} / \textbf{0.9208} \\
Two-operation chains
& 0.6934 / 0.3840
& 0.6903 / 0.3418
& 0.8554 / 0.5521
& \textbf{0.8633} / \textbf{0.5647} \\
Three-operation chains
& 0.5863 / \textbf{0.2723}
& 0.4972 / 0.1618
& 0.6408 / 0.2688
& \textbf{0.6427} / 0.2286 \\
All chains
& 0.6399 / 0.3281
& 0.5937 / 0.2518
& 0.7481 / \textbf{0.4104}
& \textbf{0.7530} / 0.3966 \\
\bottomrule
\end{tabular}}
\label{Chained_Postprocessing_Summary}
\vspace{-0.2cm}
\end{table}

\textbf{Results.}
On two-operation chains (\textcolor{refcolor}{Tables}~\ref{Chained_Postprocessing_Table} and~\ref{Chained_Postprocessing_Summary}), SpCon-ER and SpCon-KR retain AUROC/AP of 0.8554/0.5521 and 0.8633/0.5647, exceeding AEDR-Matched on both metrics; three-operation chains reduce them to 0.6408/0.2688 and 0.6427/0.2286. Attribution is not monotonic in pixel-level distortion: Brightness $\rightarrow$ JPEG has the lowest mean PSNR (13.42\,dB), yet SpCon-KR retains 0.9568/0.8126, whereas Resize $\rightarrow$ JPEG (34.33\,dB) reaches 0.7677/0.3308. Under KR, every chain containing Resize stays below 0.77 AUROC, although Resize alone retains 0.9988 (\textcolor{refcolor}{Table}~\ref{SpCon_Seen}); degradation thus stems mainly from composing resampling with other operations rather than from distortion magnitude, pointing to composition-aware reference components as a direct extension.
%==============================================================================

\section{Adaptive-Adversary Stress Test}
\label{Adaptive_Adversary}

The main protocol evaluates non-adversarial post-processing. This section complements it with adaptive perturbations under three levels of attacker knowledge.

\subsection{Threat Model}
\label{Adaptive_Threat}

The attacker holds a belonging image and seeks to have it ranked as non-belonging under the target VAE. We consider three knowledge levels: \textbf{(i) Transferred.} The attacker knows that attribution is reconstruction-based but cannot query the audited VAE. Perturbations are optimized against an SD1.5 surrogate VAE and evaluated on SDXL and FLUX.1, which take no part in the optimization. \textbf{(ii) Representation-aware.} The attacker knows the spectral-contraction representation of \textcolor{refcolor}{Equation}~\ref{SpCon_Z} but not the auditor's fitted densities. The attack optimizes 16 radial band gains so that the response moves toward the mean response of an independent ImageNet reference set. \textbf{(iii) Full knowledge.} The attacker additionally holds the fitted components $\mathcal{A}^{+}$ and $\mathcal{A}^{-}_{\mathrm{KR}}$, and directly minimizes the frozen KR score of \textcolor{refcolor}{Equation}~\ref{Attribution_Score_Eq}. \textbf{(i)} and \textbf{(ii)} require only publicly released artifacts. \textbf{(iii)} additionally requires the auditor-side fitted densities, which are not part of any released model.

\subsection{Protocol}
\label{Adaptive_Protocol}

Attacks are applied to \emph{Clean} images only. This experiment therefore measures adaptive evasion on unprocessed inputs and is reported separately from the $1{+}18{+}16$ macro-averages of \textcolor{refcolor}{Section}~\ref{Experimental_Setup}.

For each candidate, 100 belonging images are attacked and compared against 700 fixed, unmodified Clean non-belonging images, preserving the $1/8$ positive ratio of the main protocol (random-ranking AP $=0.125$). Only positives are perturbed, so all reported degradation is attributable to the attack rather than to a changed test composition. Perturbations are bounded in $L_\infty$ with $\epsilon\in\{1,2,4,8\}/255$; PGD runs 30 steps of size $\epsilon/10$, with each larger budget warm-started from the solution at the next smaller budget, and band equalization runs 15 steps. Every perturbed image is projected to the pixel-domain budget and persisted as an 8-bit PNG; AUROC, AP, PSNR, SSIM, and AlexNet-LPIPS are all recomputed from the stored files, and all 6{,}400 stored images satisfy a measured $\|\tilde{\mathbf{x}}-\mathbf{x}\|_\infty\le\epsilon$. PSNR is converted from the cell-averaged pixel MSE. The reference densities stay frozen: the auditor performs no adversarial fitting. Results are computed per candidate and macro-averaged with equal weight. OCC-CLIP is excluded from this adaptive stress test because its unattacked clean performance is substantially weaker than the reconstruction-based baselines, making additional attack-induced degradation less informative; its non-adversarial results remain in the main experiments.

\begin{table}[t]
\centering
\caption{SpCon-KR under adaptive perturbations (AUROC / AP), by attacker knowledge and $L_\infty$ budget. Levels (i) and (ii) use only publicly released artifacts. The three tiers differ in attack information and candidate coverage--Transferred PGD reports two frozen victims (SDXL, FLUX.1), whereas the other two cover all seven candidates--and are therefore reported per tier rather than ranked against one another.}
\vspace{0.1cm}
\scriptsize
\setlength{\tabcolsep}{7pt}
\renewcommand{\arraystretch}{1}
\resizebox{0.95\columnwidth}{!}{
\begin{tabular}{c|cccc|>{\columncolor{skyblue!30}}c}
\toprule
\textbf{Attack} & \textbf{Budget} & \textbf{PSNR (dB)} & \textbf{SSIM} & \textbf{LPIPS}
& \multicolumn{1}{c}{\textbf{SpCon-KR}} \\
\midrule

\multirow{5}{*}{\makecell[c]{(i) Transferred PGD\\[-1pt]\scriptsize SDXL, FLUX.1}}
& Unattacked & $\infty$ & 1.0000 & 0.0000 & 1.0000 / 0.9998 \\
& $1/255$ & 50.73 & 0.9949 & 0.0039 & 0.9991 / 0.9951 \\
& $2/255$ & 46.91 & 0.9886 & 0.0168 & 0.9954 / 0.9769 \\
& $4/255$ & 42.81 & 0.9727 & 0.0535 & 0.9752 / 0.8907 \\
& $8/255$ & 38.86 & 0.9388 & 0.1077 & 0.9521 / 0.7819 \\
\cmidrule{1-6}

\multirow{5}{*}{\makecell[c]{(ii) Band Equalization\\[-1pt]\scriptsize 7 candidates}}
& Unattacked & $\infty$ & 1.0000 & 0.0000 & 0.9999 / 0.9995 \\
& $1/255$ & 48.39 & 0.9975 & 0.0023 & 0.9997 / 0.9984 \\
& $2/255$ & 42.92 & 0.9891 & 0.0140 & 0.9983 / 0.9902 \\
& $4/255$ & 37.28 & 0.9653 & 0.0410 & 0.9867 / 0.9298 \\
& $8/255$ & 31.93 & 0.9239 & 0.0755 & 0.9463 / 0.7598 \\
\cmidrule{1-6}

\multirow{5}{*}{\makecell[c]{(iii) White-box PGD\\[-1pt]\scriptsize 7 candidates}}
& Unattacked & $\infty$ & 1.0000 & 0.0000 & 0.9999 / 0.9995 \\
& $1/255$ & 50.66 & 0.9951 & 0.0030 & 0.3105 / 0.0880 \\
& $2/255$ & 46.85 & 0.9884 & 0.0127 & 0.1975 / 0.0778 \\
& $4/255$ & 42.84 & 0.9724 & 0.0388 & 0.1629 / 0.0755 \\
& $8/255$ & 39.20 & 0.9438 & 0.0755 & 0.1471 / 0.0746 \\
\bottomrule
\end{tabular}}
\label{Adaptive_Main}
\vspace{-0.2cm}
\end{table}

\subsection{Attacks from Publicly Available Knowledge}
\label{Adaptive_Public}

Attacks that do not read auditor-side parameters leave attribution largely intact (see \textcolor{refcolor}{Table}~\ref{Adaptive_Main}). At $\epsilon=8/255$, band equalization reduces AUROC from 0.9999 to 0.9463 and transferred PGD from 1.0000 to 0.9521, a loss of at most 0.054; AP decreases to 0.7598 and 0.7819, respectively. At $\epsilon\le2/255$, both attacks remain above 0.9950 AUROC. At $\epsilon=8/255$, band equalization yields 31.93\,dB PSNR and 0.0755 LPIPS, while transferred PGD yields 38.86\,dB PSNR and 0.1077 LPIPS.

Two mechanisms are consistent with this outcome, neither tested here. Surrogate-VAE perturbations may transfer poorly because the response depends on the specific frozen operator queried rather than on reconstruction in general; band equalization may fail because the score depends on the joint channel-band configuration modeled by the full scale matrix (\textcolor{refcolor}{Appendix}~\ref{Scale_Matrix_Ablation}) rather than on any marginal band-wise statistic.

\subsection{Full-Knowledge White-Box Attack}
\label{Adaptive_Whitebox}
When the attacker holds the auditor's fitted densities, the KR score becomes a differentiable white-box objective. Consistent with the general vulnerability of differentiable forensic detectors under white-box access~\citep{White-Box}, minimizing it reduces AUROC/AP from 0.9999/0.9995 to 0.3105/0.0880 at $\epsilon=1/255$ and to 0.1471/0.0746 at $\epsilon=8/255$. Because the reference densities are fitted offline and never released with the model, the reference-based design allows them to remain auditor-side.

\subsection{Scope}
\label{Adaptive_Scope}

The tested attacks are applied to Clean inputs and cover transferred, representation-aware, and full-knowledge gradient perturbations. Randomized, generative-space, and query-based perturbations, attacks that promote non-belonging images as belonging, and adaptive attacks on top of post-processing are left to future work, together with rate-limited scoring interfaces and adversarial fitting of the reference densities.

\end{document}